\documentclass[acmtog]{acmart}

\newcommand{\final}{1}

\usepackage{color}
\usepackage{ifthen}
\usepackage{float}
\usepackage{alltt}
\usepackage{amsmath}
\usepackage{amsthm}
\usepackage{newlfont} 
\usepackage{floatflt}
\usepackage{wrapfig}
\usepackage{fixltx2e}
\usepackage{subfig} 
\usepackage{multirow}
\usepackage{booktabs}
\usepackage{cleveref}
\usepackage{hyperref}

\usepackage{xspace}
\usepackage[export]{adjustbox}
\usepackage{mathtools, cuted}
\usepackage{CJKutf8} 
\usepackage[ruled]{algorithm2e}

\usepackage{siunitx, booktabs, lipsum}
\usepackage[flushleft]{threeparttable}

\usepackage{xcolor}
\usepackage{arydshln}
\usepackage{adjustbox}
\setcitestyle{square}
\def\textrightarrow{$\rightarrow$}

\SetAlFnt{\small}
\SetAlCapFnt{\small}
\SetAlCapNameFnt{\small}
\SetAlCapHSkip{0pt}
\IncMargin{-\parindent}

\let\oldcaption\caption
\renewcommand{\caption}[2][]{\oldcaption[#1]{{\em #1} #2}}

\definecolor{SithColor}{rgb}{0.7,0,0} 
\newcommand{\qisun}[1]{{\color{SithColor} Qi: #1 $\qed$}}
\definecolor{ConsularColor}{rgb}{0,0.4,0} 
\definecolor{GuardianColor}{rgb}{0,0,0.8} 
\newcommand{\praneeth}[1]{{\color{GuardianColor} Praneeth: #1 $\qed$}}

\definecolor{zeshiBlue}{rgb}{0,0.0,0.9}

\newcommand{\yujie}[1]{{\color{ConsularColor} Yujie: #1}}
\newcommand{\warning}[1]{{\it\color{red} #1}}
\newcommand{\note}[1]{{\it\color{blue} #1}}
\newcommand{\nothing}[1]{}

\definecolor{AudioColor}{rgb}{0.56,0.34,0.62}

\definecolor{figred}{rgb}{1,0,0}
\definecolor{figgreen}{rgb}{0,0.6,0}
\definecolor{figblue}{rgb}{0,0,1}
\definecolor{figpink}{rgb}{1,0.63,0.63}

\newcommand{\revise}[1]{{\color{black} #1}}
\newcommand{\todo}[1]{{\it\color{orange} TODO: #1}}

\newcommand{\rerevise}[1]{{\color{black} #1}}

\ifthenelse{\equal{\final}{1}}
{
\renewcommand{\qisun}[1]{}
\renewcommand{\praneeth}[1]{}
\renewcommand{\yujie}[1]{}
\renewcommand{\warning}[1]{}
\renewcommand{\note}[1]{}
\renewcommand{\todo}[1]{}
\renewcommand{\revise}[1]{{#1}}
\renewcommand{\rerevise}[1]{{#1}}

}

\newcommand{\pseudocode}{Algorithm}
\floatstyle{plain}
\floatname{algorithm}{\pseudocode}

\newcommand{\filename}[1]{\url{#1}}
\newcommand{\foldername}[1]{\url{#1}}

\let\oldparagraph\paragraph
\iffalse

\renewcommand{\paragraph}[1]{\oldparagraph{\textbf{#1}.}} 
\else

\renewcommand{\paragraph}[1]{\oldparagraph{{#1}.}}
\fi

\ifdefined\email
\else
\newcommand{\email}[1]{\url{#1}}
\fi

\graphicspath{
{images/}
{images/source_images/}
{images/1080p_wild/}
}

\acmJournal{TOG}
\acmArticle{}
\acmDOI{}
\acmSubmissionID{papers\_xxx}

\begin{document}

\setlength{\abovecaptionskip}{1.0ex}
\setlength{\belowcaptionskip}{1.0ex}
\setlength{\floatsep}{1.0ex}
\setlength{\dblfloatsep}{\floatsep}
\setlength{\textfloatsep}{2.0ex}
\setlength{\dbltextfloatsep}{\textfloatsep}
\setlength{\abovedisplayskip}{1.0ex}
\setlength{\belowdisplayskip}{1.0ex}


\title{Pupil-Adaptive 3D Holography Beyond Coherent Depth-of-Field}

\author{Yujie Wang}
\affiliation{%
  \institution{Shandong University,}
  \city{Qingdao}
  \institution{Peking University and State Key Laboratory of General Artificial Intelligence}
  \city{Beijing}
  \country{China}
  }
 \email{yujiew.cn@gmail.com}

\author{Baoquan Chen}
\affiliation{%
\department{School of AI}
  \institution{Peking University and State Key Laboratory of General Artificial Intelligence}
  \city{Beijing}
  \country{China}
}
\email{baoquan@pku.edu.cn}
\authornote{Corresponding author}

\author{Praneeth Chakravarthula}
\affiliation{%
  \institution{UNC Chapel Hill}
  \city{Chapel Hill}
  \country{USA}
}
\email{cpk@cs.unc.edu}
 


\begin{teaserfigure}
\begin{center}
\includegraphics[width=18cm]{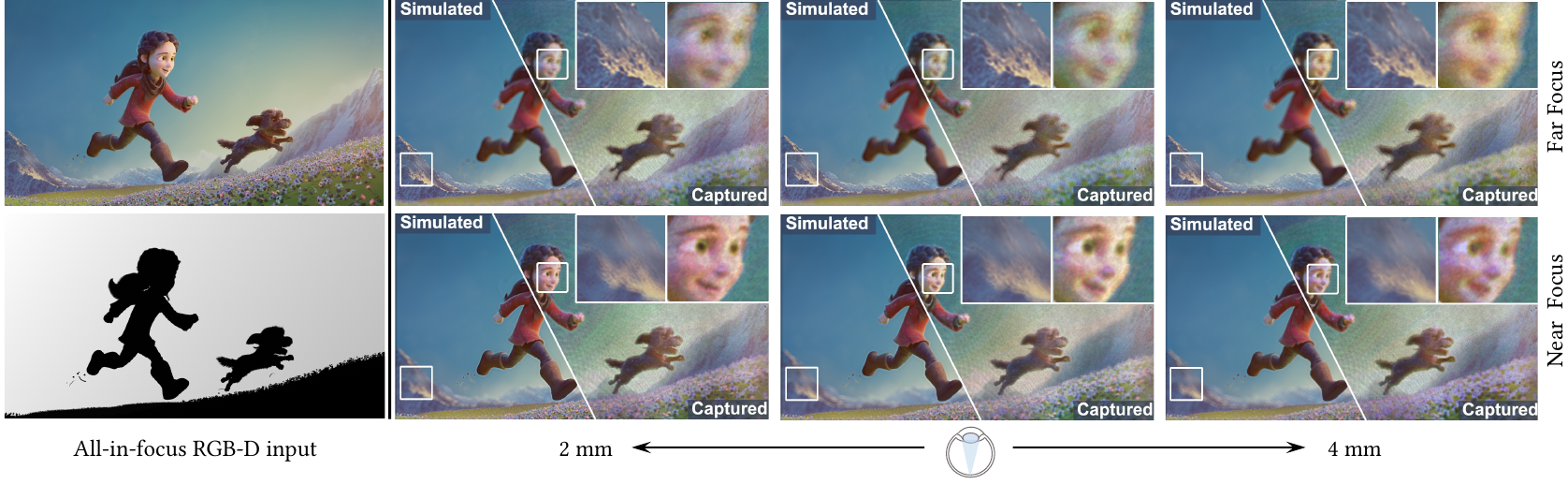}
\end{center}
\caption{\emph{Pupil-adaptive 3D holography}. In reality, the pupil size of human eyes continuously vary throughout the day.
When the pupil size changes, the degree of the observed defocus effects also changes accordingly. However, providing such natural pupil-dependent defocus effects produced by incoherent light in real world on a coherent light-based holographic display is challenging.
To overcome this, we propose a unified neural framework to synthesize 3D holograms producing natural defocus effects, accommodating various continuously changing eye pupil sizes. 
We validate our method both in simulations and on an experimental prototype display, and demonstrate its ability to produce pupil-dependent natural defocus blur, as highlighted in the insets above.
}
\label{fig:teaser}
\end{teaserfigure}

\begin{abstract}

Recent holographic display approaches propelled by deep learning have shown remarkable success in enabling high-fidelity holographic projections. However, these displays have still not been able to demonstrate realistic focus cues, and a major gap still remains between the defocus effects possible with a coherent light-based holographic display and those exhibited by incoherent light in the real world. Moreover, existing methods have not considered the effects of the observer's eye pupil size variations on the perceived quality of 3D projections, especially on the defocus blur due to varying depth-of-field of the eye. 

In this work, we propose a framework that bridges the gap between the coherent depth-of-field of holographic displays and what is seen in the real world due to incoherent light. To this end, we investigate the effect of varying shape and motion of the eye pupil on the quality of holographic projections, and devise a method that changes the depth-of-the-field of holographic projections dynamically in a pupil-adaptive manner. 
Specifically, we introduce a learning framework that adjusts the receptive fields on-the-go based on the current state of the observer's eye pupil to produce image effects that otherwise are not possible in current computer-generated holography approaches.
We validate the proposed method both in simulations and on an experimental prototype holographic display, and demonstrate significant improvements in the depiction of depth-of-field effects, outperforming existing approaches both qualitatively and quantitatively by at least 5 dB in peak signal-to-noise ratio. 

%
\end{abstract}

%
%



\maketitle

\section{Introduction}

Immersive virtual and augmented reality (AR/VR) systems are starting to become commonplace as wearable AR/VR displays are now readily available to consumers and at affordable prices. As these real and virtual experiences start to converge, it becomes imperative that the user experience through these visual displays is seamless. This has indeed been a topic of interest for a long time among the computer graphics, optics and vision science researchers, with significant focus on mitigating the visual fatigue caused by such displays~\cite{Konrad_2017,koulieris2017accommodation}. 
Nonetheless, the emulation of natural focus cues while maintaining image resolution in near-eye displays remains an ongoing research pursuit, and has not yet been fully realized.
Recently, computer-generated holographic (CGH) displays have shown the potential to support natural focus cues, high-resolution imagery, and aberration correction capabilities for both visual and optical errors, thereby providing more realistic and immersive user experiences~\cite{doublePhase,kim2022holographic}. 
Unlike conventional flat panel displays, holographic displays modulate an input light wave to create interference patterns that create a desired image. This control over the entire wavefront of light also enables holographic displays to create images with minimal optical elements, by encoding much of the optics into the SLM phase pattern, thereby enabling compact form factors and multi-focal capabilities required for near-eye display applications.

Recent studies have demonstrated that holographic displays can attain image quality approaching that of conventional flat-panel displays~\cite{choi22time, chakravarthula2022pupil,shi2022light, HoloNet}. 
These methods calibrate for hardware display errors and remove constraints on holographic phase calculations by incorporating neural network predictors \cite{chakravarthula2020learned} and camera-in-the-loop calibration~\cite{HoloNet}. 
Although recent learned approaches using differentiable wave propagation models and smooth object phase are capable of producing high-quality images, they typically suffer from low axial resolution with a small eyebox \cite{b-sgd}. Kim et al. \shortcite{kim2022accommodative} demonstrated that smooth phase tends to display less pronounced defocus effects compared to random-phase holograms. However, random phase results in severe speckle noise, e.g., those produced by Gerchberg-Saxton (GS) algorithm ~\cite{gerchberg1972practical}.
This limitation significantly restricts the attainable retinal blur and, as a result, the monocular accommodation cues, presenting a significant drawback for holographic displays~\cite{kim2022accommodative}.
We study the effects of both smooth and random phases on the eyebox and focal cues in \Cref{fig:existing_diffPupil}.
The deviations from real-world incoherent imaging phenomena in a coherent light-based holographic display and the presence of unnatural defocus effects can be clearly seen in both the cases.

Ensuring the faithful delivery of retinal defocus blur is of paramount importance in holographic displays, yet this domain remains inadequately explored. Holographic displays function based on a coherent-light imaging model, in stark contrast to the incoherent light transport model of the physical world. 
This disparity in the light transport models give rise to a discrepancy in the retinal defocus blur between real and (holographically generated) virtual objects, thus compromising the quality of 3D holographic images and impeding the realization of photorealistic hologram rendering.
\rerevise{To address this issue, recent iterative methods \cite{b-sgd, realistic_blur} propose improving out-of-focus blur in holograms by using synthesized focal stacks from RGB-D input. However, these iterative approaches are computationally intensive, particularly the current state-of-the-art method by Lee et al.~\shortcite{b-sgd}, which optimizes multiple frames for time-multiplexing. Alternatively, Yang et al.~\shortcite{deh2022} introduce a neural network-based solution, utilizing rendered focal images as supervision for predicted holograms, yielding promising results. However, this approach assumes a fixed pupil diameter, limiting its adaptability to varying pupil sizes of the user. In reality, pupil size often changes due to factors like ambient lighting, typically ranging from 2-4 mm even in consistent lighting conditions. Considering the user's pupil state is essential for achieving proper accommodation, enhancing depth perception and realism\cite{hennessy1976effect, ward1985effect, bittermann2007blur}. 
Unfortunately, current methods do not incorporate pupillary changes during hologram generation.}

In this work, we propose a framework to tackle the abovementioned issues in hologram generation: 1) bridging the gap between different light transport models to achieve real-world defocus effects caused by incoherent light on a coherent holographic display, and 2) integrating real-time pupillary changes to generate accommodation cues that adapt to the user's pupil size.
Our primary objective is to achieve a balance between perceptual fidelity and computational efficiency, and to this end, we introduce novel learning techniques to overcome the associated challenges. 
Our approach centers on a unified architecture capable of predicting diverse holograms based on varying input pupil sizes. The key to our framework is 
to automatically adjust the neural framework's receptive field, and consequently, adapts the depth-of-field in 3D holographic images according to the pupil size.
To overcome the disparity in light propagation models, we create a dataset of 3D focal stacks, aligning the camera parameters with those of the human eye. Subsequently, our neural network is trained by employing the coherent wave field propagation of the predicted hologram to supervise the reconstruction of focal stacks. Note that the predicted focal stacks are compared to target ground truth focal stacks generated using an incoherent imaging model and corresponding pupil aperture sizes.

Overall, we summarize our contributions as follows:
\begin{itemize}
    \item We analyze the effect of pupil states within the eyebox of a 3D holographic display and show that developing a coherent light-based holographic display that simultaneously achieves a large eyebox, natural defocus cues and high image quality is challenging due to inherent trade-offs.
    \item We develop a single adaptive learning framework that incorporates the effect of dynamic pupillary changes of the eye via an adjustable deformable convolutional neural network for 3D hologram synthesis, thereby allowing for dynamic depth-of-field and defocus changes in generated images.
    \item We present a training approach that leverages photorealistically generated focal stacks with a range of pupil aperture sizes to efficiently guide the training of our framework.
    \item We demonstrate significant improvements featuring convincing defocus effects and diverse depth-of-fields under varying pupil conditions both in simulation and on a functional hardware prototype. Additionally, we offer a comprehensive evaluation and analysis of the efficacy of our proposed framework.
\end{itemize}

We will release all the code and datasets for this paper. 

\paragraph{Scope and Limitations}
This paper investigates the tradeoffs involved in achieving high-quality noise-free 3D images, natural and pupil-dependent focus cues seen in real-world, and a large eyebox-wide light energy distribution, simultaneously. We demonstrate that achieving all the above in a single system is challenging and we propose a method to adapt to pupil diameter via an adjustable deformable convolutional network that dynamically adjusts the receptive field, and generate high fidelity holograms with natural pupil-dependent defocus. Note that pupil-dependent 3D focus cues is heavily dependent on desired image quality and speckle noise, eyebox energy, and image formation models used, and, as we also validate in our paper, a 3D hologram implicitly does not guarantee pupil-dependent realistic focus cues.

Our work does not implement a closed-loop system where the pupil state is measured to drive the rendering. As we validate in the paper, achieving a wide eyebox energy distribution simultaneously with high fidelity 3D image and focus cues is challenging and warrants physically steering the eyebox, which we do not implement in the current work. Also, a metric to accurately measure the perceptual quality of retinal focus cues can further improve our work. We note that human eyes distort the phase of incident light in a way specific to the individual observer \cite{Chakravarthula:2021:GCR}, with the resulting errors compensated by the human brain  \cite{artal2004neural} while using the chromatic aberrations to drive the accommodation \cite{chromablur}. Employing additional hardware and software for incorporating eyebox steering and the above holistic perception of the human visual system into the proposed framework, augmented with perceptual evaluations, is an exciting direction and our planned future work. 

\section{Related Work}
Our work is related to a large body of display and holography research. In this section, we review relevant prior literature that the proposed framework builds upon.

\subsection{Holographic Displays}
Holographic displays are capable of reproducing the entire continuous light field of an underlying scene, and hence are capable of providing appropriate depth and view dependent effects, making them a promising technology to achieve unprecedented capabilities for future AR/VR applications \cite{doublePhase, Kim2022glass}. 
Recent optimization \cite{wirtinger, chakravarthula2022hogel, chakravarthula2022pupil} and deep learning-based methods \cite{shi2021nature, choi2021neural3d, chakravarthula2020learned} moreover have shown compelling image quality showing holographic displays have the potential of approaching the quality of conventional displays. 
Alongside digital holographic displays, there is also a growing interest in utilizing holographic optical elements (HOEs) in near-eye displays to replace conventional optical elements such as refractive lenses and prisms.
For instance, Maimone et al. \cite{doublePhase} demonstrated a design for an augmented reality holographic display, and more recently, a thin and lightweight virtual reality display~\cite{maimone2020holographic} using analog holographic optical elements.
Kim et al. \shortcite{Kim2022glass} designed a compact holographic virtual reality near-eye display system supporting focus cues. 
\rerevise{
While holography can offer ultimate display capabilities, unfortunately, most of the existing computer-generated holography (CGH) display methods support a very small eyebox severely limiting the view- and depth-dependent effects. 
Note that while holographic optical elements recorded via analog holography reduces bulky optical stacks paving way for compact wearable displays, a significant benefit of holographic displays come from CGH methods.
In CGH approaches, the object wavefronts are numerically simulated and encoded into phase patterns that are displayed on a phase-only spatial light modulator (SLM). Therefore, the quality of the computed phase patterns greatly influence the fidelity of perceived visual cues and imagery. In this work, we propose a framework to achieve high-fidelity 3D imagery with appropriate defocus cues based on the current pupil state.}

\subsection{Computer Generated Holography}
In this section, we review existing CGH methods and categorize them into non-learning-based traditional phase retrieval methods and neural network-based learned phase retrieval approaches, with focus on 3D hologram generation.

\begin{figure*}[!t]
\includegraphics[width=0.99\textwidth]{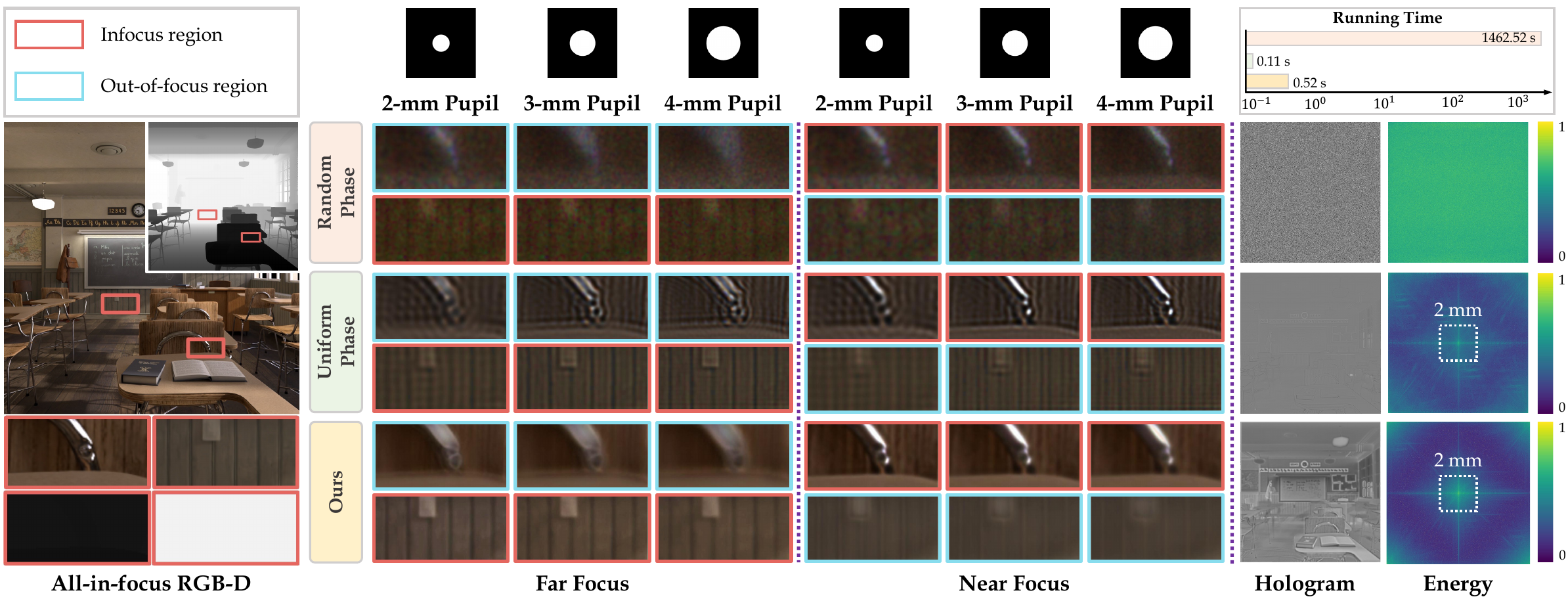}
\vspace{0.1 mm}
\caption{
\emph{Evaluation of smooth phase and random phase holograms in simulation.} 
We simulate holographic reconstructions of smooth and random phase holograms with their respective state-of-the-art methods. For the smooth phase holograms \cite{shi2022light}, the light energy is concentrated at the center of the eyebox and fails to reproduce correct pupil-dependent depth-of-field effects. On the other hand, random phase holograms \cite{b-sgd} achieves uniform energy distribution across the eyebox and correct defocus effects for different pupil sizes, but suffers from severe speckle noise, more notably for smaller pupil sizes. Please see \Cref{sec:motivation} for a detailed discussion.
}
\label{fig:existing_diffPupil}
\end{figure*}
\subsubsection{Traditional Phase Retrieval.} 
\label{sec:classical} 
Conventional phase retrieval methods are employed to generate holograms from various 3D scene representations, including meshes, point clouds, sliced layers, and rendered images. When calculating holograms directly from non-planar scene representations like polygonal meshes and point clouds, a common approach is to treat each element within the scene as an individual emitter \cite{benton2008holographic, Ogihara15}. However, achieving higher display quality using this method requires densely sampled primitives, which in turn necessitates significant computational resources. Despite the presence of numerous computational acceleration techniques, such as GPU parallelization ~\cite{petz2003fast,masuda2006computer,chen2009computer} and use of look-up tables~\cite{Kim08}, it remains a challenge for these methods to accurately reproduce occlusion and defocus effects \cite{chakravarthula2022hogel} because of their treatment of individual primitives.

Image-based scene representations such as focal stacks and light fields offer a more efficient way of signal processing for CGH.
For instance, layer-based techniques \cite{okada2013, Zhao2015} use a stack of intensity layers to represent a 3D scene and propagate these layers toward the spatial light modulator (SLM) plane using wave propagation models.
This approach necessitates dense scene sampling to produce nearly accurate focus cues, but struggles representing occlusions in the scene \cite{Zhang17}.
Recent work Kavakli et al.~\shortcite{realistic_blur} supervises the reconstruction planes using a multi-plane representation, convolving out-of-focus regions at each plane with a Gaussian kernel. However, this multi-plane representation, constructed from an RGB-D pair, has limitations due to inaccessible occluded parts, affecting defocus blur within the target planes.
Lee et al.~\shortcite{b-sgd} and Choi et al.~\shortcite{choi22time} utilize a focal stack as supervision, combined with time-multiplexing strategies, to achieve natural defocus blur, but it requires focal stack or light field data for hologram computation. Notably, both approaches entail iterative optimization for multiple holograms, resulting in significant computational time.
\rerevise{As for 3D hologram synthesis from light field data, it demands interference of wavefronts propagated from each view point within the light field \cite{zhang2015, zhang2016, Shi17, padmanaban2019holographic}. 
While hogel-free holography \cite{chakravarthula2022hogel} addresses some limitations of hogel representation in existing methods, it comes with high computational and time costs.
}

\subsubsection{Learned Phase Retrieval.} \label{sec:neural}
In recent years, there have been notable advancements in leveraging neural networks to enhance both the quality and efficiency of computer-generated holography. 
Eybposh et al.~\shortcite{eybposh2020deepcgh} demonstrated rapid 3D hologram synthesis using a CNN-based framework with well-designed supervision strategies.
\rerevise{
Shi et al.~\shortcite{shi2021nature,shi2022light} illustrated that, with an RGB-D input, a lightweight neural network trained on large-scale photorealistic renderings can generate high-quality 3D holograms.
Choi et al.~\shortcite{choi2021neural3d} proposed a 3D hologram optimization framework with parameterized wave propagation modules implemented by neural networks.
Shui et al.~\shortcite{zheng22unsupervised} introduced a neural framework that supervises predicted phase-only holograms using masked in-focus areas at different focal planes. 
Most recently, Yang et al.~\shortcite{deh2022} employed rendered varifocal images to supervise image reconstruction at various planes for 3D hologram synthesis, achieving natural defocus blur. It is worth noting that although Yang et al.'s approach is relevant to our study, it is limited to achieving defocus blur within a fixed pupil setting, requiring separate training runs for each pupil state variation.
Motivated by the continuous fluctuations in human pupil sizes throughout the day, our aim is to develop a comprehensive and unified 3D hologram synthesis framework capable of generating realistic defocus blur across a range of pupil size conditions.}

\subsection{Adaptive Neural Networks}
Adaptive neural networks, which can tailor their parameters according to specific inputs, have found applications in various domains \cite{shaham2021spatially}.
Notably, Mildenhall et al.~\shortcite{mildenhall2018burst} utilized a network to predict filtering kernel weights for each pixel, achieving impressive image denoising results.
\rerevise{
Dai et al. \shortcite{dai2017deformable} introduced a deformable convolutional layer that dynamically adapts the size and shape of convolution kernels to different spatial positions, proving effective in tasks like object detection and semantic segmentation.}
Note that in deformable convolutional layers, kernel adjustments are solely influenced by the input features or image.
Additionally, hyper-networks, which can also adapt to input conditions, have demonstrated their utility in various tasks, including image processing \cite{fan2018decouple}, deblurring \cite{fan2018decouple}, and super-resolution \cite{hu2020meta}. Hyper-networks are capable of uniformly adapting the receptive field for every position, ensuring that the operation kernel size remains the same across all positions. Deformable convolutions, on the other hand, allow for non-uniform adjustment of the receptive field for each spatial position.
\rerevise{In our work, we introduce the first adaptive framework within the holography domain, with the objective of synthesizing 3D holograms featuring realistic defocus effects under varying pupil sizes.}
Inspired by the model of defocus blur formation, we devise an adjustable deformable convolutional layer that enables dynamic and adaptive kernel adjustments at each position, thereby providing focus cues corresponding to the current pupil size.


\section{Motivation}
\label{sec:motivation}
Recent state-of-the-art computational approaches on holography have begun to demonstrate high-quality photorealistic holographic projections with mitigated artifacts. These holograms which typically utilize a smooth object phase show a high energy concentration in a limited angular spectrum thereby severely restricting the eyebox of the display~\cite{chakravarthula2022pupil}. On the other hand, holograms with uniformly distributed angular spectrum can offer a larger eyebox, but suffers from severe speckle noise in the reconstructed images due to random object phase~\cite{schiffers2023stochastic}. Unfortunately, these two physical phenomena attributed to the exsiting CGH methods significantly limit the support of focus cues, which is one of the biggest advantages of holographic displays~\cite{kim2022accommodative}. Moreover, changes in eye pupil size alters the depth-of-field and hence significantly affects focus cues and depth perception. 

In holographic displays, there exists a tradeoff between eyebox size, image quality, and defocus effects due to the inherent challenges in achieving a balance between these factors.  
Incorporating pupil-dependent depth-of-field effects and real-time pupillary changes further complicates this task, as holograms encode a fixed depth-of-field and cannot adapt to dynamic pupil variations. Additionally, replicating real-world incoherent light-induced defocus cues on a coherent laser-based holographic display is challenging due to the inherent disparities in the light transport models.


We investigate the ability of existing methods to \emph{statically adapt} to pupillary changes to produce pupil-depdendent depth-of-field effects in \Cref{sec:pupil-dependent-investigation}. We then investigate recently proposed pupil-aware hologram optimization approach \cite{chakravarthula2022pupil,schiffers2023stochastic} to simultaneously achieve image quality and defocus effects within a large eyebox using light field data to explore the inherent tradeoffs in \Cref{sec:pupil-aware-strategy}. The observations in this section motivate us to devise the pupil-adaptive holography approach to achieve pupil-dependent depth-of-field effects mimicing incoherent light, which we later describe in \Cref{sec:approach}. 
 

\subsection{Static Adaptation to Pupillary Changes}
\label{sec:pupil-dependent-investigation}

We assess the ability of existing solutions to statically adapt to pupil-dependent defocus effects by simulating holographic images generated with recent neural network-based and iterative method \cite{shi2021nature, b-sgd} using varying pupil sizes. The results presented in \Cref{fig:existing_diffPupil} demonstrate the tradeoff discussed previously in existing hologram approaches. 
The neural-network-based approach by Shi et al.~\shortcite{shi2021nature, shi2022light} produces holograms with smooth phase patterns and good image quality, but exhibit erroneous defocus effects, especially showing unrealistic and nearly constant depth-of-fields across different pupil sizes.
In contrast, the iterative method by Lee et al.~\shortcite{b-sgd} creates time-multiplexed holograms with random phases featuring a uniformly spread eyebox energy distribution, offering more reasonable defocus trends. That is, larger the pupil size, greater the blur effect in out-of-focus areas.
However, they suffer from pronounced speckle noise, especially with smaller pupils.
These findings emphasize the challenge of achieving noise-free holographic projections with appropriate depth-of-field variations by statically adapting to different pupil sizes, even when using time-multiplexing strategies.

\subsection{Pupil-aware Light Field Holography}\label{sec:pupil-aware-strategy}

Recently Chakravarthula et al.~\shortcite{chakravarthula2022pupil} and Schiffers et al.~\cite{schiffers2023stochastic} suggested that employing a pupil-aware optimization can result in energy distributed across a larger eyebox. However, neither methods were able to demonstrate noise-free imagery and good defocus. We explore the feasibility of pupil-aware optimization with light field as input to statically adapt to pupil size variations and simultaneously achieve image quality, defocus cues and a wide eyebox. We initialize the optimizer with a random phase-only hologram and supervise the reconstructions over sampled pupils of size 3~mm or 4~mm and focal distances. It's important to highlight that the energy distribution of the hologram is directly influenced by the selection of the pupil coverage area during the optimization process. 
This impact arises because the supervision process corresponds with the light field views associated with sampled pupil positions. 
Consequently, we explore the impact of the pupil sampling interval $z$ (see \Cref{fig:pupil_array}) on the hologram's energy distribution and how this influences image quality. For more comprehensive illustrations and additional details, please see the Supplementary Material.

\setcounter{figure}{3}
\setlength{\columnsep}{8pt}
\begin{wrapfigure}{r}{0.42 \linewidth}
\centering
\vspace{-12px}
\includegraphics[width=.86\linewidth]{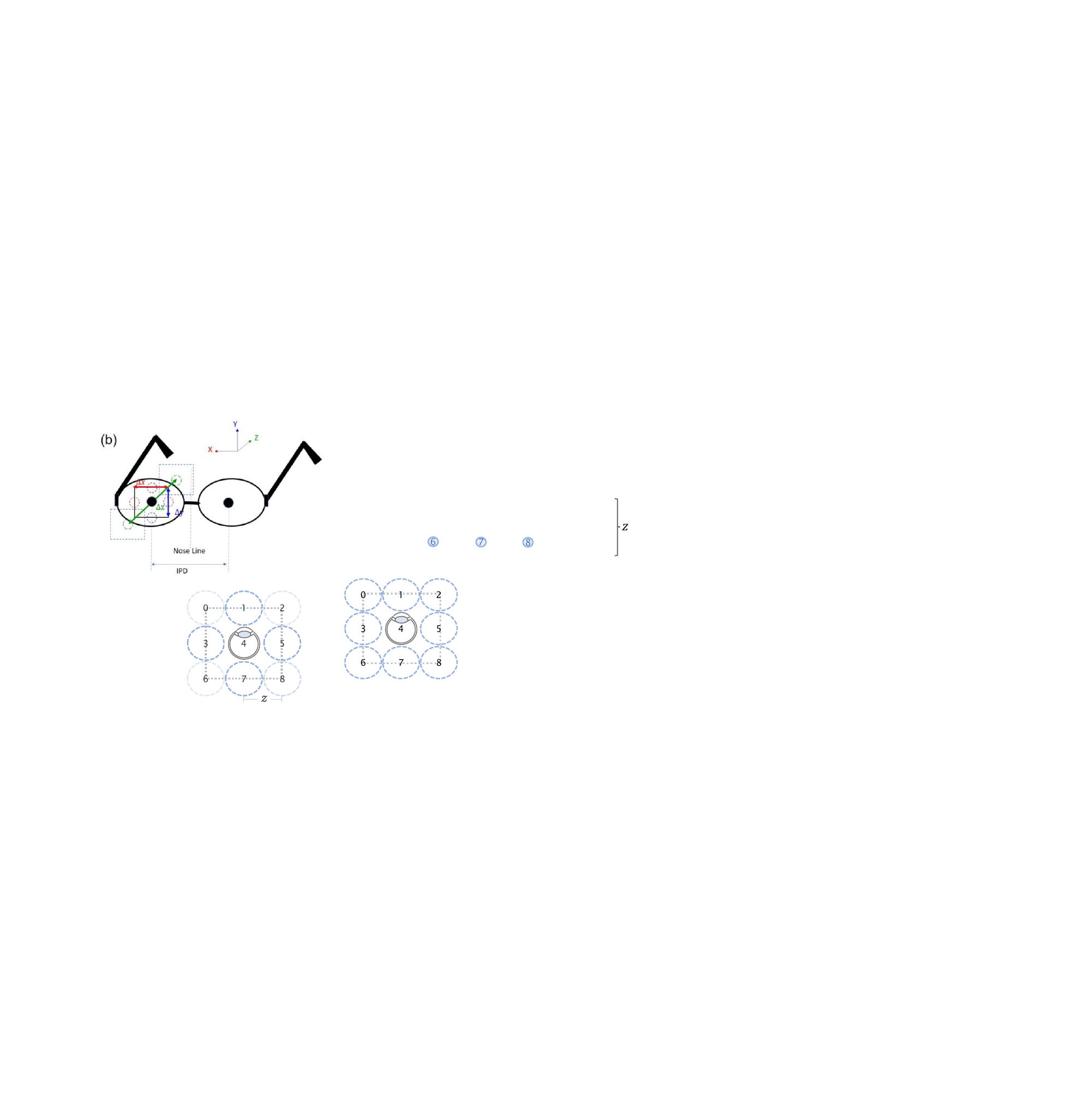}
\caption{\label{fig:pupil_array} Eyebox pupil sampling.}
\vspace{-12px}
\end{wrapfigure}

\paragraph{Observation} The pupil-aware optimization approach yields promising results, shown in \Cref{fig:LF_focus}, when the distance $z$ between adjacent pupil positions in a $3\times3$ pupil sampling array remains within $1$ mm range, for pupil sizes of $\{3\:\text{mm}, 4\:\text{mm}\}$. 
\setcounter{figure}{2}
\begin{figure}[!t]
\includegraphics[width=0.48\textwidth]{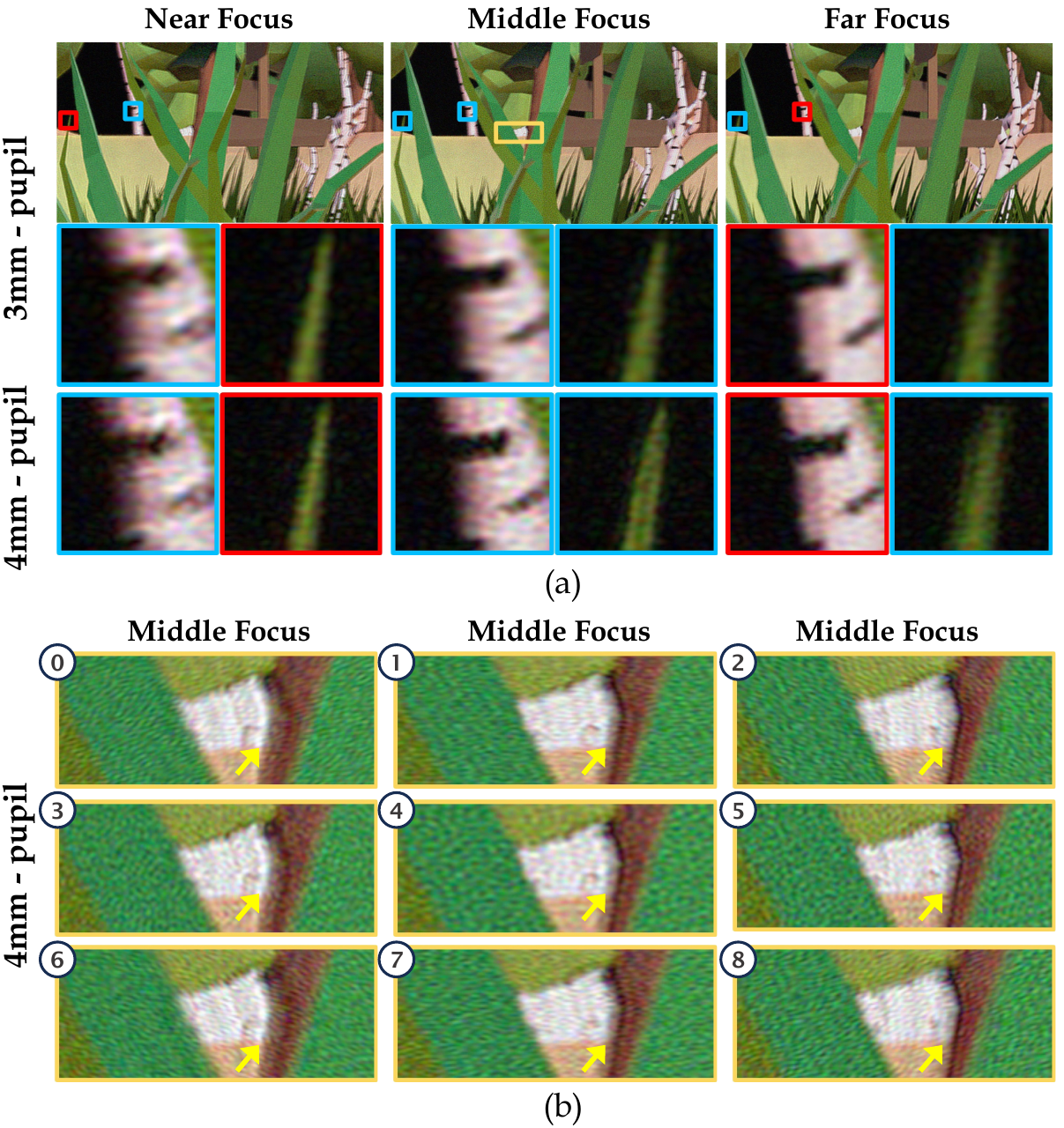}
\caption{Simulated reconstruction from the optimized hologram. (a) Reconstructed images at three focal planes for a 3mm-puil and 4-pupil at the center view. In-focus regions are highlighted in red and out-of-focus regions are in blue.
(b) Illustration of the parallax effects. Insets are taken from reconstructions at middle focal plane at $9$ viewpoints with a 4mm-pupil.
}
\label{fig:LF_focus}
\end{figure}
\setcounter{figure}{4}
\begin{figure*}[!htbp]
\includegraphics[width=18cm]{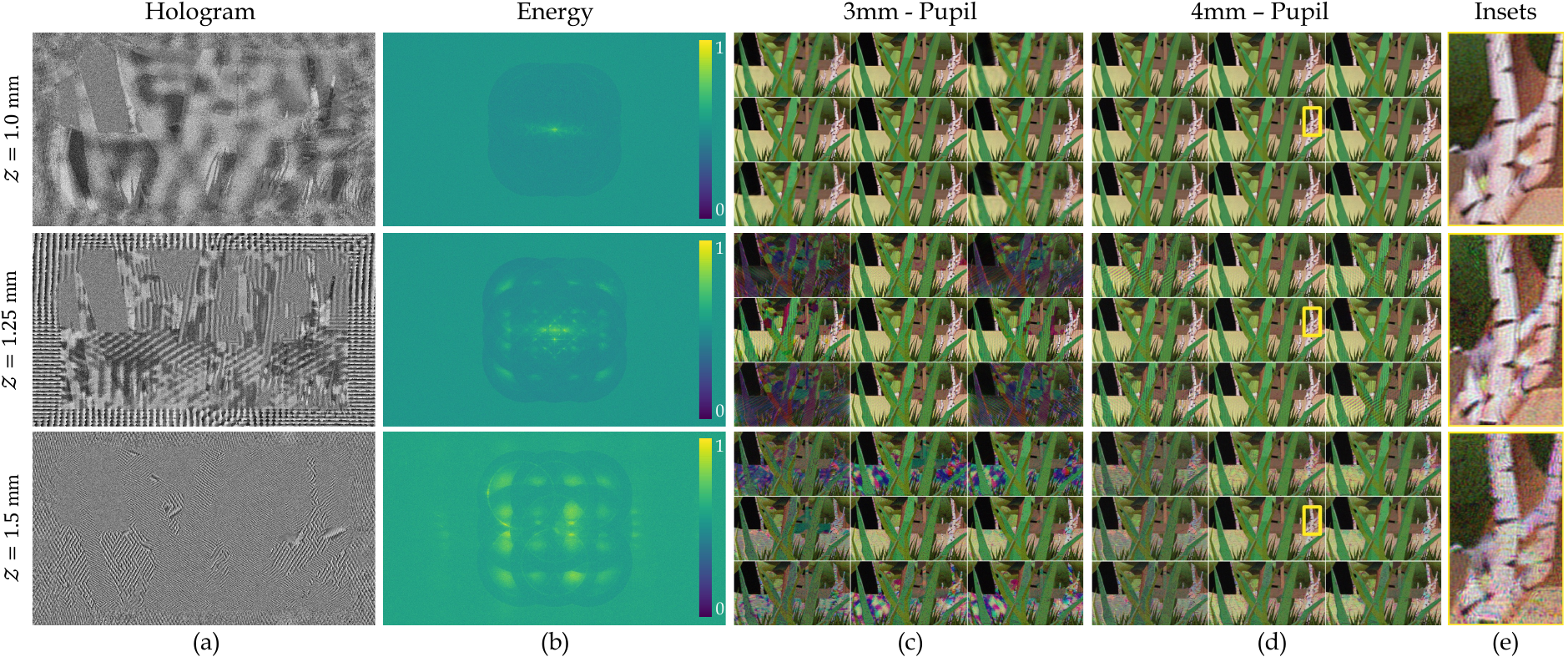}
\caption{Pupil-aware light field holography with different pupil sampling intervals. We show the optimized holograms (green channel) in column (a) and corresponding energy distributions in (b). A full set of the reconstructed views (at middle depth plane) for a 3-mm pupil are shown in (c) and for a 4-mm pupil in (d). The insets in (e) highlights an enlarged region of the reconstructed center view for a 4-mm pupil. As the pupil interval $z$ increases, the phase pattern exhibits increased randomness and a broad eyebox energy distribution. However, this results in a gradual decline in the quality of reconstructed images.}
\label{fig:LF_interval}
\end{figure*}
\Cref{fig:LF_focus}(a) presents simulated reconstructions from the center view of the light field using both 3-mm and 4-mm pupils, demonstrating pupil-dependent focus cues and variations in defocus effects with change in pupil sizes.
\Cref{fig:LF_focus}(b) emphasizes the parallax effects at the middle focus plane across all the nine light field views obtained from the pupil sampling positions. 
It can be seen that the occlusion extent between foreground and background objects varies as the viewpoints change, for instance, the occlusion around the plant as highlighted by yellow arrows.
However, note that the reconstructed images are corrupted with speckle noise, an observation also recently validated by Schiffers et al. \shortcite{schiffers2023stochastic}

In \Cref{fig:LF_interval}, we show the energy distribution of optimized holograms with varying intervals between adjacent sampled pupil positions. A larger interval between sampled pupil positions leads to a more dispersed energy distribution, contributing to an expanded eyebox. However, this expansion comes at a cost of compromised image quality and parallax. 
For instance, when the interval $z$ is increased to 1.25 mm, while reconstructions at various views remain reasonable for the 4-mm pupil, significant reconstruction errors appear for a 3~mm pupil, particularly at corner views (positions 0, 2, 6, 8). However, severe image artifacts are observed in the reconstructions for a 4~mm pupil when the $z$ interval is increased to $1.5$~mm as highlighted in \Cref{fig:LF_interval}(e). For more discussion on the effect of sampling interval $z$ of the eyebox, please refer to the Supplementary Material.

\paragraph{Discussion}
Our investigation reveals a trade-off between the quality of image reconstructions, the quality of defocus effects, and the energy distribution in holograms computed from light fields: \\
\noindent
\textbf{Tradeoffs with large eyebox:}
Optimizing for a large eyebox can reduce the image quality as the holographic content is spread across a wider viewing area, diminishing signal-to-noise ratio, spatial resolution and object clarity. 
With degradation in image quality, distinguishing between defocus effects near in-focus and out-of-focus regions may also come challenging.
Moreover, a larger eyebox necessitates uniform angular distribution of light, limiting the system's ability to provide strong and convincing defocus cues. \\
\noindent
\textbf{Tradeoffs with image quality:} Conversely, to achieve the best possible image quality, holographic displays concentrate the holographic content within a narrower viewing area with much of the light interfering in-phase to maintain sharpness and clarity, leading to a smaller eyebox. \\
\noindent
\textbf{Tradeoffs with defocus effects:} 
Akin to light field displays \cite{lanman2013near}, achieving defocus effects often involves a controlled distribution of light angles, which reduces the eyebox size by restricting viewing positions. 
On the other hand, as discussed in \Cref{sec:pupil-dependent-investigation} and shown in \Cref{fig:existing_diffPupil}, adapting to arbitrary pupil sizes and positions necessitates a uniformly spread eyebox energy and a wide eyebox \cite{chakravarthula2022pupil}. 
Moreover, this also brings a conflict with ensuring high image quality which often constrains energy distribution to a narrow effective eyebox.



%
While solutions combining random phases and time-multiplexing techniques may partially alleviate the image quality-eyebox energy trade-off, they sacrifice time resolution and demand the computation of multiple holograms, further increasing computational and time costs. To address these challenges and achieve fast hologram generation that adapts to pupil size variations for improved 3D image quality, we propose a \emph{dynamic modification of the holographic wavefield's depth-of-field properties in a pupil-adaptive manner}, which we will discuss in the following sections.

\begin{figure}[t!]
\includegraphics[width=0.45\textwidth]{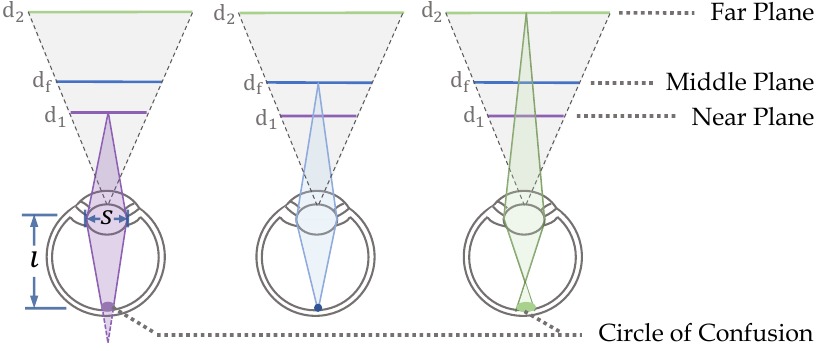}
\caption{When a human eye is focusing at a distance (e.g., the middle plane), an object point appearing away from the focal plane will occur dilated (within an area) on the retina rather than as a clear point. The area on the retina is called the circle of confusion (CoC) and its diameter is proportional to the distance between the object point and the focal plane.}
\label{fig:coc}
\end{figure}
\section{Pupil-adaptive 3D holography} \label{sec:approach}

To enable pupil-dependent depth-of-field effects in holography, we have developed a unified pupil-adaptive framework for 3D hologram generation that incorporates pupil size as a key parameter. Our framework takes an RGB-D image and the current pupil size as input, and integrates the spatially-varying properties of defocus blur and the relationship between pupil size and the circle-of-confusion (or simply blur circle) size at each spatial position. Central to our framework is the use of an adjustable deformable convolution layer to change the receptive field of the network on demand. We trained our framework using photo-realistically rendered focal stacks of randomly generated scenes with camera parameters matching the aperture and depth-of-field of the eye under various pupil conditions, allowing us to synthesize 3D holograms under various pupil-size conditions and mimicking incoherent defocus cues.

In \Cref{sec:prerequisites}, we provide a concise overview of the correlation between pupil diameter and blur size, as well as the connection between the reconstruction planes of the 3D hologram and the displayed focal distances. Following this, we offer an in-depth exploration of our framework's design, presenting the architecture in \Cref{sec:arch} and outlining the training objectives in \Cref{sec:objecive}. In \Cref{sec:extension}, we discuss a variant of our framework that receives evenly sampled focal images as input.


\subsection{Pupil Effects on Displayed Imagery} \label{sec:prerequisites}

\subsubsection{Pupil-blur Relationship} \label{sec:blur_size}
As shown in \Cref{fig:coc}, according to a thin lens imaging model, an object point contributes to an area on the imaging sensor with a diameter proportional to the distance between it and the focal plane. 
This area is called Circle of Confusion (CoC) and its diameter can be derived from the depth $d$ of the object point, focal distance $d_f$, the aperture size of the lens $s$, and the distance $l$ between the lens and the sensor, which is given by
\begin{equation} \label{eq:coc}
c = s\cdot l \cdot \big|\frac{1}{d} - \frac{1}{d_f}\big|.
\end{equation}
Similarly, we can model human eyes using a thin-lens approximation, enabling the computation of observed defocus effects through \Cref{eq:coc}. Since objects in a scene are distributed across various depth planes, the sizes of the contributing CoC areas vary spatially and shift with alterations in pupil size, as indicated by \Cref{eq:coc}.

\begin{figure}[htbp!]
\includegraphics[width=0.47\textwidth]{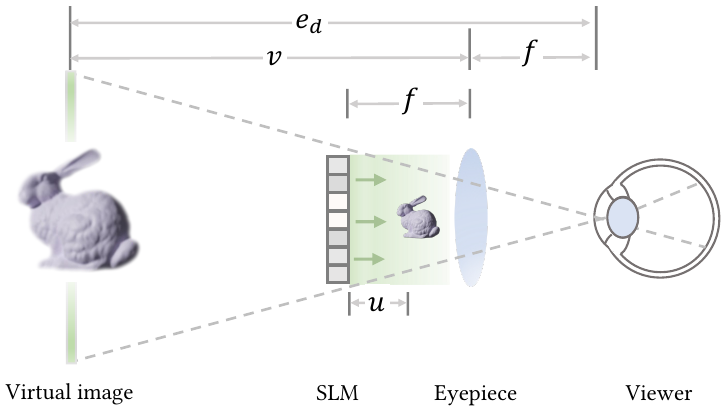}
\caption{Illustration of the display optical system. The spatial light modulator (SLM) is placed close to the focal plane of the eyepiece. The reconstructed image formed at a distance $u$ from the SLM plane forms a magnified virtual image at a distance $e_d$ from the viewer, whose eyebox forms at the focal plane of the eyepiece.}
\label{fig:eye_distance}
\end{figure}
\begin{figure*}[!t]
\includegraphics[width=0.98\textwidth]{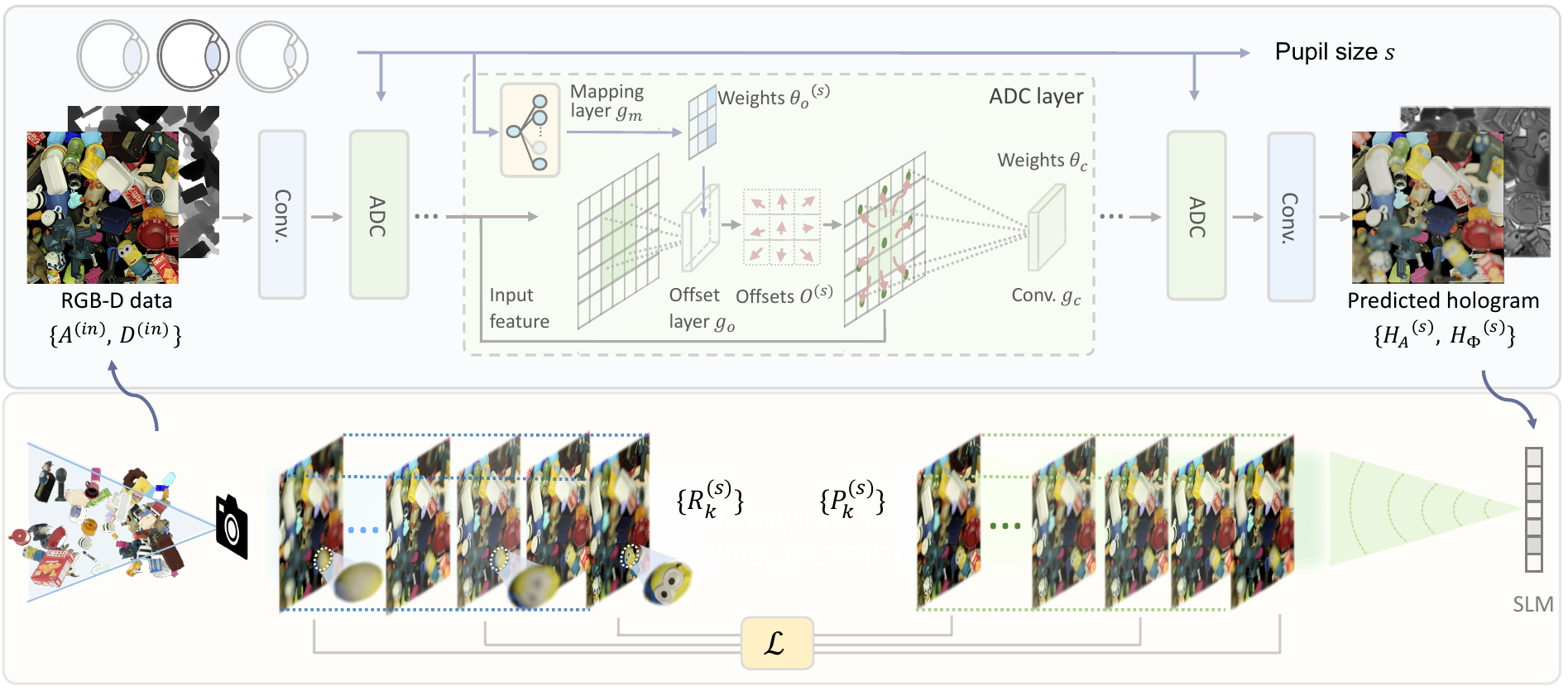}
\vspace{0.1 mm}
\caption{Pupil-adaptive 3D holography framework. Given the input RGB-D data and a specified pupil size, our proposed framework synthesizes a 3D hologram capable of producing defocus effects according to the current pupil size. An adjustable deformable convolutional layer incorporates the current pupil information by dynamically adjusting the neighbors within the convolution kernel of each spatial position according to the pupil size and feature map. To supervise the training, we use path tracing to render focal stack images with defocus cues from randomly generated scenes by setting a virtual camera with parameters matched to that of an eye. This supervision strategy enables achieving defocus effects of incoherent light on a coherent light-based holographic display.
}
\label{fig:pipeline}
\end{figure*}
\subsubsection{Display Model}

As depicted in the simplified illustration in \Cref{fig:eye_distance}, a phase-only hologram pattern displayed on an SLM modulates the incoming wavefront, and the reconstructed images are magnified with an eyepiece. Specifically, as shown in \Cref{fig:eye_distance}, if a wave modulated by the SLM placed close to the eyepiece focal plane projects an image at a distance $u$, the distance $v$ to the magnified virtual image formed by the eyepiece lens can be calculated as
%
\begin{equation}
    v = \frac{f(f-u)}{u},
\end{equation}
where $f$ denotes the focal length of the eyepiece.
And the eyebox is typically formed at the focal plane of the eyepiece, and hence 
the distance between the viewer (camera in the rendering) and the virtual image (focal stack image in rendering) is 
\begin{equation}\label{eq:virtual_distance}
    e_d = \frac{f^2}{u}.
\end{equation}
Given the least distance of distinct vision is $25$~cm, it is essential that $e_d$ is greater than $25$ cm. Therefore, we set the minimum distance between the virtual image and the viewer to $35$ cm. We employed \Cref{eq:virtual_distance} to generate defocus images at the required distances, which enabled the creation of our synthetic training dataset.

\subsection{Architecture}\label{sec:arch}
Our work centers on a unified learning framework that synthesizes holograms according to the input pupil size, removing the necessity for training distinct networks for each pupil condition.
As depicted in \Cref{fig:pipeline}, the input to the network is an all-in-focus RGB-D image pair, denoted as ${A^{(\text{in})}, D^{(\text{in})}}$, and a pupil size $s$.
Subsequently, the system generates the corresponding complex 3D hologram ${H_A^{(s)}, H_{\Phi}^{(s)}}$ while considering the pupil size $s$.
The blur at any given focal plane is spatially varying as discussed in \Cref{sec:blur_size}. Therefore, we account for the maximum blur size required by the synthesized hologram to accommodate various pupil conditions for each position. 
For example, pixels corresponding to objects at the closest or farthest distances necessitate larger blur sizes compared to those at intermediate depths. 
The receptive field of a neural network influences the degree of blur in reconstructed images. An exhaustive empirical study demonstrating this can be found in Section S4 of the Supplementary Material.
Therefore, the framework's receptive field must adapt to these depth cues and also be flexible to accommodate further changes in blur due to variations in pupil size.

As illustrated in \Cref{fig:pipeline}, the proposed framework incorporates a custom adjustable deformable convolutional (ADC) layer. This layer exhibits the capability to dynamically fine-tune the receptive field individually for each pixel position, guided by the pupil diameter.
In particular, drawing inspiration from deformable convolution as introduced by Dai et al.~\shortcite{dai2017deformable}, our ADC layer introduces dynamic offsets for each location within a $3\times3$ kernel. These offset fields are learned during the network training by an additional convolutional layer $g_o$. Moreover, to accommodate variations in the pupil size $s$, we parameterize $g_o$ with learned adjustable weights ${\theta_o}^{(s)}$ produced from a mapping MLP layer $g_m$ that takes pupil size as input. 
Consequently, the ADC layer successfully attains distinct receptive fields for each specific pixel position and pupil size.
Within the ADC block, we employ a convolutional layer $g_c$ that operates on the modified neighborhood grids, where each grid is adjusted based on the learned offsets. These adjustments often result in fractional locations, and to extract features from these locations, bilinear interpolation is applied. 
%
The entire workflow of our custom ADC layer is summarized in Algorithm \ref{alg:adc_layer} and also illustrated in \Cref{fig:pipeline}.

Built upon this design, our framework is able to predict a complex hologram from the input RGB-D data for a given pupil size. This process is formulated as
\begin{equation}\label{eq:model}
H_A^{(s)}, H_{\Phi}^{(s)} = G \left(A^{(in)}, D^{(in)}; s\right).
\end{equation}
where G is our framework, i.e., the hologram formation model.
We then encode the complex field into a phase-only hologram using an anti-aliasing double phase encoding approach \cite{shi2021nature}.


\subsection{Network Training and Supervision Strategy}\label{sec:objecive}
As depicted in \Cref{fig:pipeline}, we jointly train our framework across various pupil-size conditions, guided by rendered focal stacks with the appropriate aperture size.
To be more specific, for a predicted hologram $H_A^{(s)}, H_{\Phi}^{(s)}$, our supervision process involves comparing the reconstructed images at various depth planes with the images present in the target focal stack.
The reconstructed images at different depth planes are obtained via free-space propagation as
\begin{equation} \label{eq:holo_recon}
P_{k}^{(s)} = \Big|f_p\left (\left \{H_{A}^{(s)}, H_{\Phi}^{(s)} \right \}, u_k \right)\Big|^2, \text{for} \:\: k = 1,2,...,\:M,
\end{equation}
where  $M$ is the number of sampled depth planes, $k$ is the index running over the sampled planes and $P_k^{(s)}$ denotes the reconstructed image at $k^{th}$ distance $u^{(k)}$. $f_p$ denotes free-space propagation using the Angular Spectrum Method (ASM) \cite{goodman2005} given by 
\begin{equation}\label{eq:fprop}
\small
\begin{split}
f_{p}(\{A, \Phi\}, d) &=\iint \mathcal{F}(Ae^{j\Phi})(u_{\xi}, u_{\eta}) \mathbb{H}(u_{\xi}, u_{\eta}) e^{j2\pi(u_{\xi} \xi + u_{\eta} \eta )} d{u_{\xi}}d{u_{\eta}},\\
\mathbb{H}(u_{\xi}, u_{\eta}) &= \begin{cases}
&e^{j2\pi d \sqrt{  \frac{1}{\lambda^2} - u_{\xi}^2 - u_{\eta}^2}}, \qquad \text{if} \quad \sqrt{u_{\xi}^2 + u_{\eta}^2} <\frac{1}{\lambda},  \\
&0,  \qquad \qquad \qquad \qquad \;\ \text{otherwise.}
\end{cases}
\end{split}
\end{equation}
where $\lambda$ denotes the wavelength, $\{A, \Phi\}$ denotes a wave field with amplitude $A$ and phase $\Phi$ , $\mathcal{F}$ is the Fourier transform, $u_{\xi}, u_{\eta}$ are spatial frequencies and $d$ is the propagation distance.
Each reconstructed image plane $P_{k}^{(s)}$ is compared against the corresponding focal image $R_k^{(s)}$ rendered at a focal distance $e_d^{(k)}$. The loss for the training objective is calculated as
\begin{equation}\label{eq:loss_all}
    {\mathcal{L}} = \alpha_\text{rec} {\mathcal{L}}_\text{rec} + \alpha_\text{perp} {\mathcal{L}}_\text{perp},
\end{equation}
where  ${\alpha}_{\text{rec}}$, ${\alpha}_{\text{perp}}$ are weight parameters. ${\mathcal{L}}_\text{rec}$ denotes the pixel-wise loss measuring the difference in image intensity and gradient, which are given by
\begin{equation}\label{eq:loss_recon}
{\mathcal{L}}_\text{rec} = \frac{1}{M}\sum_{k=1}^{M} W_k^{(\text{coc})} \cdot \bigg |P_k^{(s)} -  R_k^{(s)} \bigg | +  \bigg | \nabla P_k^{(s)} - \nabla R_k^{(s)} \bigg |.
\end{equation}
$W_k^{(\text{coc})}$ is a weight map calculated from the CoC map $C_k$ at the focal distance $e_d^{(k)}$ 
to emphasize the in-focus regions of the reconstructed image, and is computed as
\begin{equation}
   W_k^{(\text{coc})} = \frac{1}{\text{log}_{2}(1 + C_k)}.
\end{equation}
 $C_k$ stores the blur size for each position and is calculated by \Cref{eq:coc}. 
As such, larger the CoC of a position, the smaller the corresponding weight.
\begin{algorithm}[t!]
    \SetAlgoLined
    \KwIn{Feature map $X$, pupil size $s$}
    \KwOut{Output feature map $V$}
     Produce modulation weight $\theta_o^{(s)}$: \: $\theta_o^{(s)} = g_m(s)$\;
     Inject $\theta_o^{(s)}$ into $g_o$: \: $g_o \gets \theta_o^{(s)}$\;
     Predict offset field $O^{(s)}\in\mathbb{R}^{U \times V \times 18}$: \: $O^{(s)} = g_o(X;\theta_o^{(s)})$\;
    \For{position $(x_i, y_i)$ within $X$}
    {
        Define a regular patch: $\mathbb{P} =\left \{(\delta x, \delta y) \:\: |\:\: \delta x, \delta y \in \{-1, 0, 1\} \right \}$ \;
        \For{location p within the patch $\mathbb{P}$}{
            Augment the location with: $\mathbb{P}_p^{(s)} = \left \{\left (\delta x_p + O^{(s)}\left(x_i, y_i, 2p\right)\:, \:\delta y_p +  O^{(s)}(x_i, y_i, 2p+1) \right) \right \}$ \;
        }
        Calculate the output using $g_c$ (with kernel weight $\theta_c$): \\ 
        $\:\:\:V(x_i, y_i) = \sum_{(\delta x,\delta y)\in \mathbb{P}^{(s)}} \mathbf{\theta_c}(\delta x, \delta y) \cdot {X} (x_i + \delta x, y_i + \delta y)$ \; 
    }
\caption{Adjustable Deformable Convolutional (ADC) layer}
\label{alg:adc_layer}
\end{algorithm}

${\mathcal{L}}_\text{perp}$ in \Cref{eq:loss_all} represents the perceptual loss \cite{LPIPS}, a well-established metric for enhancing textural intricacy. This loss quantifies disparities found in the multi-scale neural features extracted from the reconstruction output and the target. Notably, utilizing ${\mathcal{L}}_{perp}$ on numerous reconstruction images within a focal stack can lead to substantial memory consumption. As a result, we choose to address this concern by randomly selecting a subset $\mathcal{N}$ of the reconstruction planes for computing ${\mathcal{L}}_\text{perp}$ as
%
 \begin{equation}\label{eq:loss_perp}
{\mathcal{L}}_\text{perp} = \frac{1}{|\mathcal{N}|}\sum_{k \in \mathcal{N}}{\mathcal{L}}_{\text{perp}}\left(P_k^{(s)}, R_k^{(s)}\right).
\end{equation}

\revise{
\subsection{Extension to Focal Stack Input} \label{sec:extension}
}
\begin{figure}
\begin{center}
\includegraphics[width=0.47\textwidth]{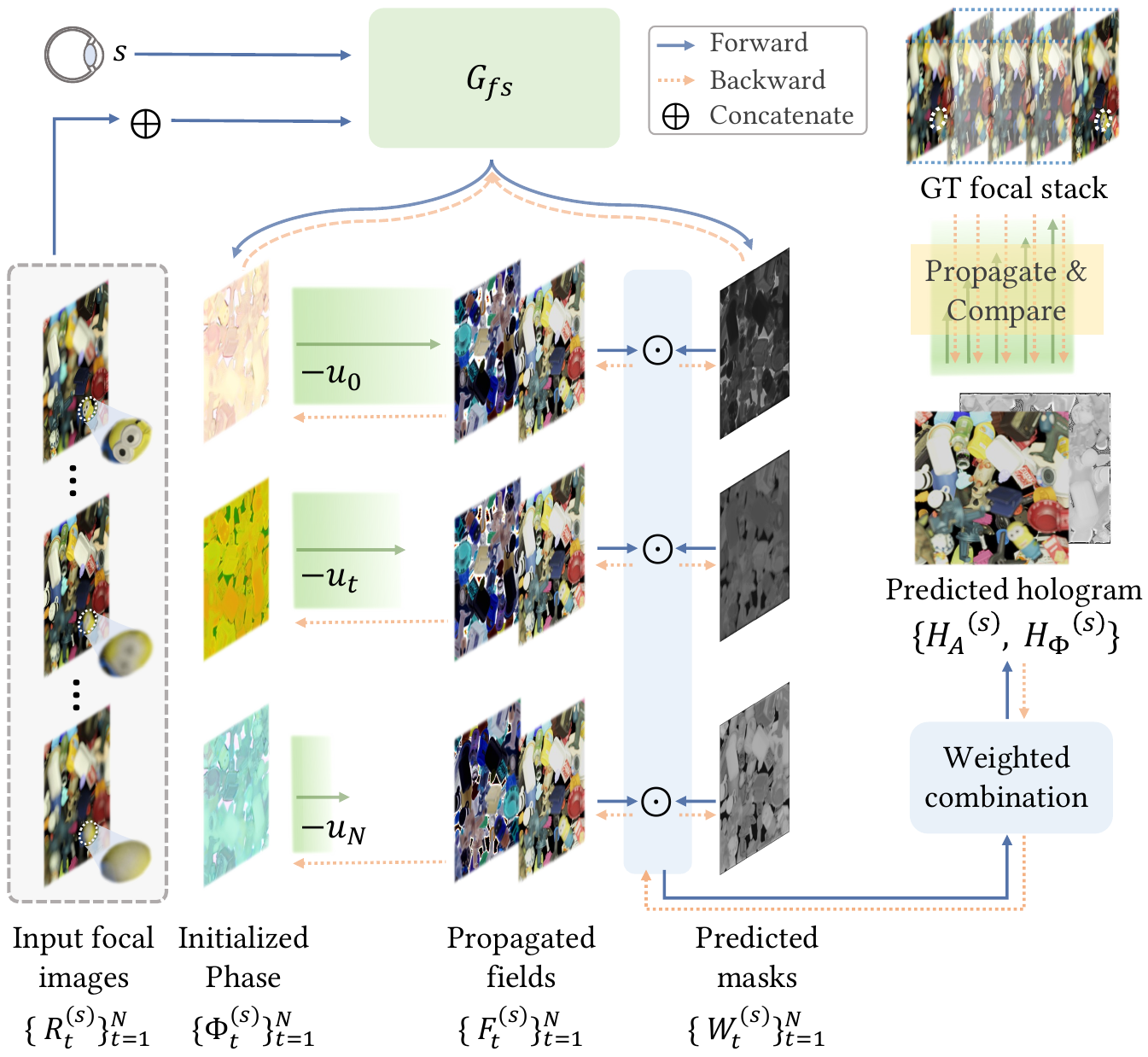}
\end{center}
\vspace{0.6 mm}
\caption{Illustration of the proposed alternative that synthesizes a 3D hologram from equally-spaced focal images. $N$ focal images are concatenated and fed into a neural network $G_\text{fs}$, which predicts an initial phase for each plane. Besides, $G_\text{fs}$ also predicts a mask for each plane that will be used to composite the final hologram. After obtaining the initial phases, the wavefield at each plane is propagated back to the SLM plane. Finally, the complex hologram is composited via a weighted combination of the propagated fields, from which a phase-only hologram is encoded.}
\label{fig:alternative}
\end{figure}
\setlength{\tabcolsep}{1.5pt}
\renewcommand{\arraystretch}{0.2}
\begin{figure*}[htbp]
\begin{center}
\small
\begin{tabular}{cccccc}
& \textbf{Ours (2mm Pupil)} & \textbf{Ours (3mm Pupil)} & \textbf{Ours (4mm Pupil)} & \textbf{Shi et al. 2021}  & \textbf{Shi et al. 2022} 
 \\[0.5ex]
\raisebox{1.85\height}{\rotatebox{90}{\textbf{Far Focus}}}
&\includegraphics[width=3.35cm]{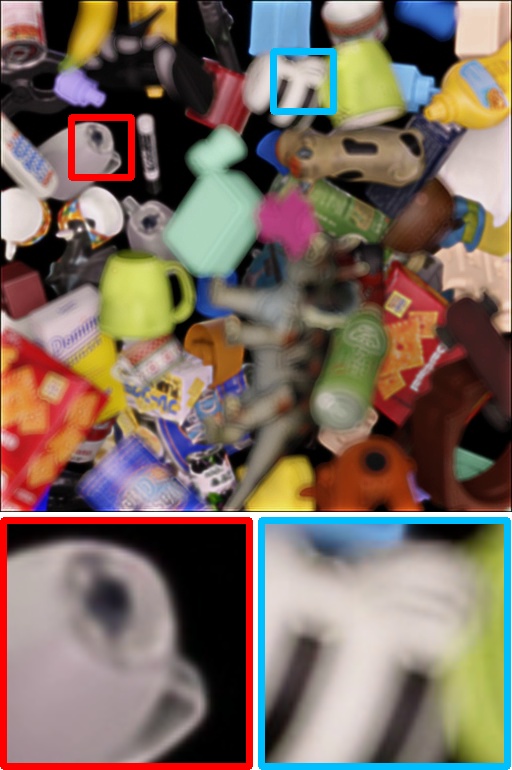}
&\includegraphics[width=3.35cm]{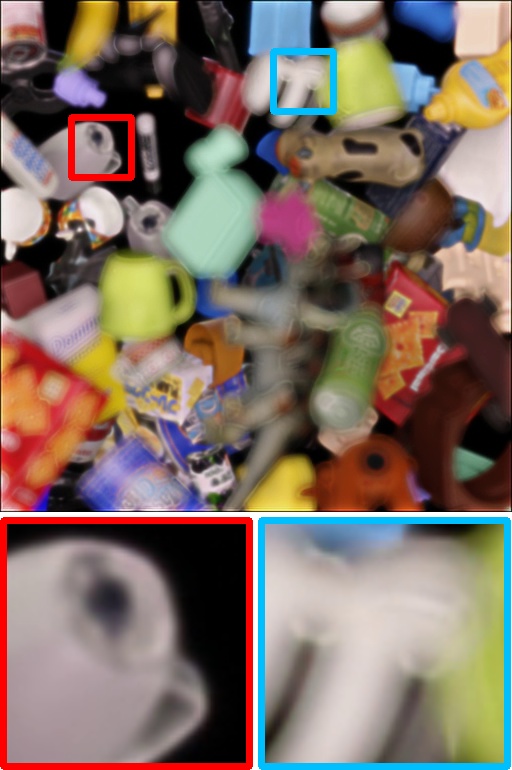}
&\includegraphics[width=3.35cm]{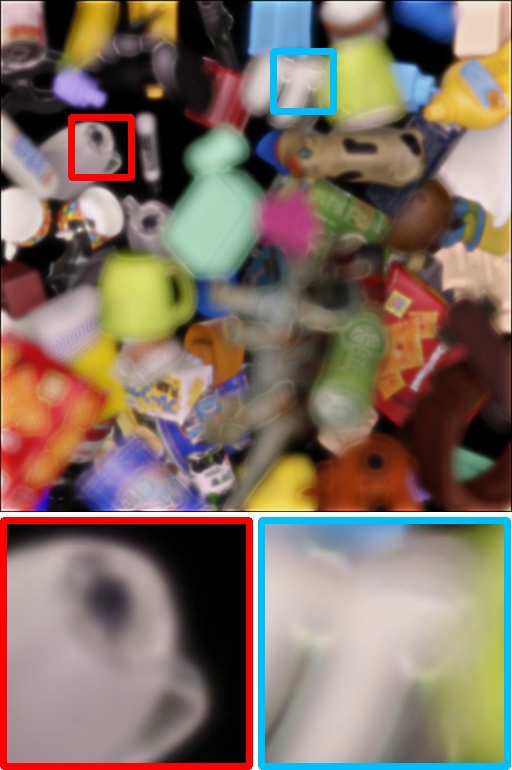} 
&\includegraphics[width=3.35cm]{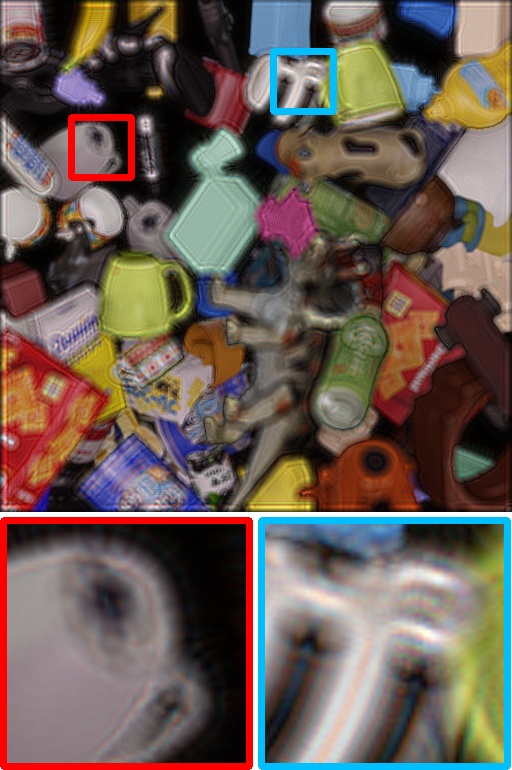} 
&\includegraphics[width=3.35cm]{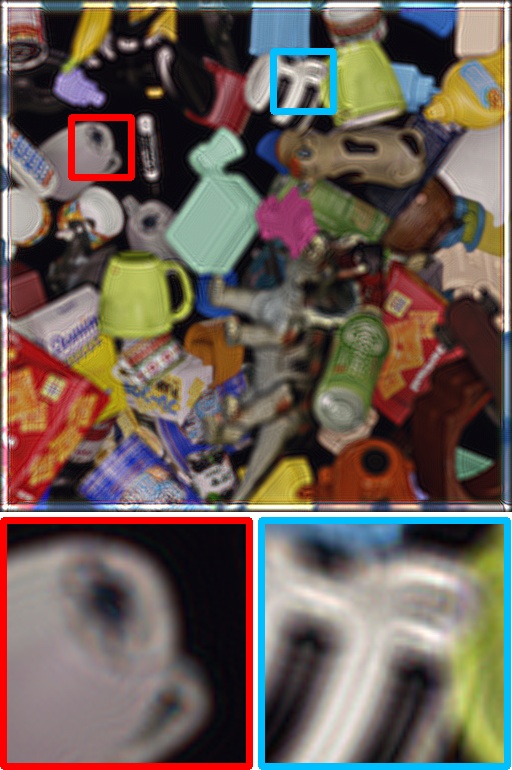}
\\
\raisebox{0.03\height}{\rotatebox{90}{\textbf{Near Focus}}}
&\includegraphics[width=3.35cm]{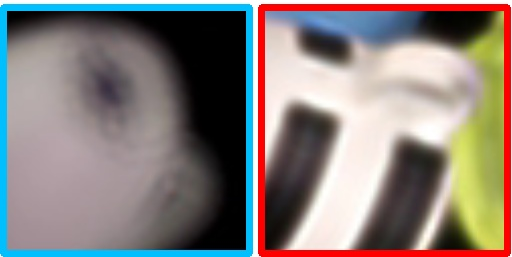}
&\includegraphics[width=3.35cm]{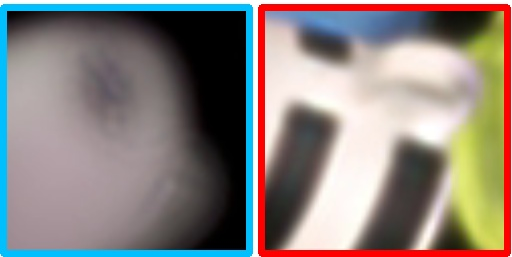}
&\includegraphics[width=3.35cm]{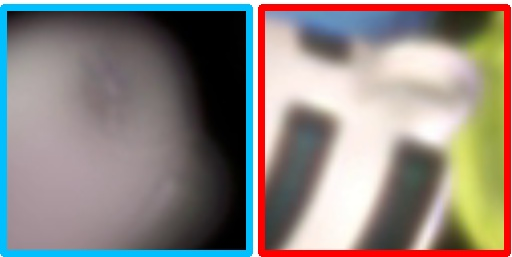} 
&\includegraphics[width=3.35cm]{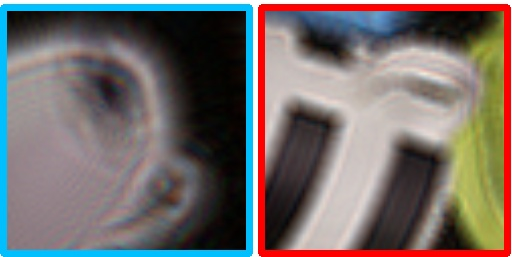} 
&\:\includegraphics[width=3.35cm]{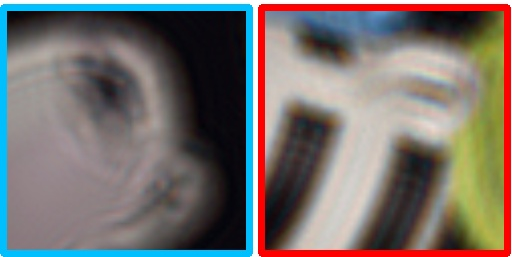}
\\[0.7ex]
&\textbf{Rendered (2mm Pupil)} &\textbf{Rendered (3mm Pupil)} & \textbf{Rendered (4mm Pupil)}  & \textbf{Choi et al. 2021} & \textbf{All-in-Focus RGB-D}
\\[0.7ex]
 \raisebox{1.85\height}{\rotatebox{90}{\textbf{Far Focus}}}
&\includegraphics[width=3.35cm]{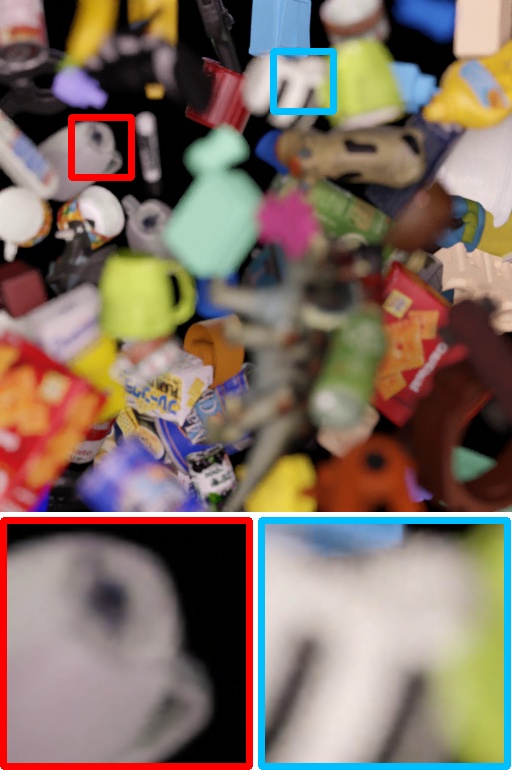} 
&\includegraphics[width=3.35cm]{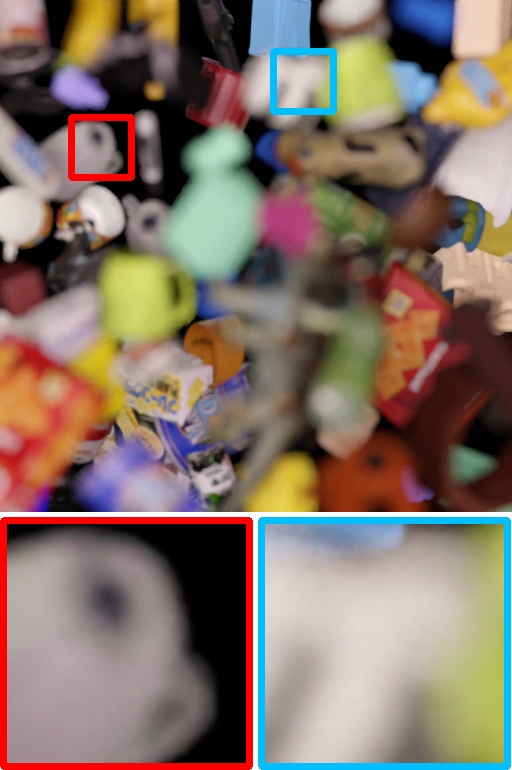} 
&\includegraphics[width=3.35cm]{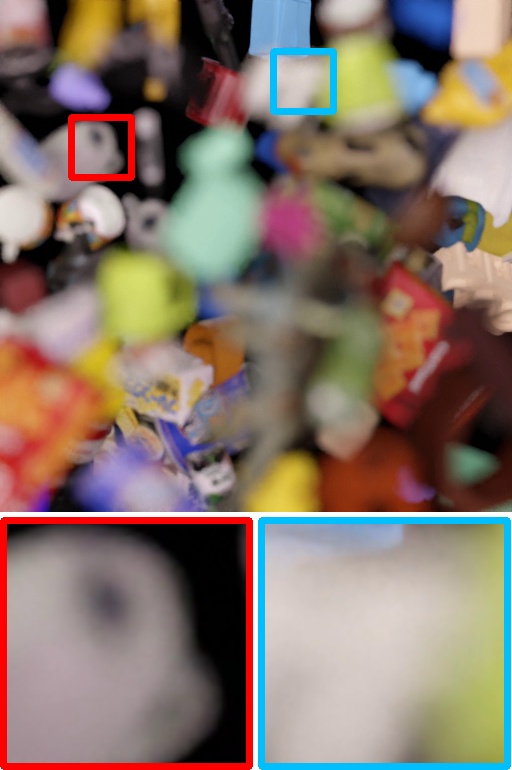} 
&\includegraphics[width=3.35cm]{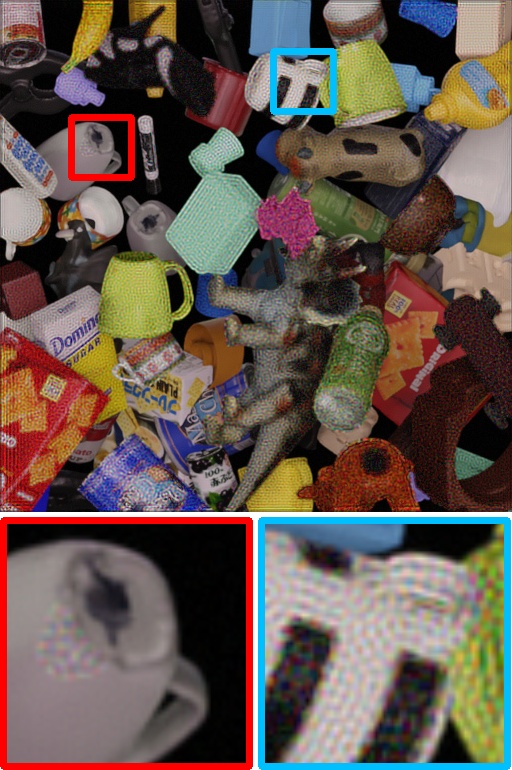}
&\includegraphics[width=3.35cm]{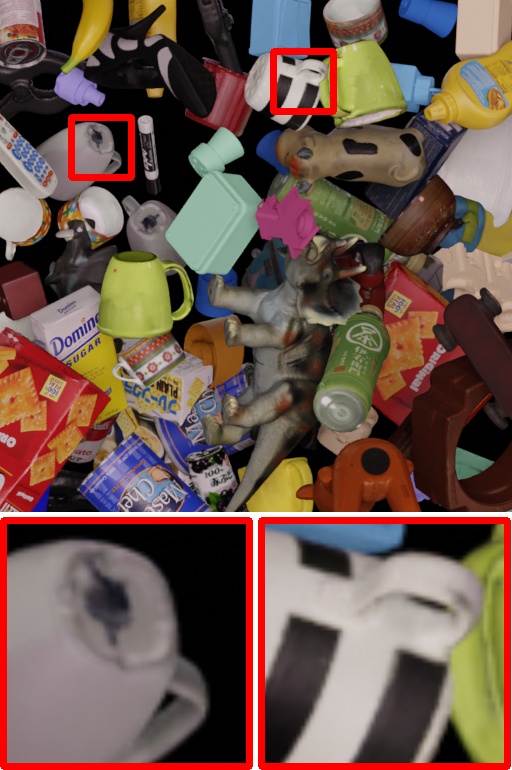}
\\
\raisebox{0.03\height}{\rotatebox{90}{\textbf{Near Focus}}}
&\includegraphics[width=3.35cm]{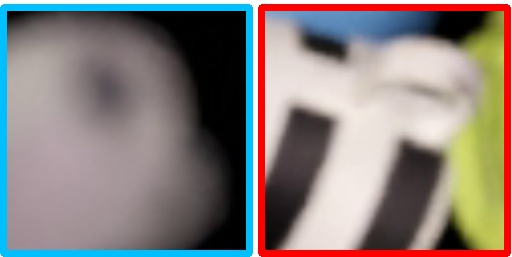} 
&\includegraphics[width=3.35cm]{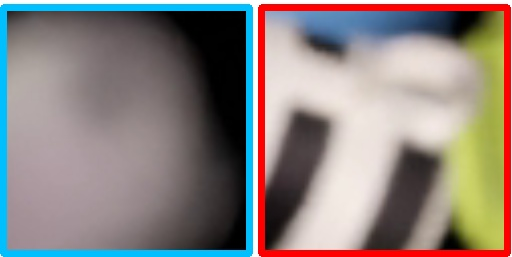} 
&\includegraphics[width=3.35cm]{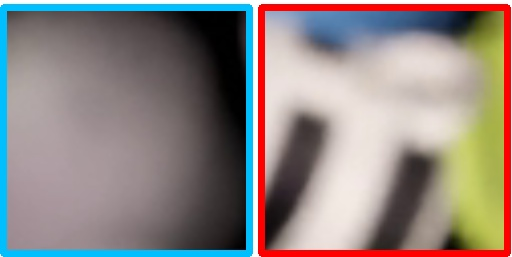}
&\includegraphics[width=3.35cm]{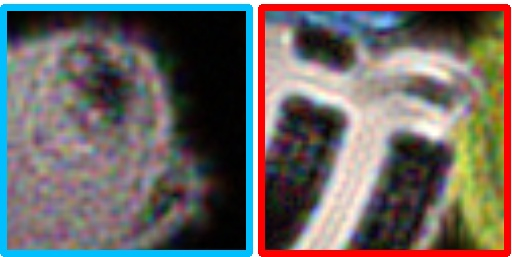}
&\includegraphics[width=3.35cm]{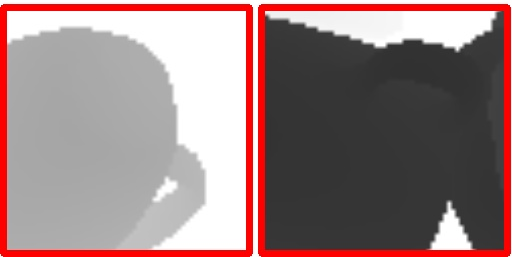}
\\ 
\vspace{2mm}
\end{tabular}
\end{center}
\caption{Simulated reconstruction of holograms synthesized by different methods on $512 \times 512$ image data. Only one set of results is provided for methods that do not consider varying pupil sizes and corresponding depth-of-field effects. The regions highlighted by red bounding boxes are closer to the focal plane while the regions in blue are more distant to the focal plane. Inaccurate and/or insufficient modeling of occlusions in Shi et al. \shortcite{shi2021nature} and \shortcite{shi2022light} produce ringing effects in out-of-focus areas. The method by Choi et al. \shortcite{choi2021neural3d} produces noise in out-of-focus areas due to the random phase. In contrast, our method demonstrates high-quality reconstructions with natural defocus effects and varying depth-of-field for different pupil sizes.
}
\label{figure:comp512}
\end{figure*}

Given that a focal stack provides more information about occluded elements within the underlying 3D scene than an RGB-D image, we have also developed an alternative approach. This method involves utilizing a neural framework to generate a 3D hologram using focal stack images that are evenly spaced.
We denote this variant as \emph{Ours (FS)} and the RGB-D framework as \emph{Ours} from hereon.
The focal stack variant of our framework is illustrated in \Cref{fig:alternative}. 
This framework utilizes a focal stack, comprising a sequence of focal images captured at evenly sampled distances, as its input data. To encourage the utilization of the rich cues present in the focal images, we employ a 
%
neural module $G_\text{fs}$ to predict intermediate phase values which we propagate and process to compute the final hologram. Furthermore, to ensure adaptability to varying pupil sizes, $G_\text{fs}$ leverages the capabilities of the ADC layer.
As shown in \Cref{fig:alternative}, $G_\text{fs}$ is used to predict a group of initial phases $\{\Phi_t^{(s)}\}_{t=1}^{N}$ and a set of masks $\{ W_t^{s} \}_{t=1}^{N}$ as
\begin{equation}\label{eq:phs_init}
\left \{{\Phi}_t^{(s)}, W_t^{(s)} \right \}_{t=1}^{N} = G_\text{fs} \left ( \left \{ R_t^{(s)}\right \}_{t=1}^N; \:\: s \right ),
\end{equation}
where $\Phi_t^{(s)}$ and $W_t^{(s)}$ denote the predicted initial phase map and mask for the $t$th image plane, respectively, for a given pupil size $s$.
We numerically propagate the wavefield of the predicted phases at each plane towards the SLM plane, computed as
\begin{equation}\label{eq:prop}
F_t^{(s)} =  f_p \left (\left\{R_t^{(s)},{\Phi}_t^{(s)} \right\}, -u_t \right ),
\end{equation}
where $F_t^{(s)}$ represents the $t$th propagated field, $-u_t$ denotes the backward distance from the reconstruction plane to the SLM plane.
As shown in \Cref{fig:alternative}, the predicted weight maps $\{W_t^{(s)}\}_{t=1}^N$ for the sampled focal planes are used to composite the output hologram $F_{h}^{(s)}$. 
The composition process is given by
\begin{equation}
F_{h}^{(s)} = \sum_{t=1}^{N} W_{t}^{(s)} \cdot F_t^{(s)}.
\end{equation}
Finally, we extract the amplitude and phase from the composited wavefield to obtain the complex hologram as
\begin{equation}
H_{A}^{(s)} =  \Big | F_{h}^{(s)} \Big|, \:\:\:
H_{\Phi}^{(s)} = \angle \left(F_{h}^{(s)}\right),
\end{equation}
where $|\cdot|$ and $\angle (\cdot)$ denotes the amplitude and phase. 
This framework is trained with the same supervision strategy described in \Cref{sec:objecive}.


\setlength{\tabcolsep}{1.5pt}
\renewcommand{\arraystretch}{0.6}
\begin{figure*}[t]
\begin{center}
\small
\begin{tabular}{cccccc}
& \textbf{Ours (2mm Pupil)} & \textbf{Ours (3mm Pupil)} & \textbf{Ours (4mm Pupil)} & \textbf{Shi et al. 2021}  & \textbf{Shi et al. 2022}  \\[0.5ex]
\raisebox{1.05\height}{\rotatebox{90}{\textbf{Far Focus}}}
&\includegraphics[width=3.4cm]{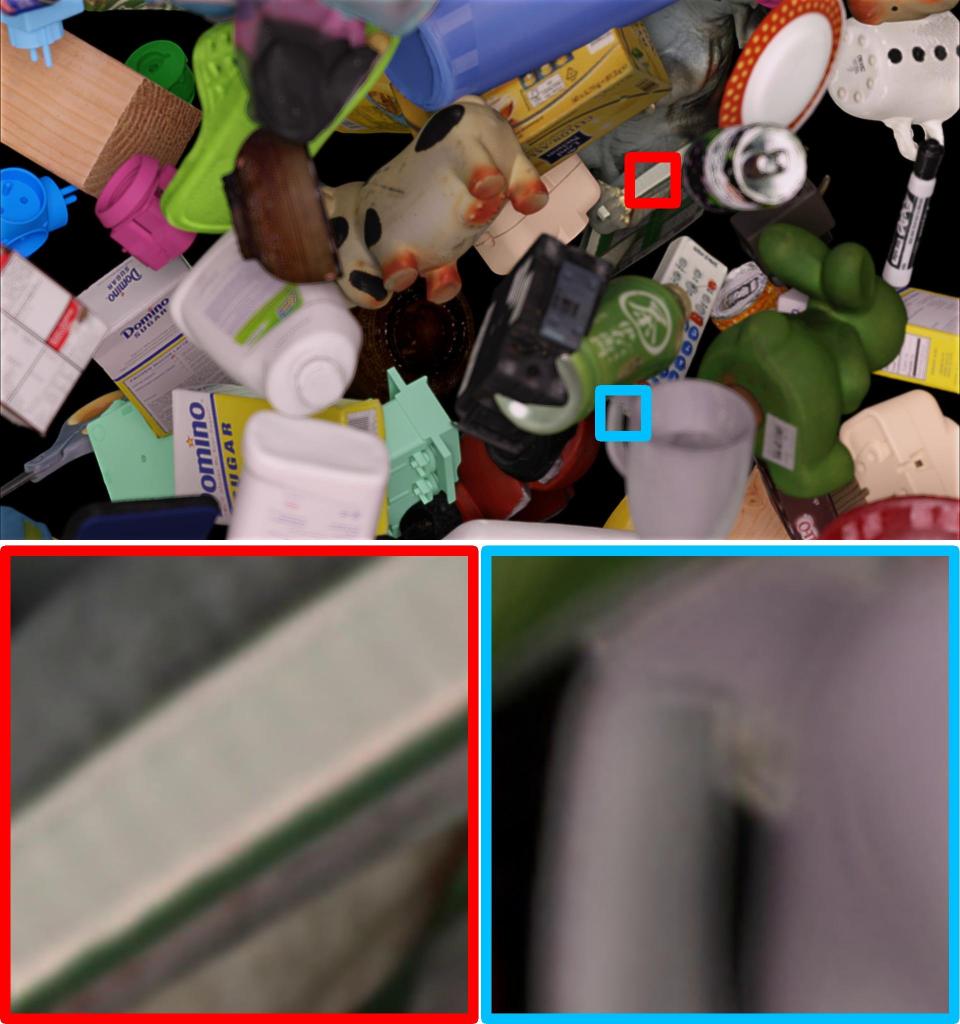}
&\includegraphics[width=3.4cm]{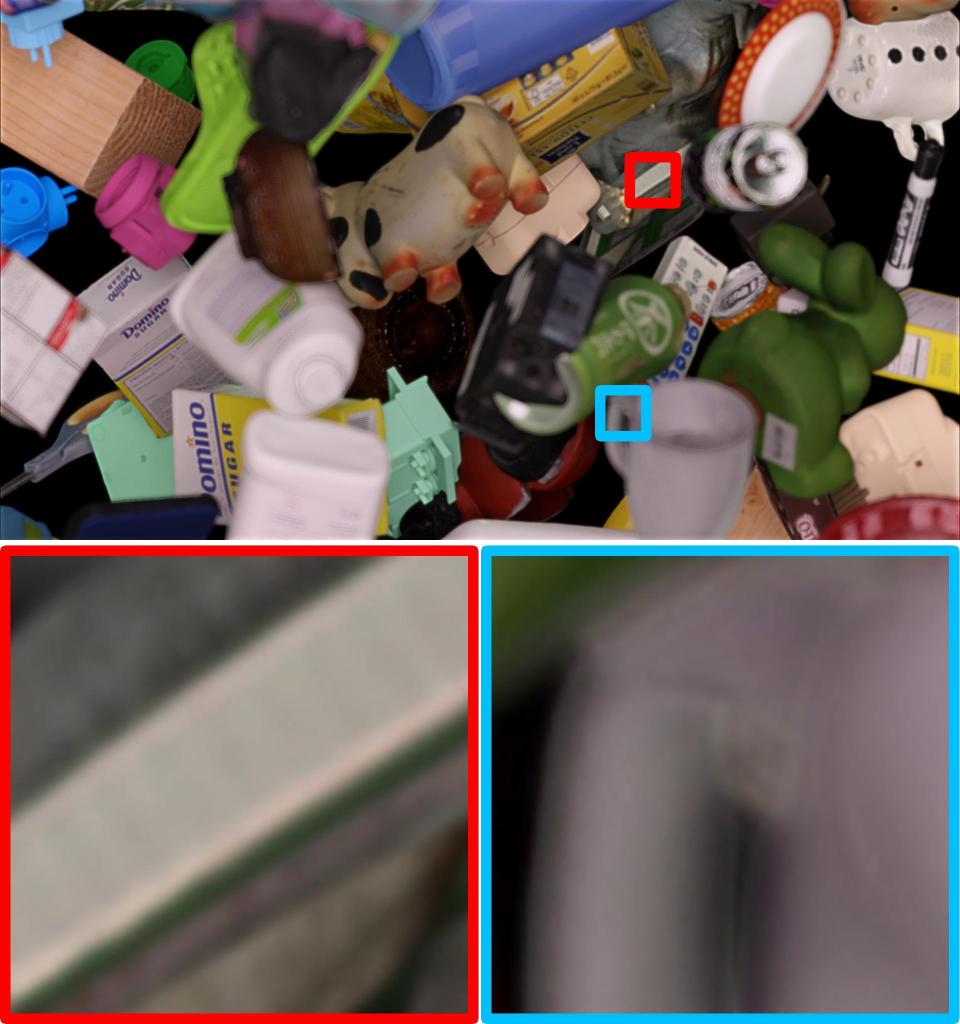}
&\includegraphics[width=3.4cm]{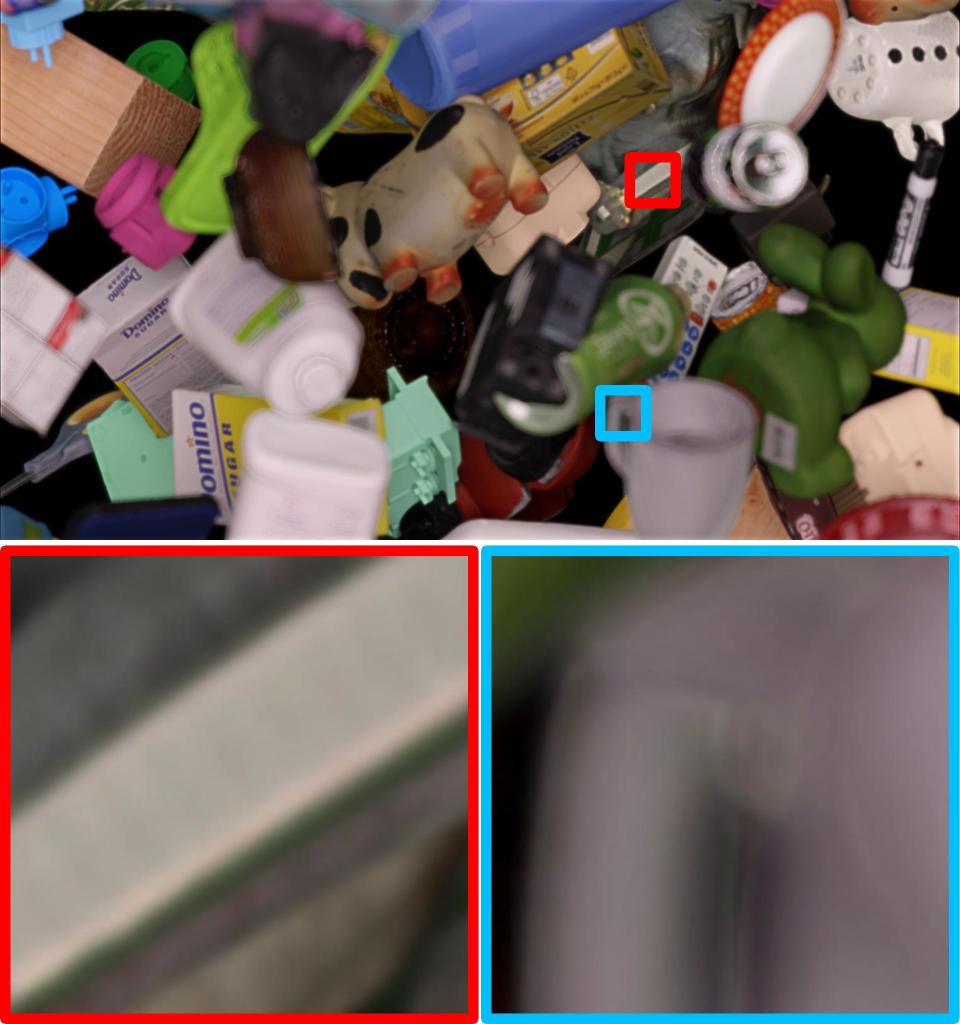} 
&\includegraphics[width=3.4cm]{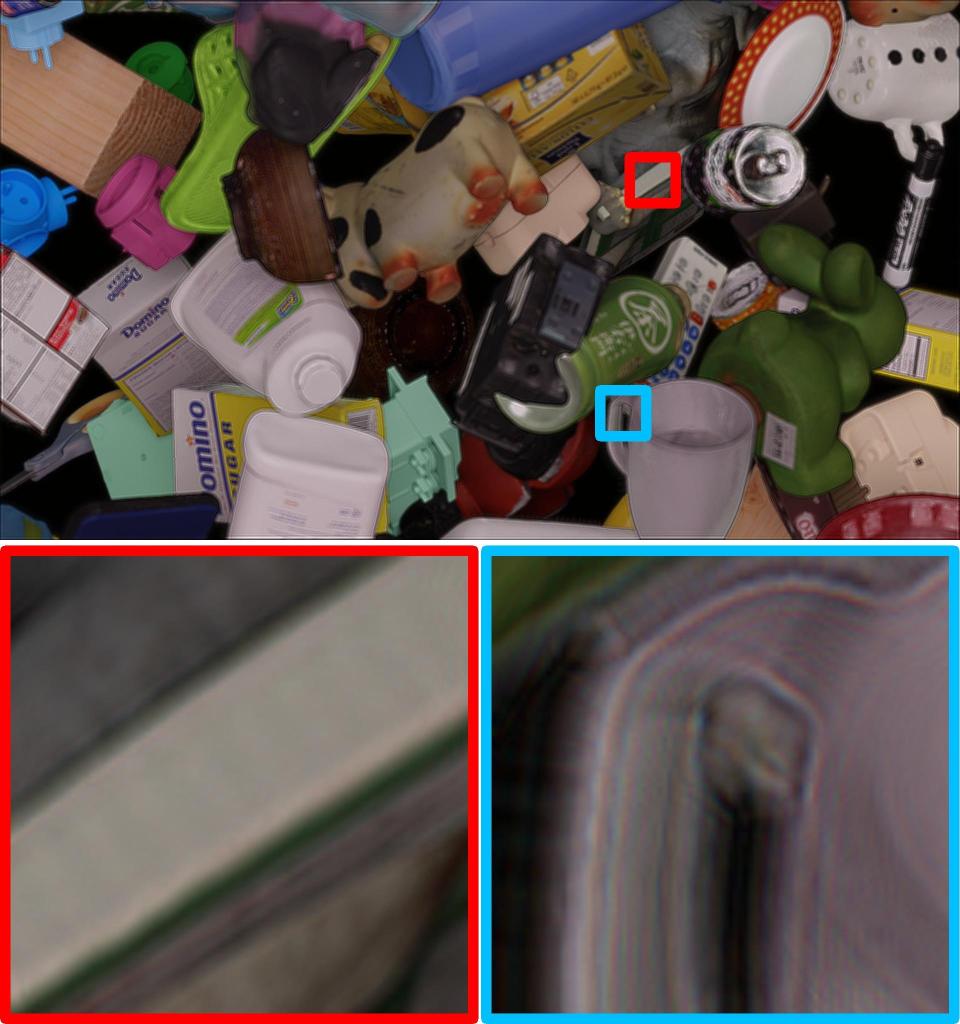} 
&\includegraphics[width=3.4cm]{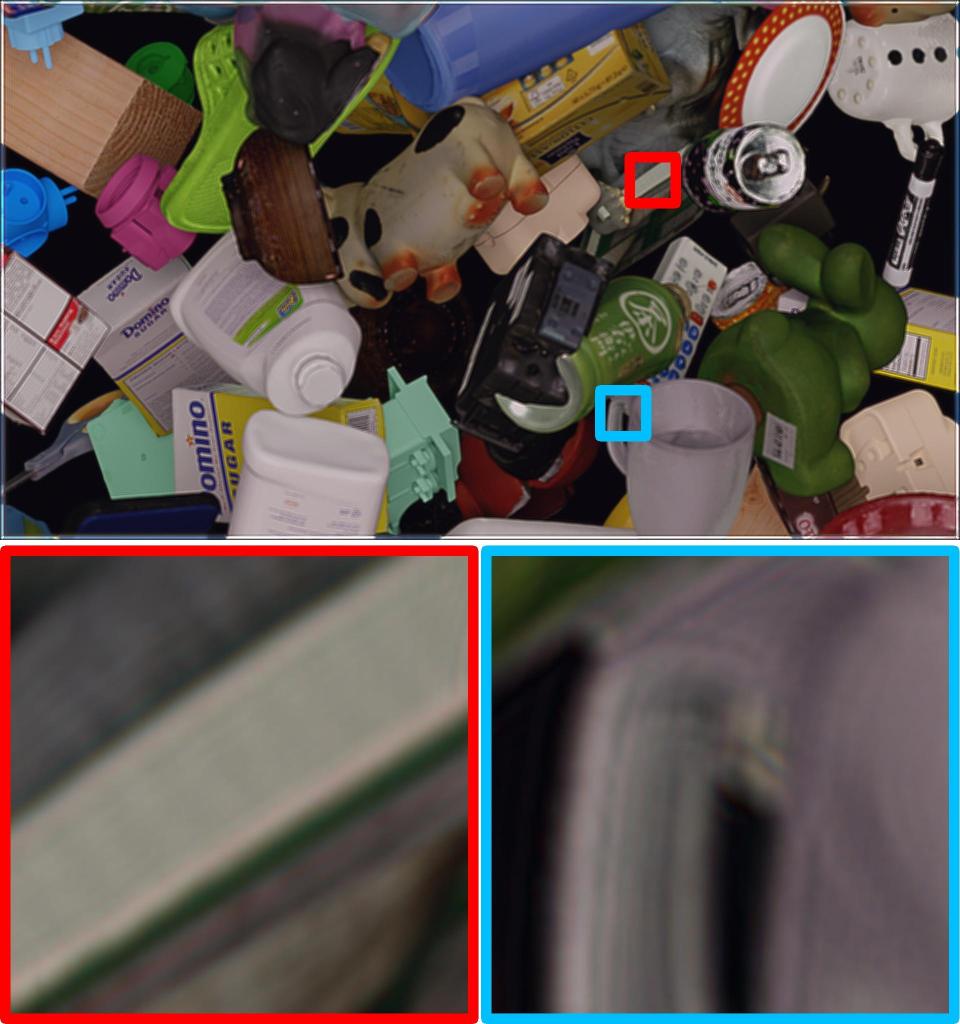}
\\
\raisebox{0.02\height}{\rotatebox{90}{\textbf{Near Focus}}}
&\includegraphics[width=3.4cm]{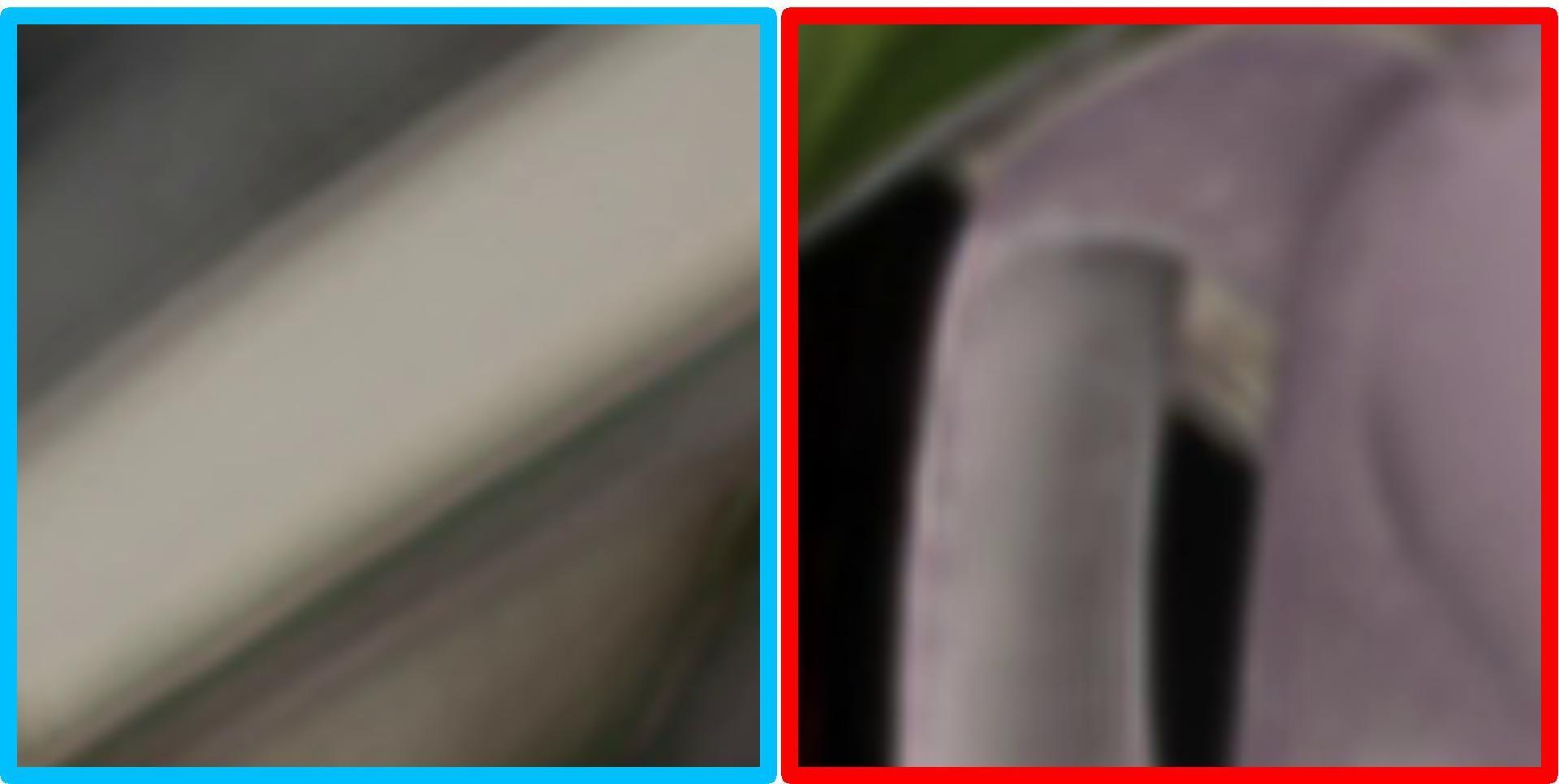}
&\includegraphics[width=3.4cm]{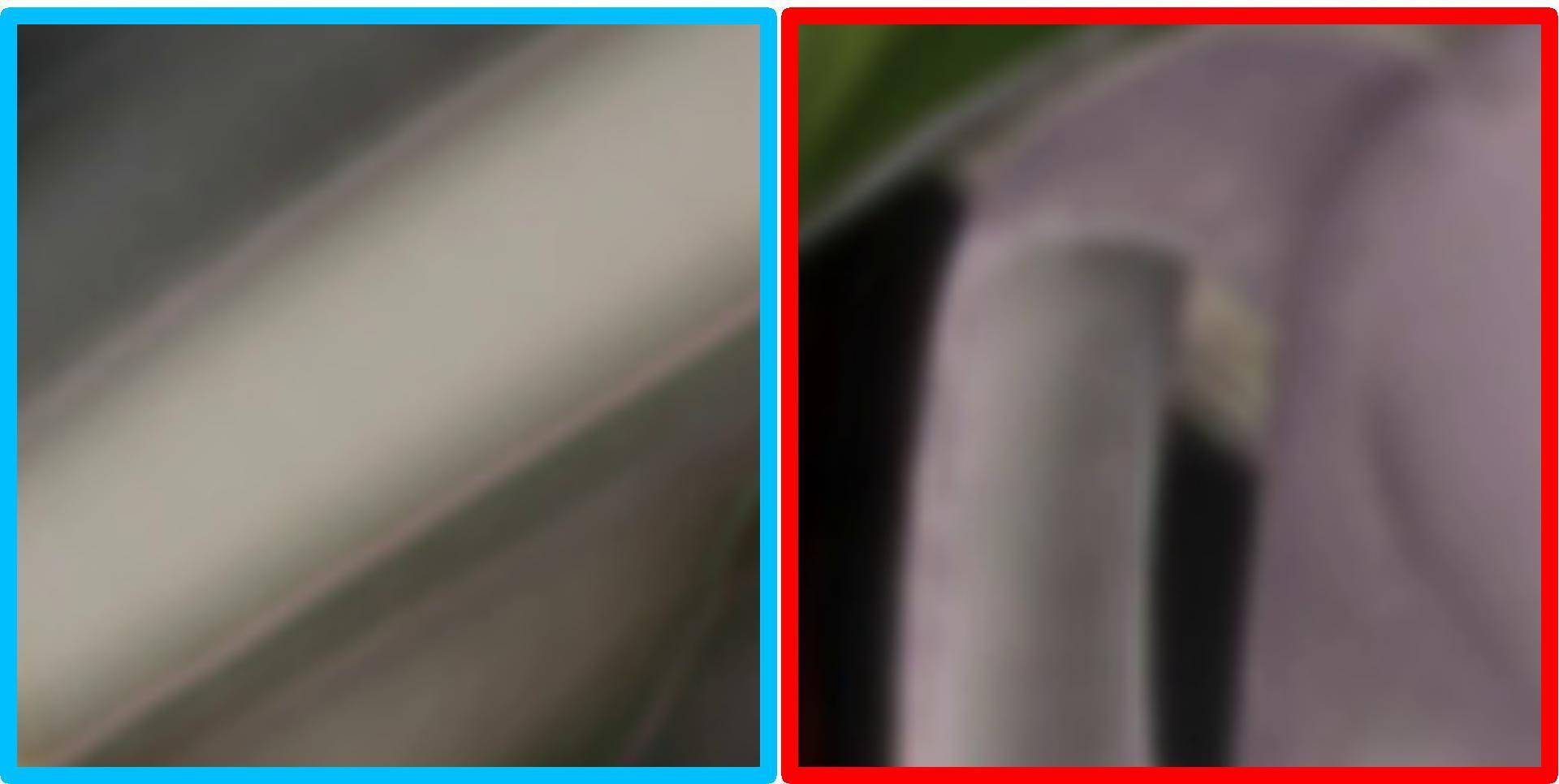}
&\includegraphics[width=3.4cm]{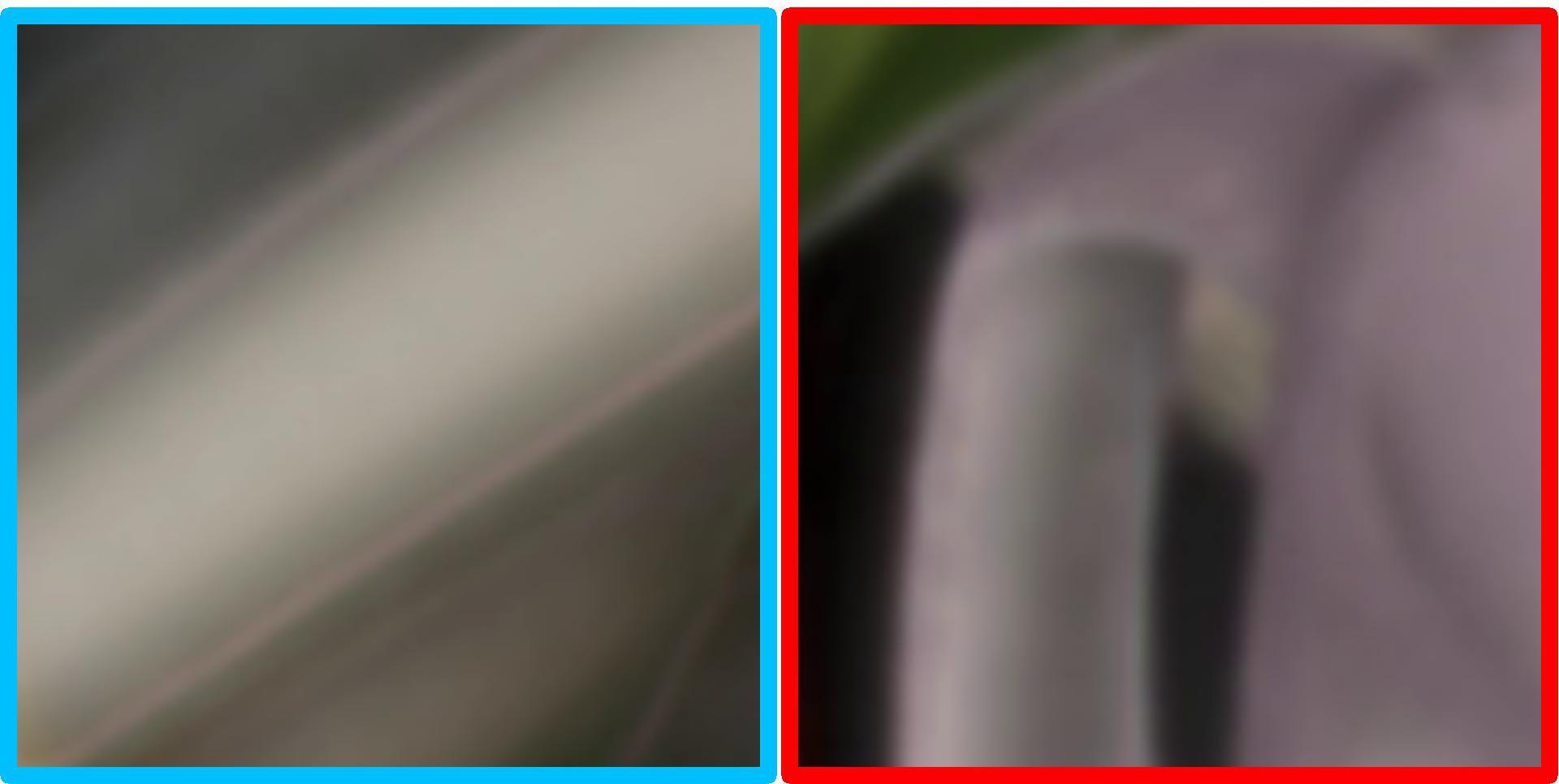} 
&\includegraphics[width=3.4cm]{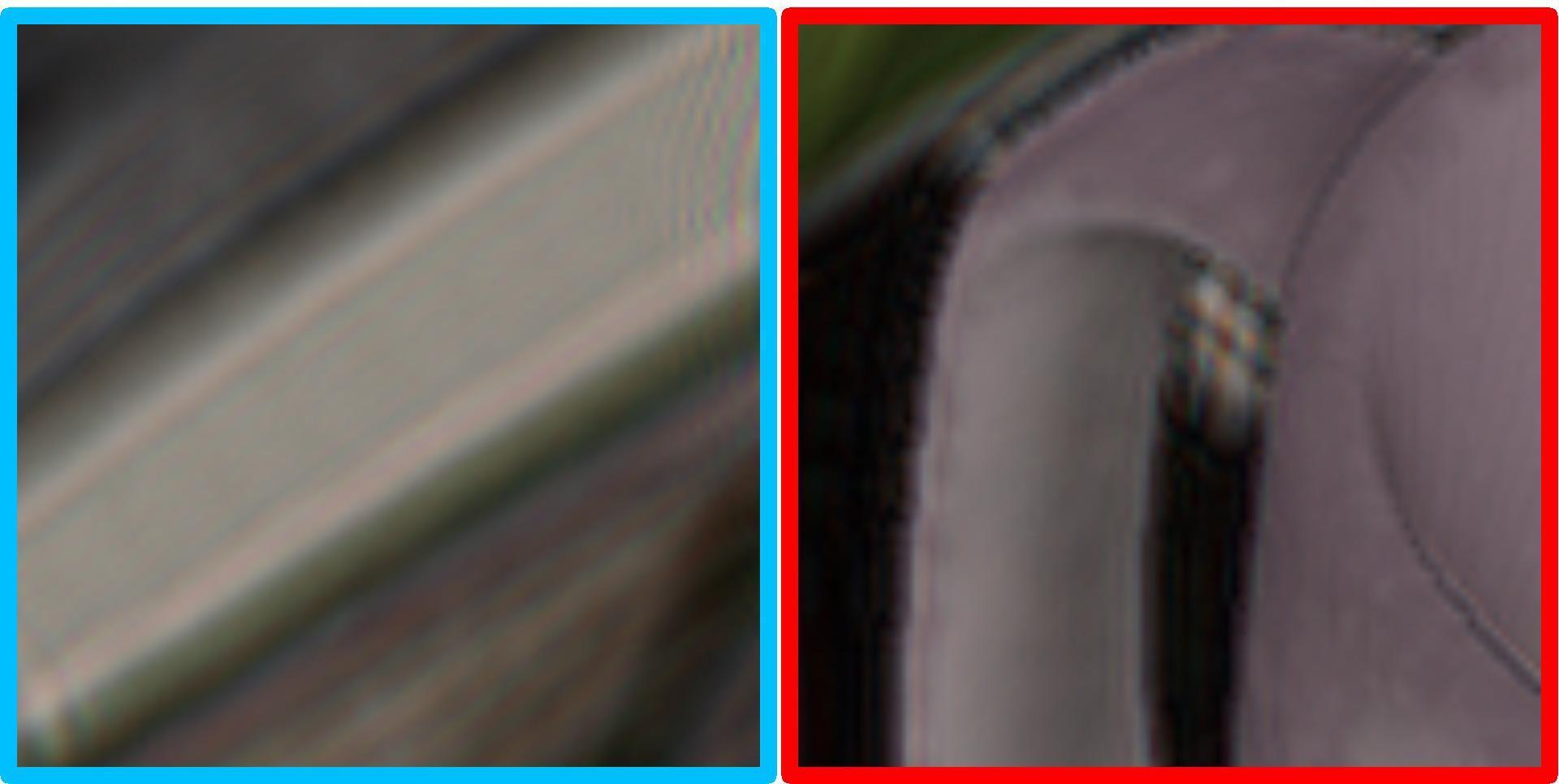} 
&\:\includegraphics[width=3.4cm]{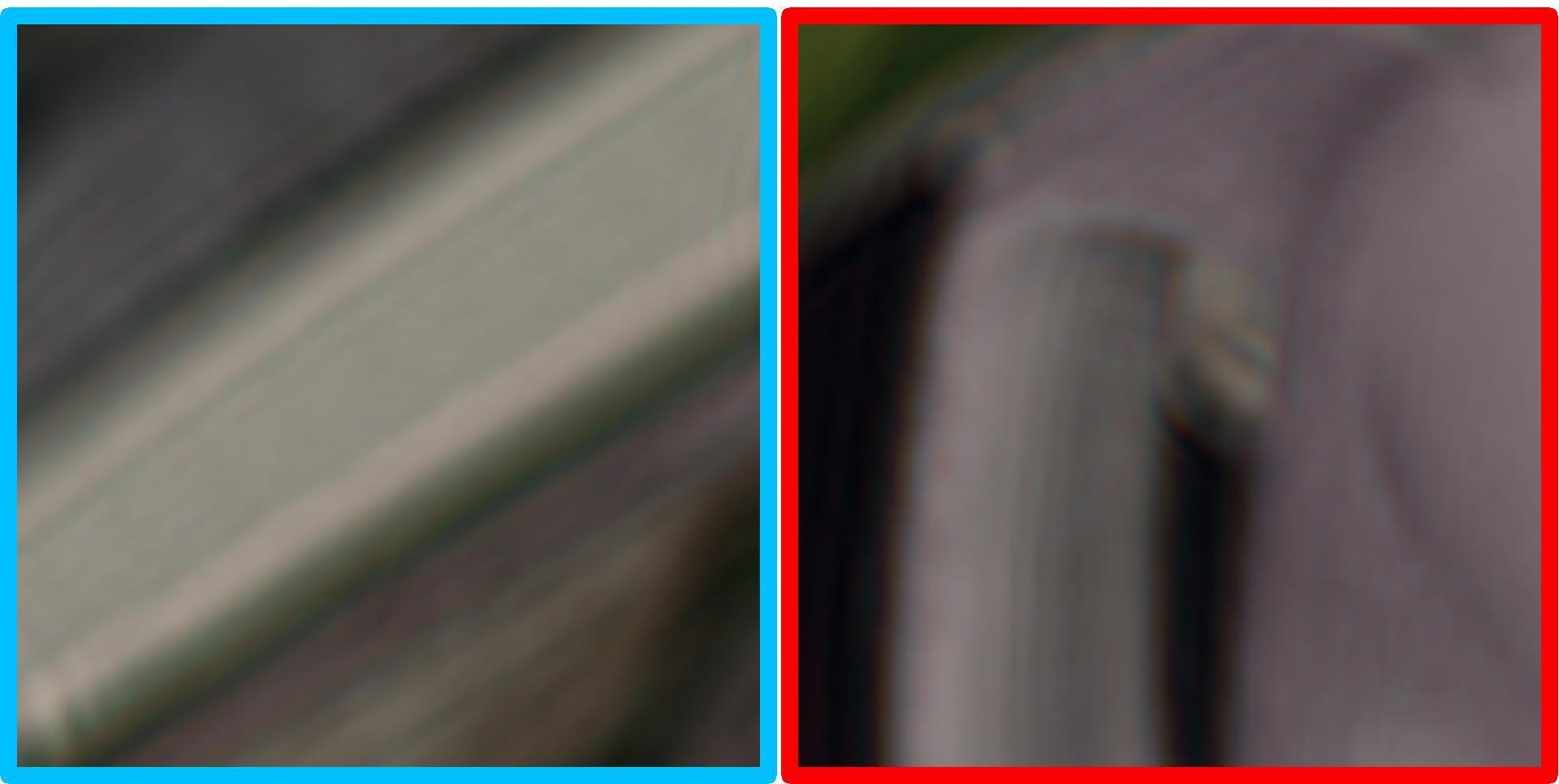}
\\[0.7ex]
&\textbf{Rendered (2mm Pupil)} &\textbf{Rendered (3mm Pupil)} & \textbf{Rendered (4mm Pupil)}  & \textbf{Choi et al. 2021} & \textbf{All-in-Focus RGB-D}
\\[0.7ex]
\raisebox{1.05\height}{\rotatebox{90}{\textbf{Far Focus}}}
&\includegraphics[width=3.4cm]{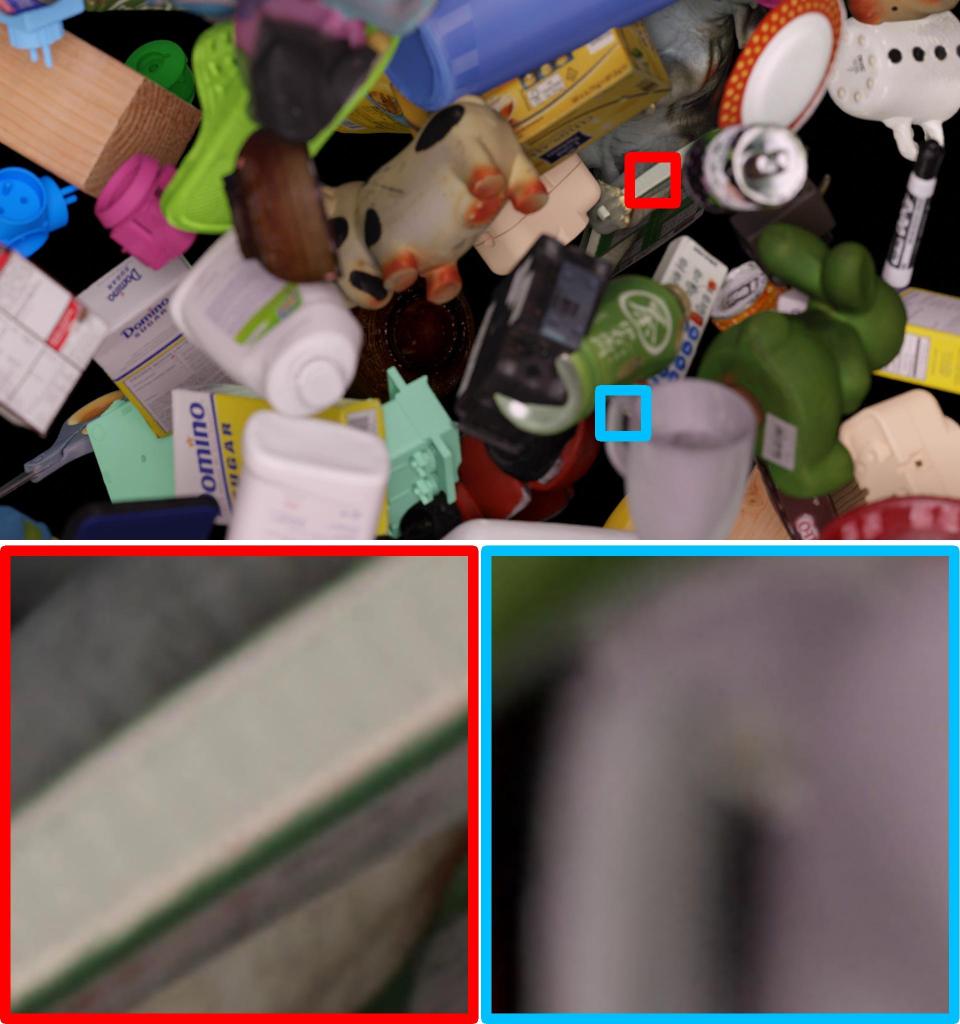} 
&\includegraphics[width=3.4cm]{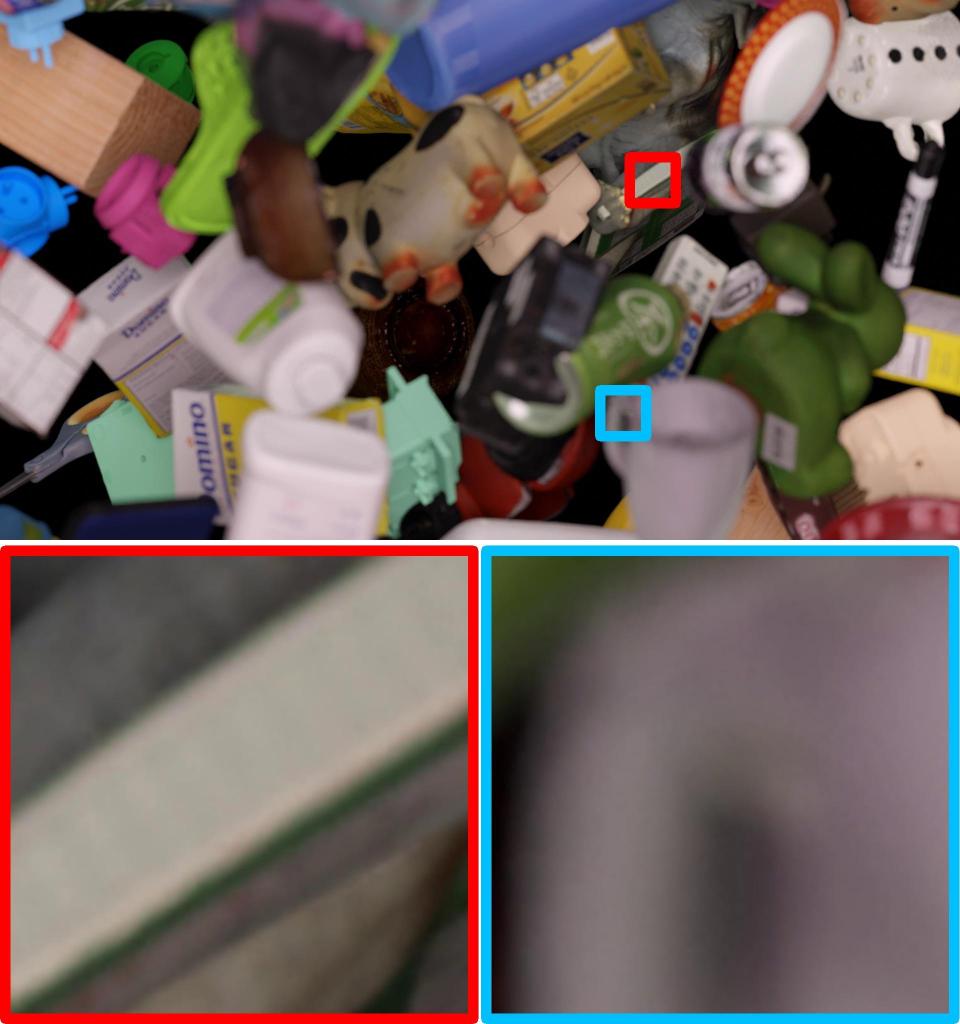} 
&\includegraphics[width=3.4cm]{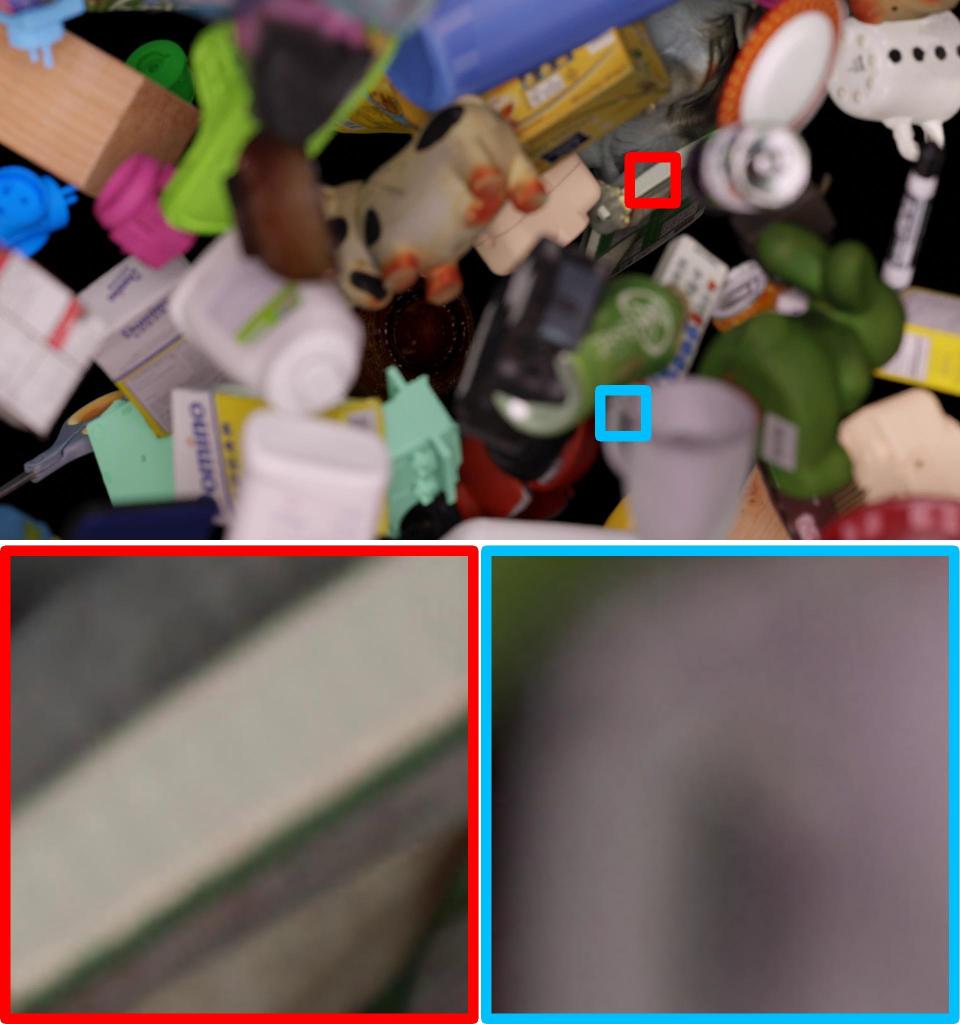} 
&\includegraphics[width=3.4cm]{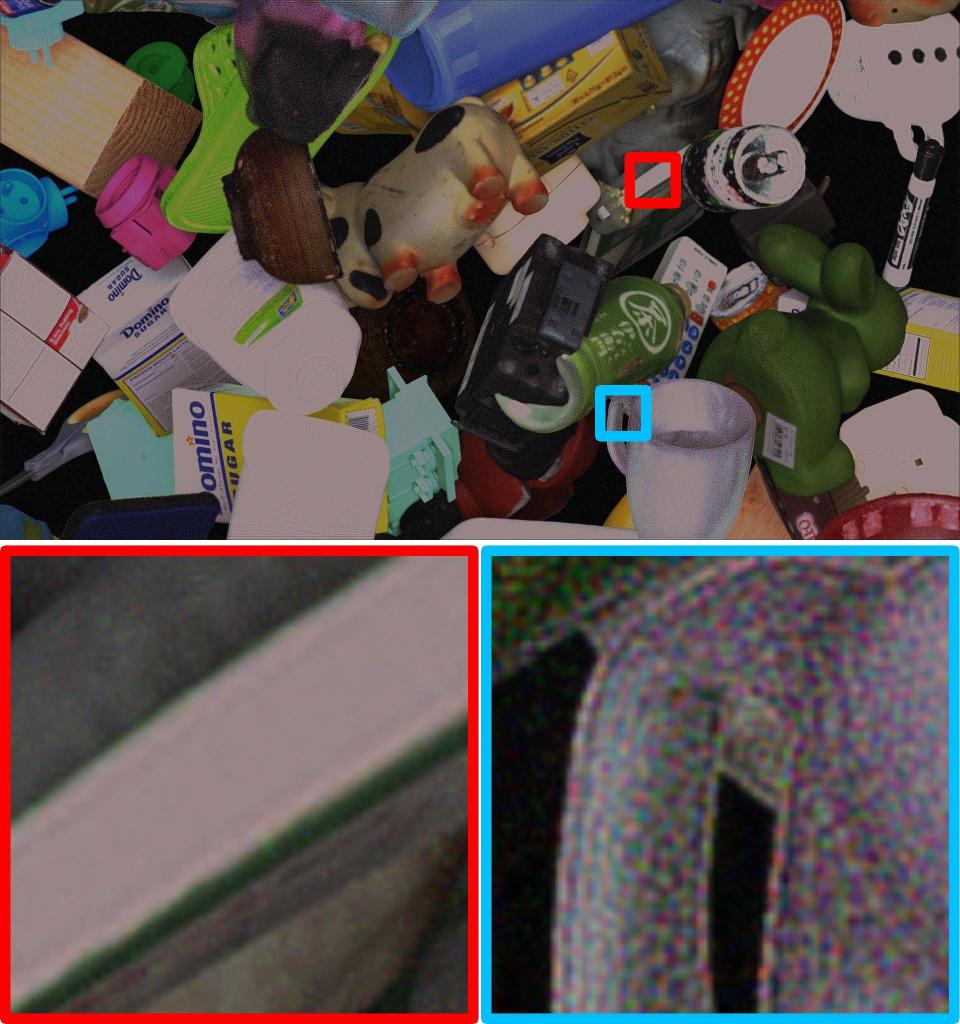}
&\includegraphics[width=3.4cm]{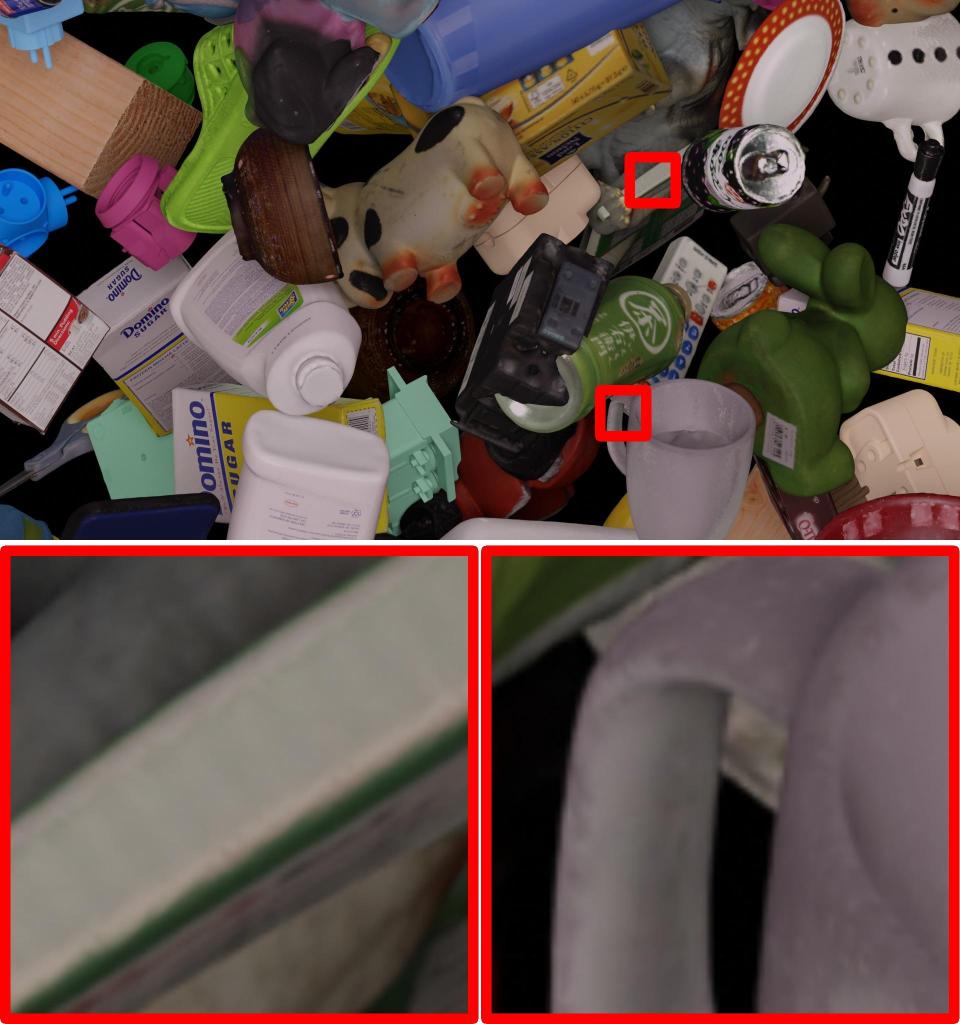}
\\
\raisebox{0.02\height}{\rotatebox{90}{\textbf{Near Focus}}}
&\includegraphics[width=3.4cm]{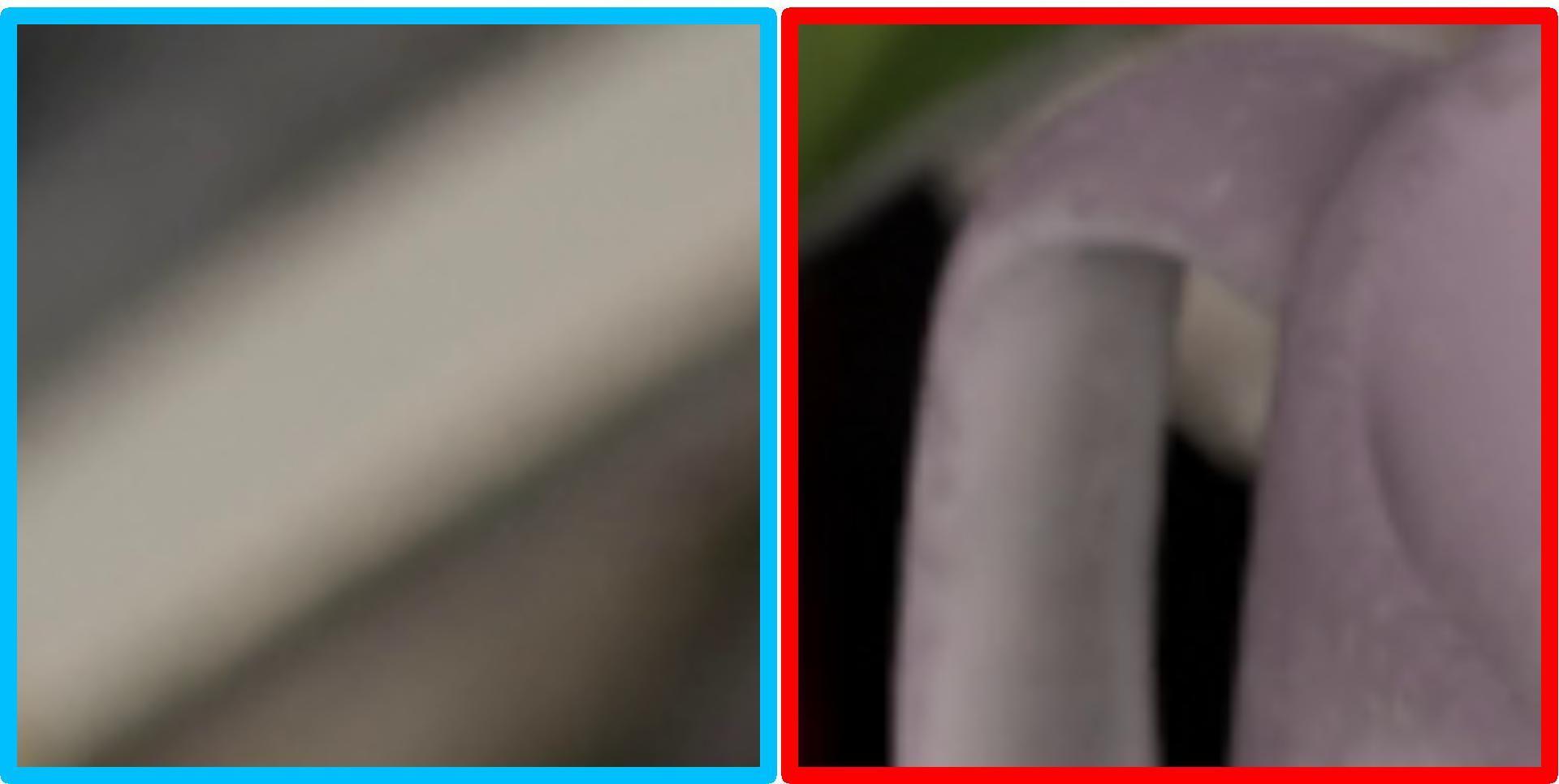} 
&\includegraphics[width=3.4cm]{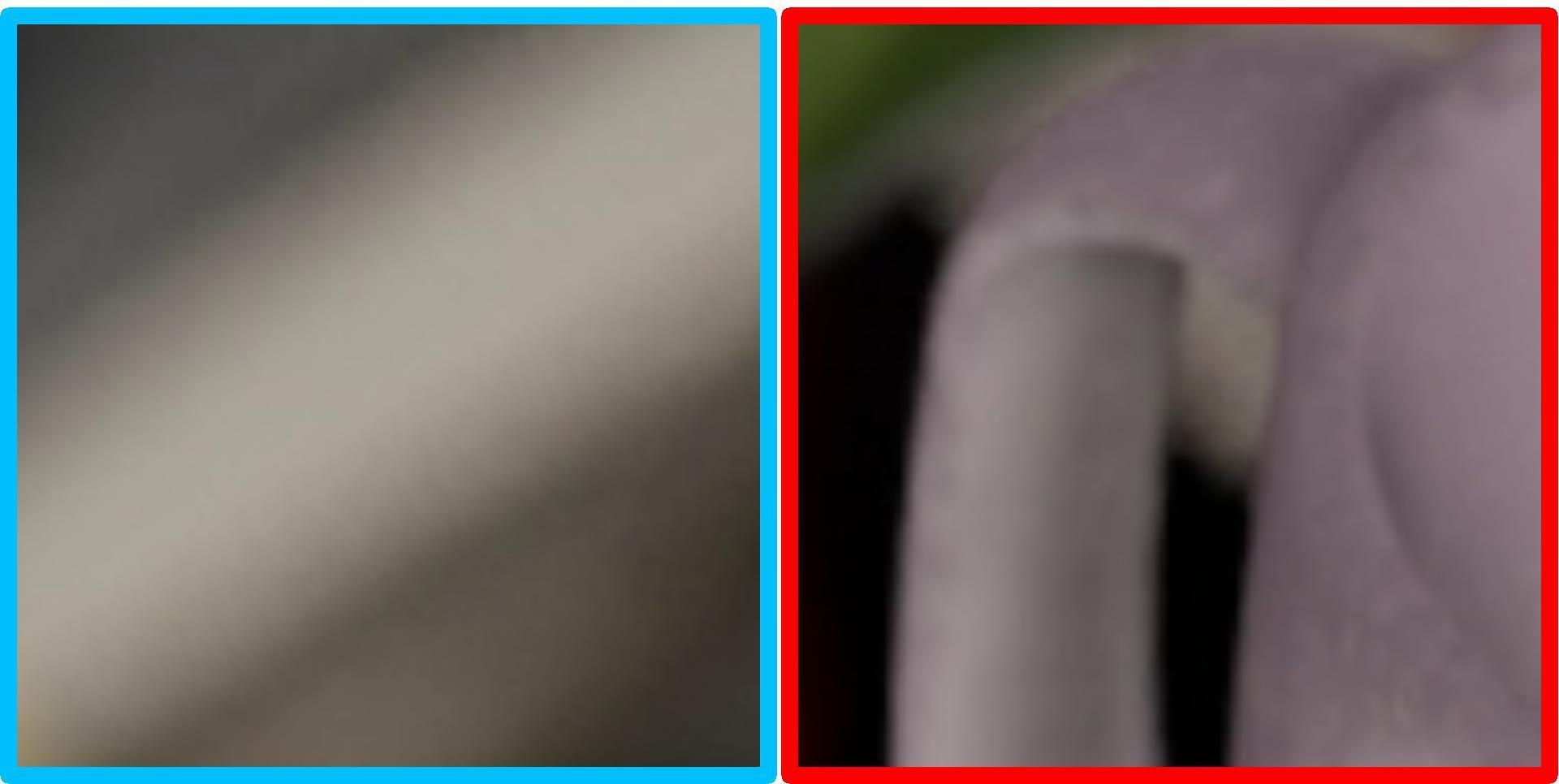} 
&\includegraphics[width=3.4cm]{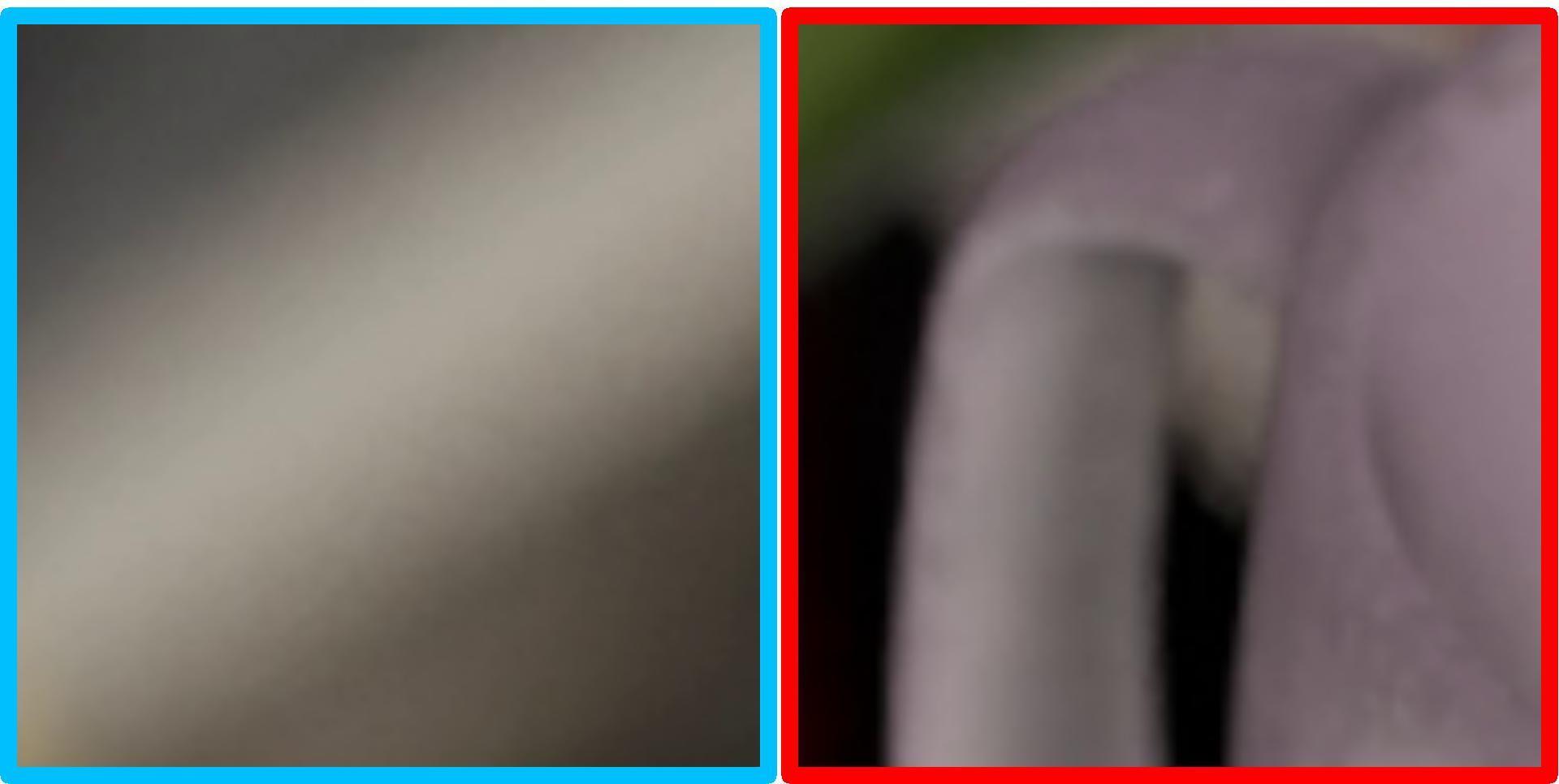} 
&\includegraphics[width=3.4cm]{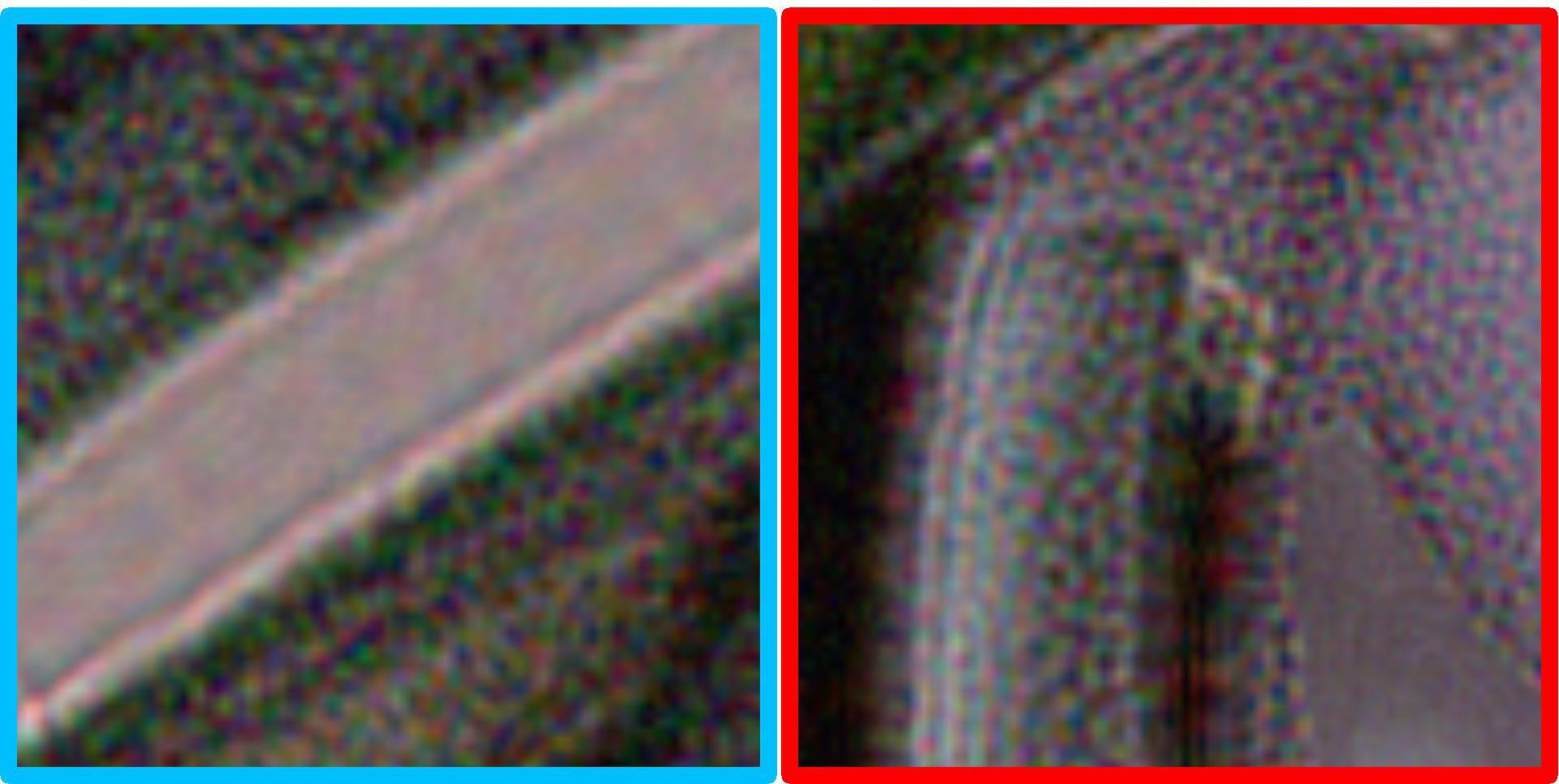}
&\includegraphics[width=3.4cm]{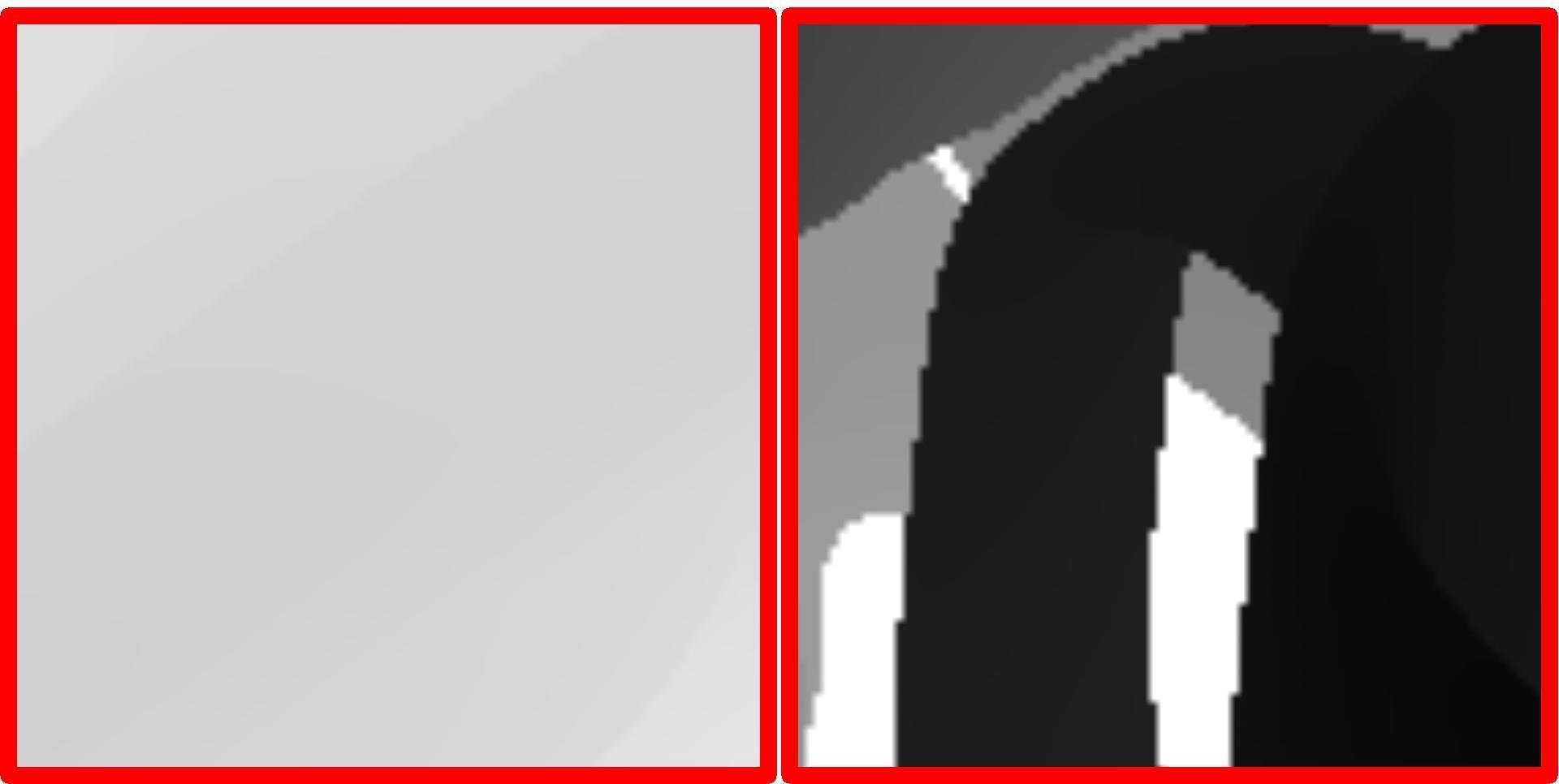}
\\  
\end{tabular}
\end{center}
\caption{Simulated reconstruction of holograms synthesized by different methods on $1920\times1080$ image data. The regions highlighted by red bounding boxes are closer to the focal plane while the regions in blue boxes are more distant to the focal plane. Only one set of results is provided for methods that do not consider varying pupil sizes and corresponding depth-of-field effects. 
The results from state-of-the-art methods present severe ringing artifacts, very subtle defocus effects or noise in out-of-focus regions. 
In contrast, our results exhibit appropriate defocus effects in out-of-focus regions, for instance, the cup handle and the white stripes pattern, also including effects from changes in pupil sizes. 
}
\label{figure:comp_1080}
\end{figure*}
\setlength{\tabcolsep}{2pt}
\renewcommand{\arraystretch}{0.6}
\begin{figure*}[t]
\begin{center}
\small
\begin{tabular}{cccl}
& \textbf{Ours (2mm Pupil)} & \textbf{Ours (3mm Pupil)} &\textbf{\:\:\:\:\:\:\:\:\:\:\:\:\:\:\:\:\:\:\:\:\:\:\:\:\:\:Ours (4mm Pupil)}   \\ [0.5ex]
\raisebox{0.45\height}{\rotatebox{90}{\textbf{Far Focus}}}
&\includegraphics[width=5.68cm]{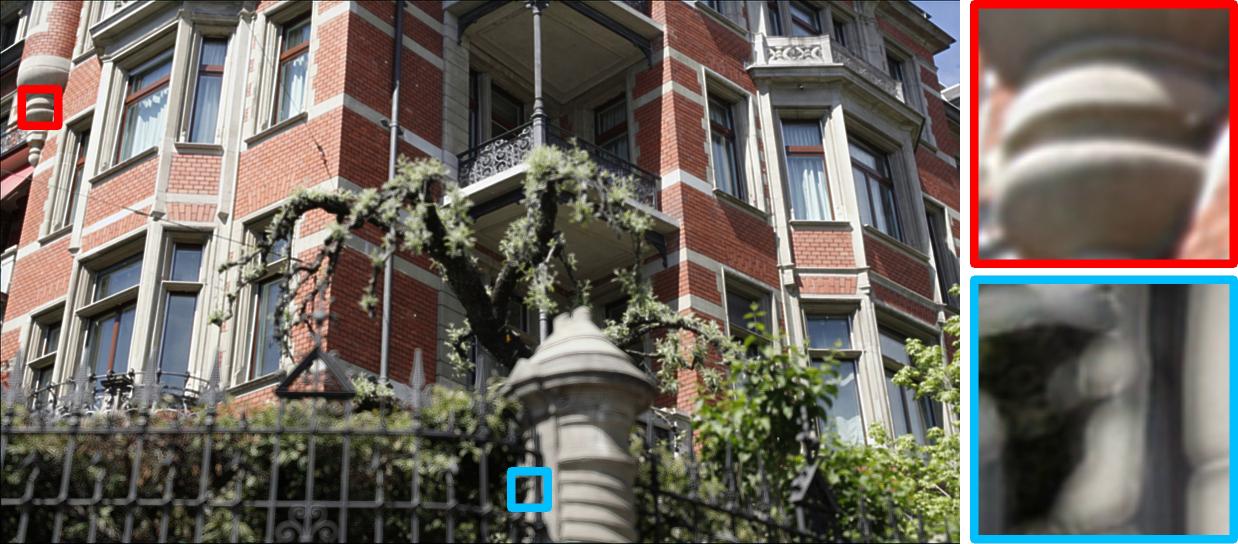}
&\includegraphics[width=5.68cm]{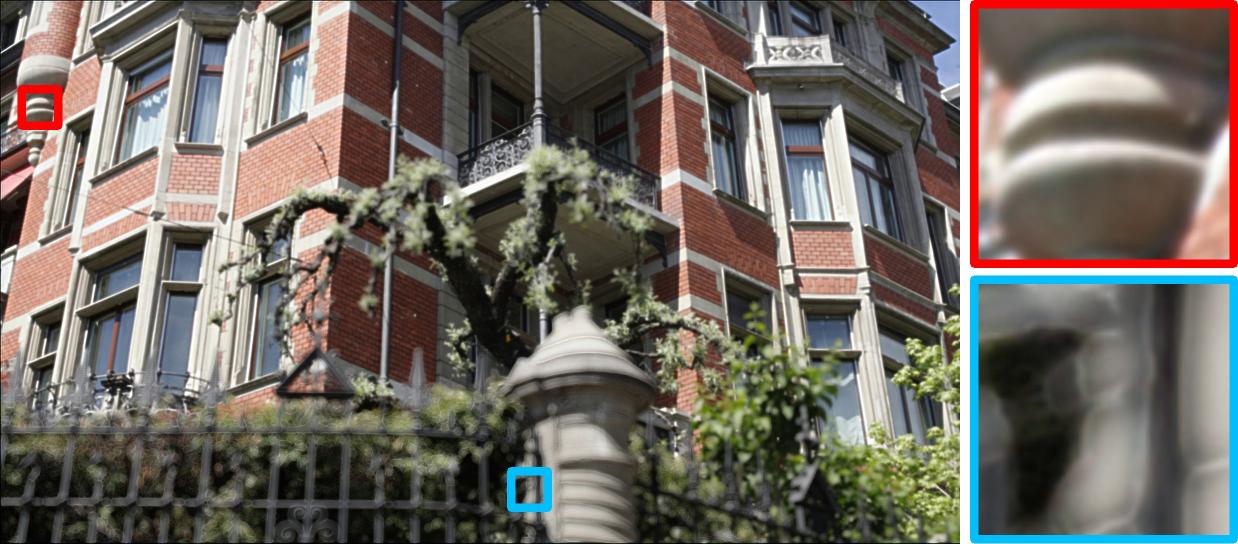}
&\includegraphics[width=5.68cm]{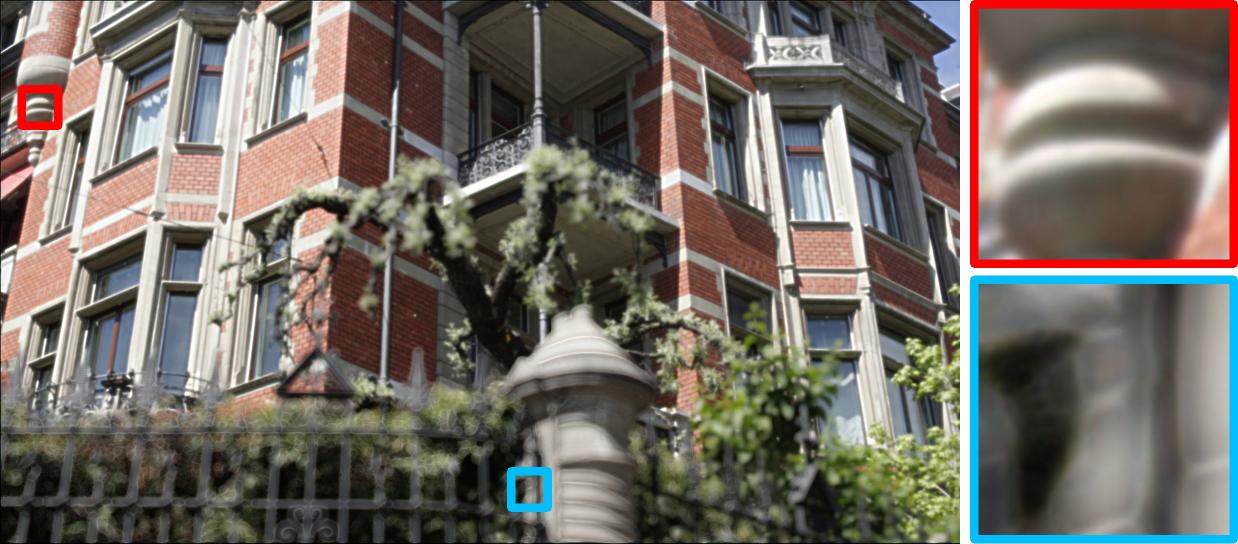}
\\
\raisebox{0.25\height}{\rotatebox{90}{\textbf{Near Focus}}}
&\includegraphics[width=5.7cm]{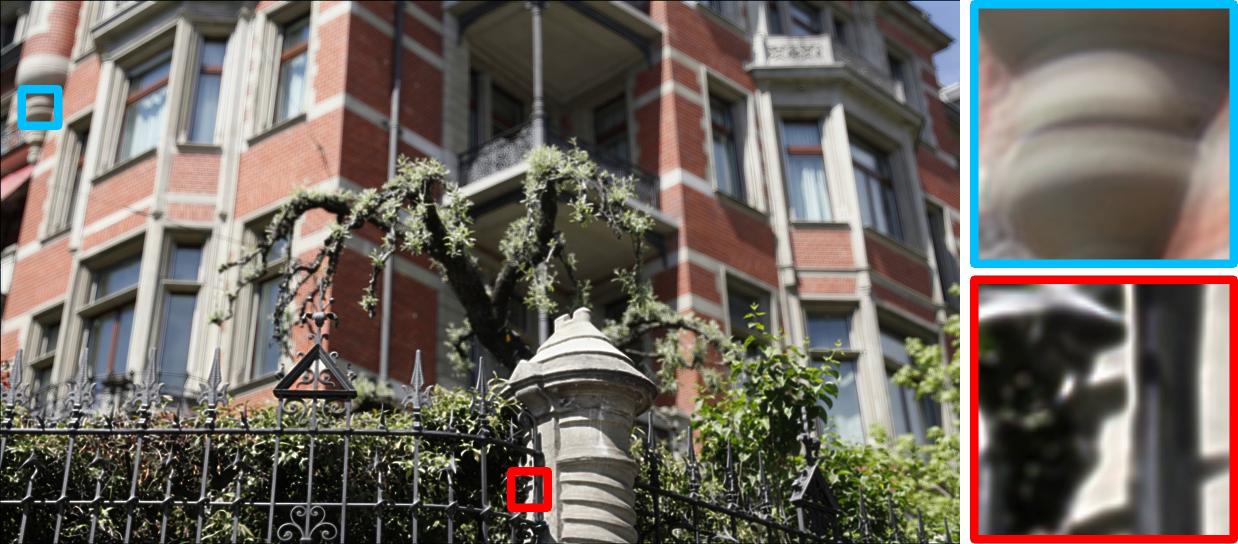}
&\includegraphics[width=5.7cm]{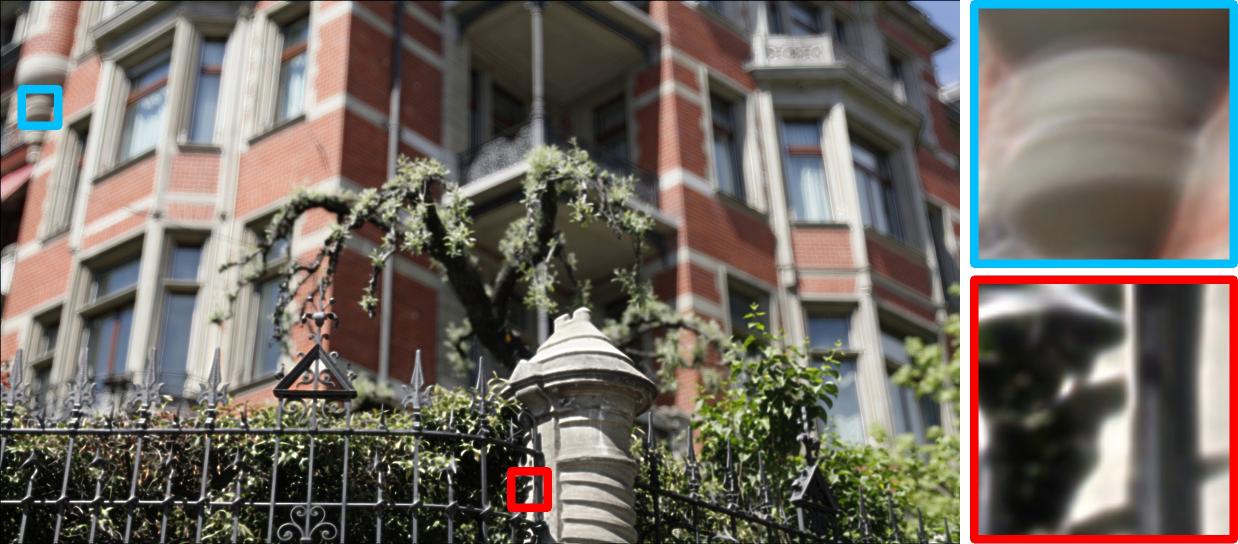}
&\includegraphics[width=5.7cm]{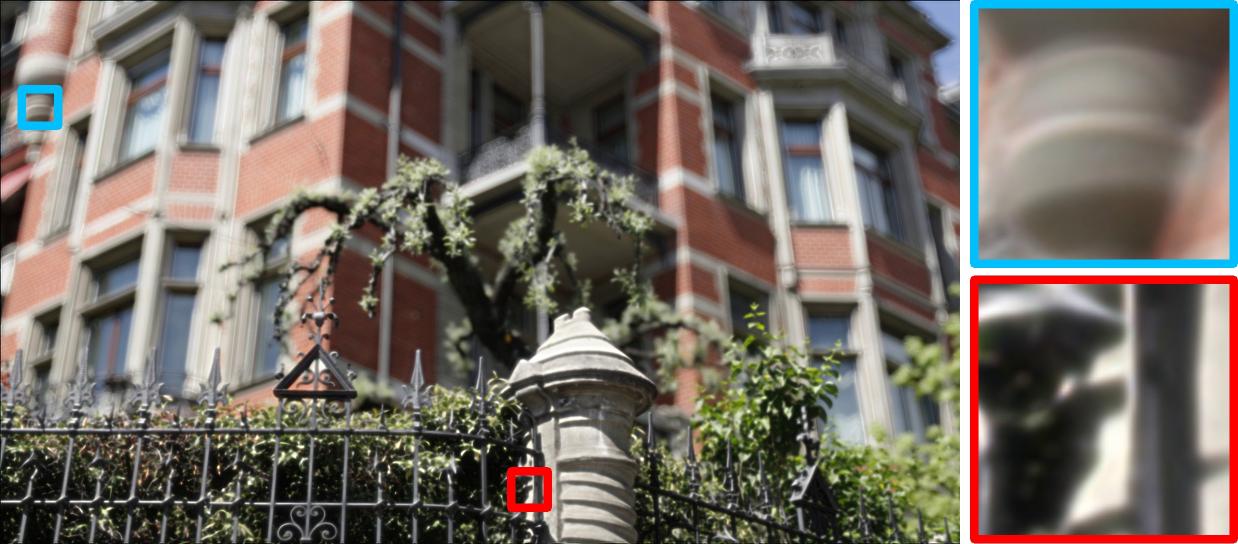}
\\[0.7ex]
& \textbf{Shi et al. 2021} & \textbf{Shi et al. 2022} & \textbf{\:\:\:\:\:\:\:\:\:\:\:\:\:\:\:\:\:\:\:\:\:\:\:\:\:\:\:\:\:\:Choi et al. 2021} \\ [0.7ex]
\raisebox{0.4\height}{\rotatebox{90}{\textbf{Far Focus}}}
&\includegraphics[width=5.7cm]{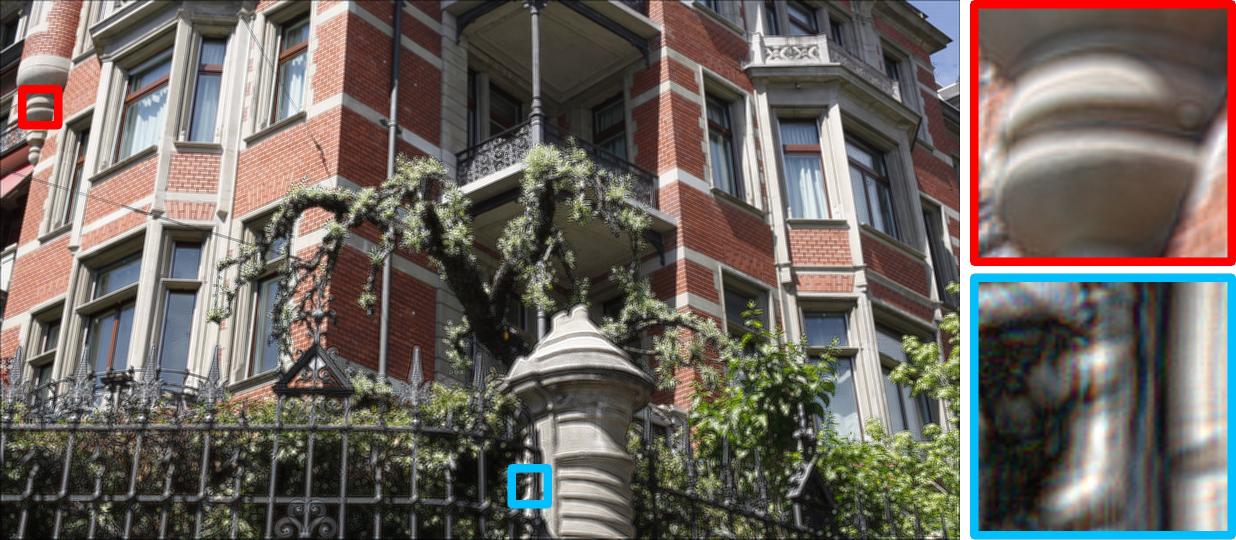}
&\includegraphics[width=5.7cm]{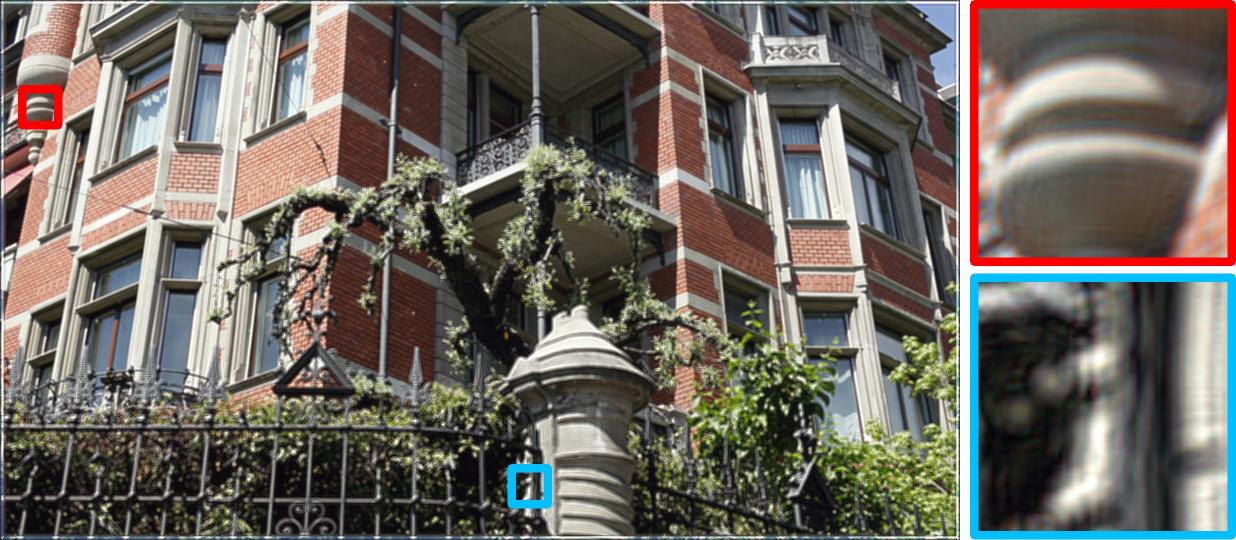}
&\includegraphics[width=5.7cm]{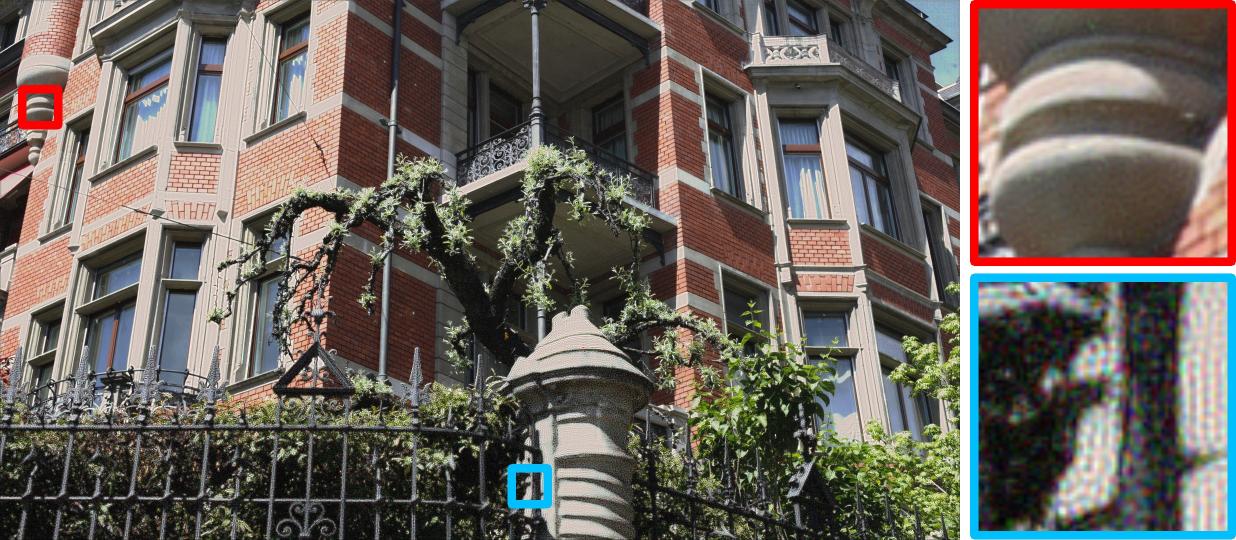}
\\
\raisebox{0.25\height}{\rotatebox{90}{\textbf{Near Focus}}}
&\includegraphics[width=5.7cm]{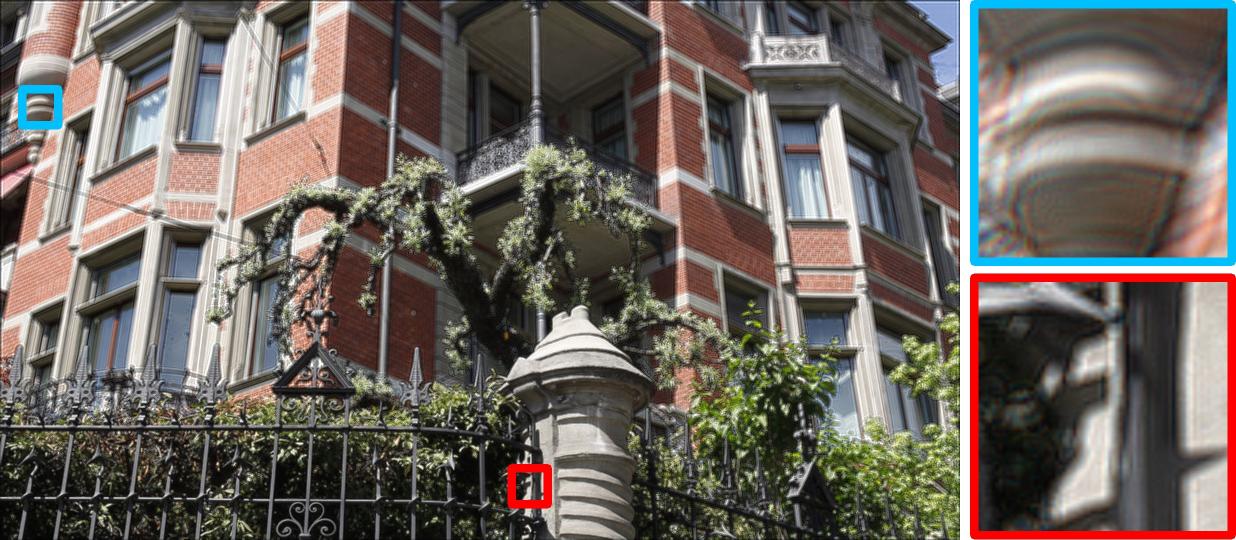}
&\includegraphics[width=5.7cm]{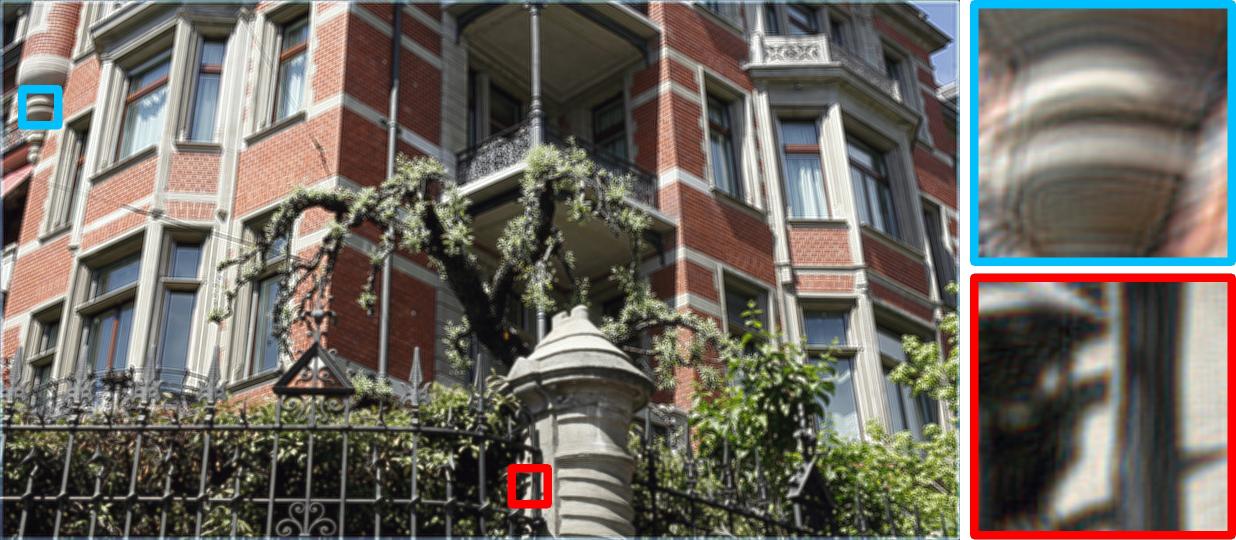}
&\includegraphics[width=5.7cm]{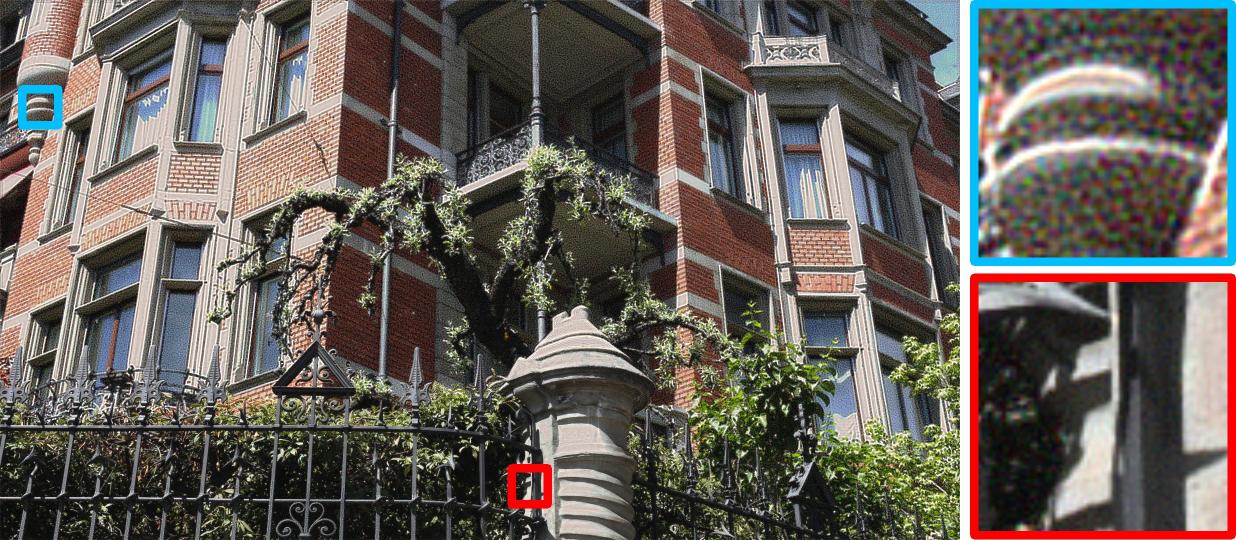}
\\
\end{tabular}
\end{center}
\caption{Simulated reconstruction of holograms obtained by different methods on in-the-wild data.
In-focus regions are highlighted in red and out-of-focus regions are marked in blue. Only one group of results is provided for methods that do not consider varying pupil sizes and corresponding depth-of-field effects. As shown in the insets (the distant building part and the closer iron fence), our method produces artifact-free defocus effects under various pupil settings as opposed to existing approaches which produces noticeable ringing artifacts, subtle defocus effects or noise in out-of-focus areas.
}
\label{figure:comp_wild}
\end{figure*}https://www.overleaf.com/project/632bf5ca9167c44f222e45cd
\section{Dataset and Implementation}

In this section, we provide implementation details of the proposed framework, including the dataset and algorithm implementation. 
\paragraph{Dataset} 
For the training of our proposed framework, we generated a comprehensive dataset using \href{https://www.blender.org}{Blender}\footnote{https://www.blender.org}. This dataset consists of 3,200 photorealistic data samples, each of which includes an all-in-focus RGB image coupled with a corresponding depth map. Additionally, distinct varifocal image stacks are rendered for each sample, each associated with a different aperture settings. 
These scenes were constructed using objects randomly selected from an object pool, which we curated from several datasets \cite{hodan2017tless, hodan2018bop, kaskman2019homebreweddb, hodan2020bop}. 
In these scenes, object positions, rotations, material properties and textures were arbitrarily assigned, with the textures sampled from the texture library \cite{cc0_texture}. 
To ensure that defocus effects across different focal planes and aperture settings are easily distinguishable, we introduced colorful polygons onto the textures before rendering. This especially enhanced the visibility of defocus effects in areas with uniform color or subtle textures. 
The training data was rendered at a resolution of $512 \times 512$, while the test data was rendered at two different resolutions, $512 \times 512$ and $1920 \times 1080$, each consisting of 100 samples.

\revise{
%
}
For each aperture setting, the number of focal planes is determined by considering the maximum blur size across the entire focal stack, ensuring that the difference in the circle of confusion size at a given spatial position between any two adjacent focal planes is less than $1$ pixel.
For instance, we rendered 28, 46, and 55 focal planes for pupil sizes of 2~mm, 3~mm, and 4~mm, respectively, with focal distances evenly distributed in diopter units. 
To test the generalization capacity of our trained framework, we rendered the test data with natural textures. 
For visual examples of the training data, please refer to the Supplementary Material.

\paragraph{Implementation}
The proposed framework is implemented using \href{https://pytorch.org}{PyTorch}\footnote{https://pytorch.org} library \cite{pytorch}. 
For the training phase, we configured a batch size of $1$ and a learning rate of $0.0001$.
During each training iteration, a scene was randomly sampled, and a pupil size parameter was chosen.
The corresponding all-in-focus RGB-D data and the complete focal stack, matching the selected pupil size, were then used as input and target data, respectively.
To calculate $\mathcal{L}_\text{rec}$, we employed all the focal images within the entire focal stack to supervise the reconstruction of the corresponding images.
When calculating $\mathcal{L}_\text{perp}$, to optimize memory consumption, we applied perceptual supervision to only $5$ randomly selected reconstruction planes in each iteration.
The full framework and all its ablation study variants were trained for $30$ epochs on an NVIDIA Tesla V100 with $32$ GB GPU, taking approximately $30$ hours.
Details of the simulation parameters for all the variants and the hardware display prototype are provided in the Supplementary Material.

\section{Results and Analysis}
In this section, we conduct an extensive analysis of the proposed framework. Section 6.1 presents a qualitative comparison and analysis of our approach against various state-of-the-art methods. In Section 6.2, we offer a comprehensive examination of the design choices in our approach. Lastly, in Section 6.3, we delve into the framework's generalization capability to diverse pupil settings, encompassing those not encountered during the training phase and sizes beyond the training range.


\subsection{Comparative Analysis} \label{sec:comparison}


In this section, we perform a comparative analysis of simulated reconstructions resulting from holograms generated by state-of-the-art 3D holography algorithms, which include TensorHolo v1 \cite{shi2021nature}, TensorHolo v2 \cite{shi2022light}, and Neural3D holography iterative method by Choi et al. \shortcite{choi2021neural3d}. As these methods very visibly neglect defocus effects in out-of-focus areas, our comparisons focus on qualitative assessments using datasets comprising both rendered and real-world in-the-wild data. Additionally, the Supplementary Material includes qualitative outcomes from two other recent works by Kavaklı et al. \shortcite{realistic_blur} and Lee et al. \shortcite{b-sgd}, both of which account for incoherent defocus blur. Notably, both the methods propose distinct approaches for creating ground truth focal images with incoherent blur, rather than relying on rendered or captured focal stacks, warranting an evaluation. All results from the various methods are based on the implementations provided by their respective authors.

\subsubsection{Results on rendered data}
In \Cref{figure:comp512} and \Cref{figure:comp_1080}, we present selected outcomes from simulated reconstructions generated by different methods alongside the corresponding rendered focal images for reference. 
As illustrated in \Cref{figure:comp512}, TensorHolo v1 exhibits modest cues for near and far focal planes but is plagued by pronounced ringing artifacts around the edges of out-of-focus areas, notably visible on the striped regions of the white mug. 
TensorHolo v2 shows some improvement in mitigating ringing effects, although these artifacts are still discernible in striped sections and the borders of the gray cup.
\setlength{\tabcolsep}{2pt}
\renewcommand{\arraystretch}{0.6}
\begin{figure*}[t]
\begin{center}
\small
\begin{tabular}{cccl}
& \textbf{Ours (2mm Pupil)} & \textbf{Ours (3mm Pupil)} &\textbf{\:\:\:\:\:\:\:\:\:\:\:\:\:\:\:\:\:\:\:\:\:\:\:\:\:\:Ours (4mm Pupil)}   \\ [0.5ex]
\raisebox{0.45\height}{\rotatebox{90}{\textbf{Far Focus}}}
&\includegraphics[width=5.7cm]{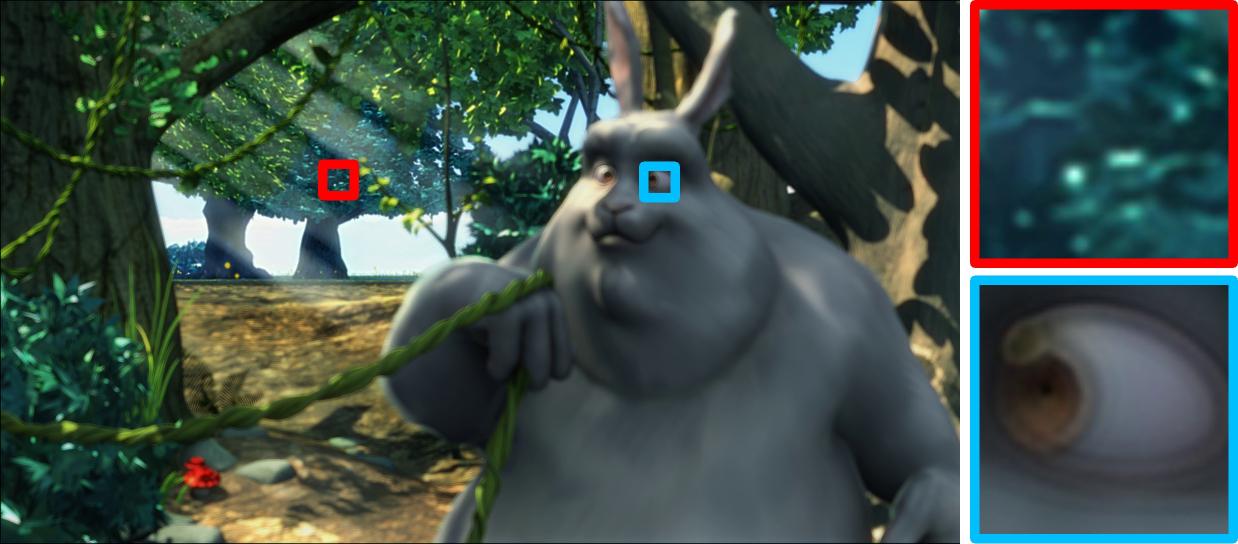}
&\includegraphics[width=5.7cm]{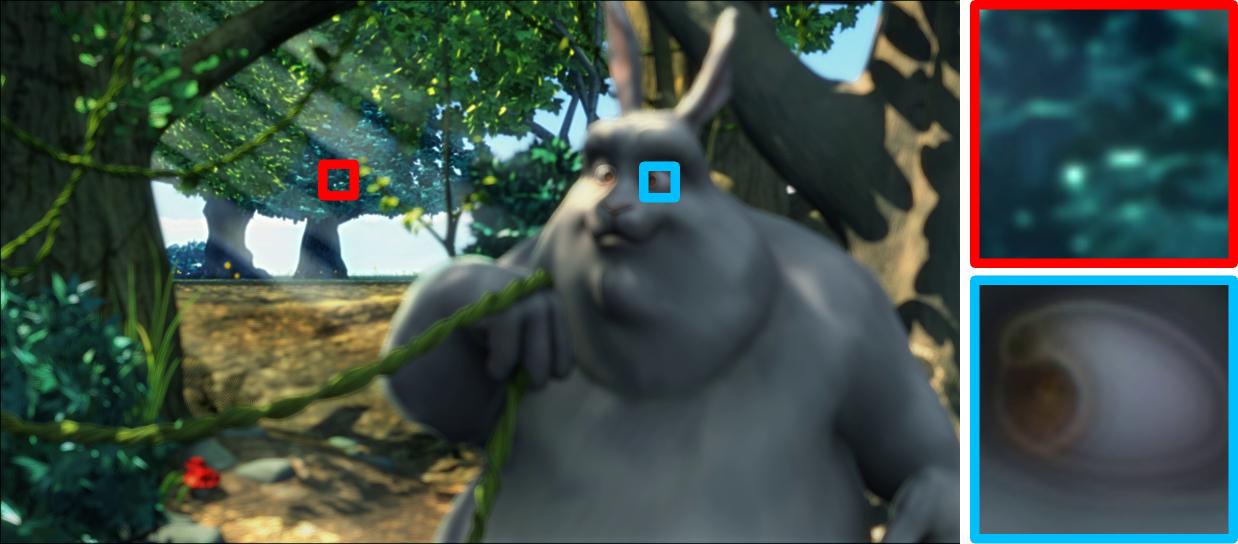}
&\includegraphics[width=5.7cm]{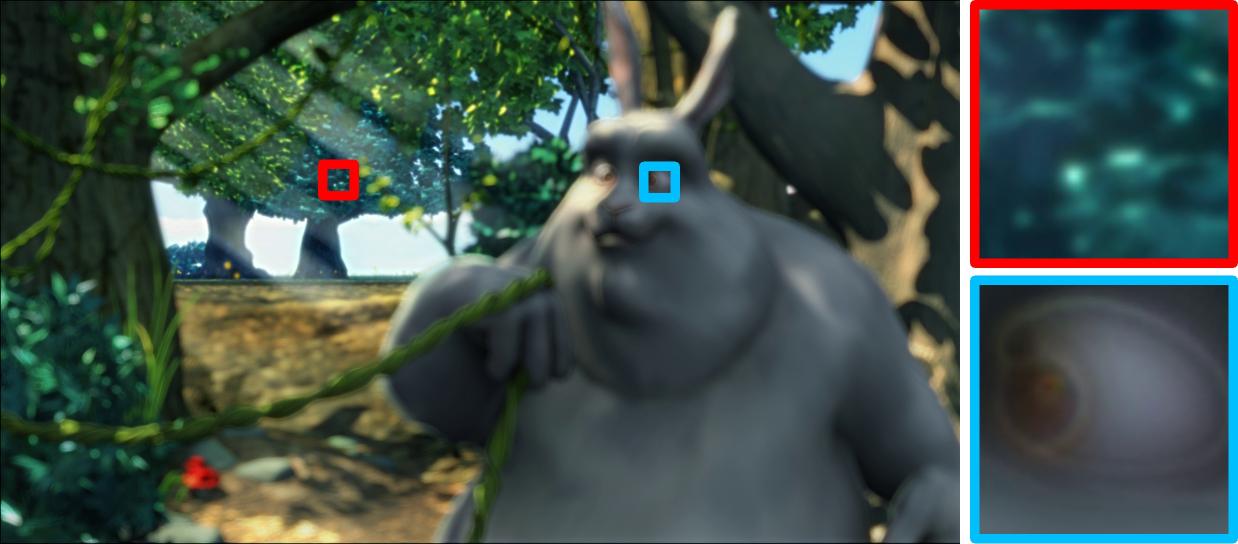}
\\
\raisebox{0.25\height}{\rotatebox{90}{\textbf{Near Focus}}}
&\includegraphics[width=5.7cm]{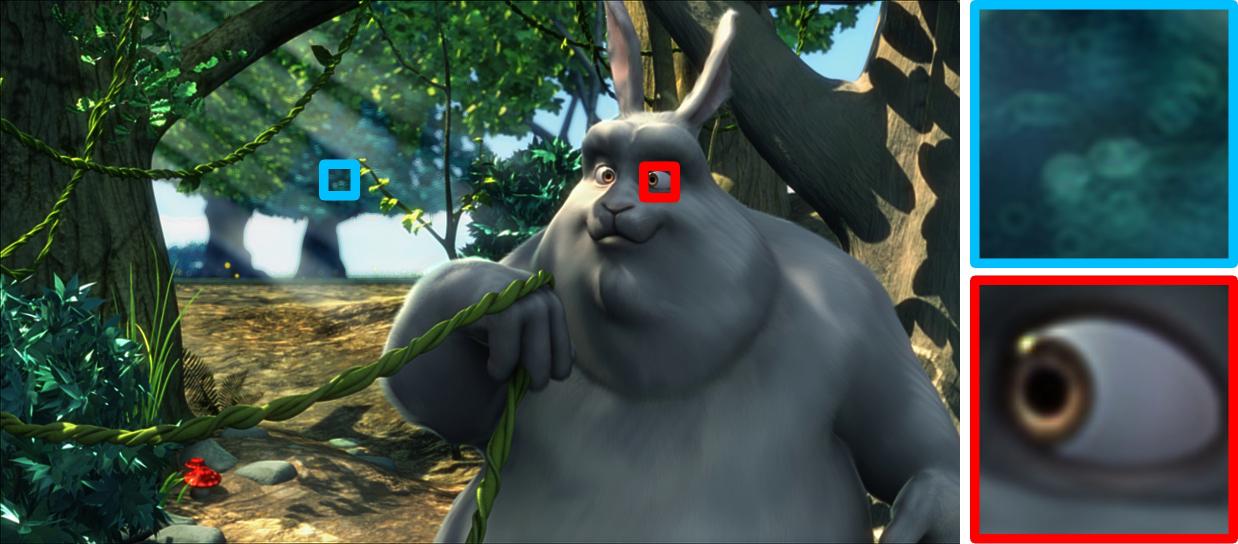}
&\includegraphics[width=5.7cm]{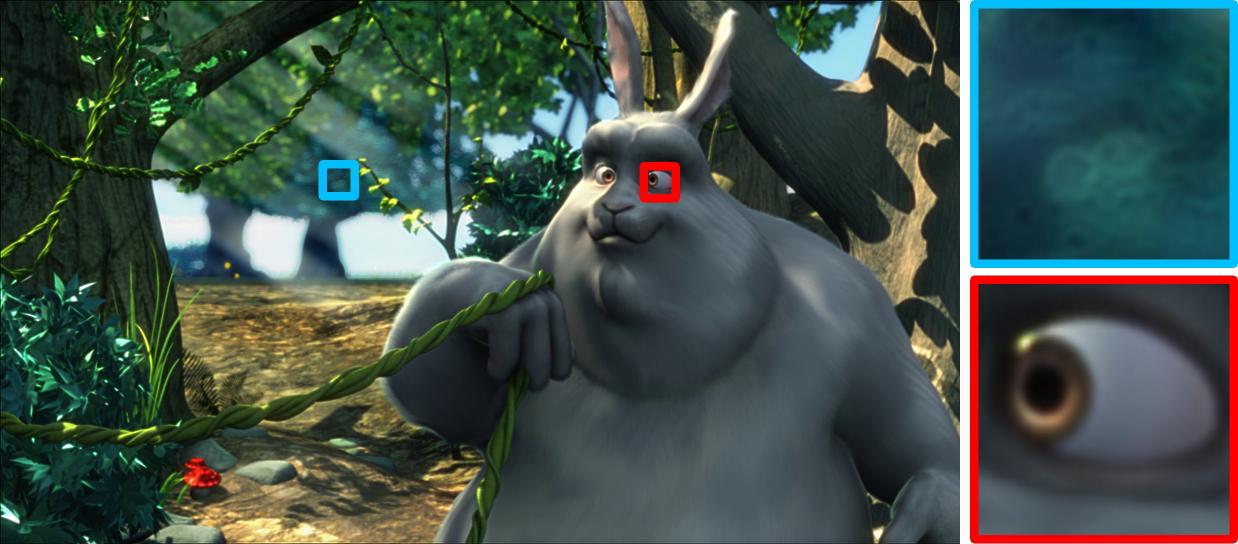}
&\includegraphics[width=5.7cm]{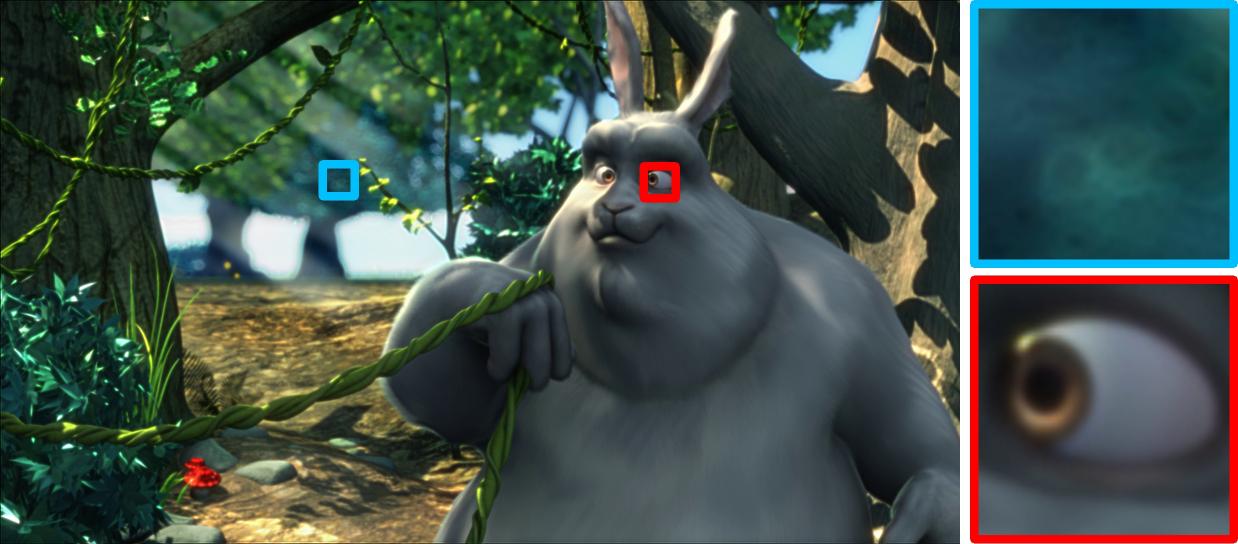}
\\[0.7ex]
& \textbf{Shi et al. 2021} & \textbf{Shi et al. 2022} & \textbf{\:\:\:\:\:\:\:\:\:\:\:\:\:\:\:\:\:\:\:\:\:\:\:\:\:\:\:\:\:\:Choi et al. 2021} \\ [0.7ex]
\raisebox{0.4\height}{\rotatebox{90}{\textbf{Far Focus}}}
&\includegraphics[width=5.7cm]{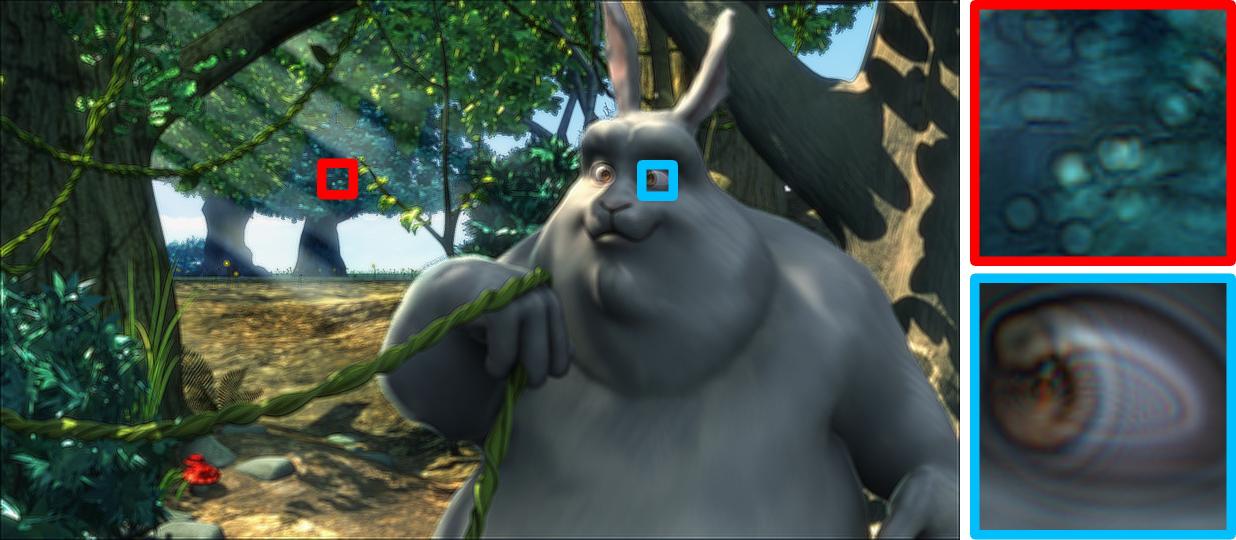}
&\includegraphics[width=5.7cm]{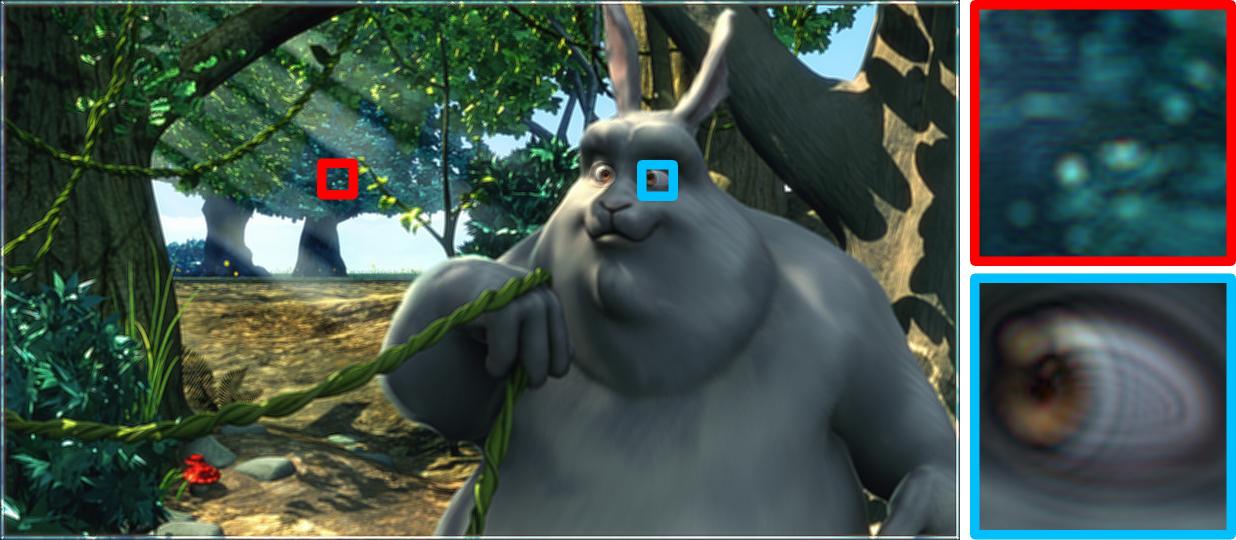}
&\includegraphics[width=5.7cm]{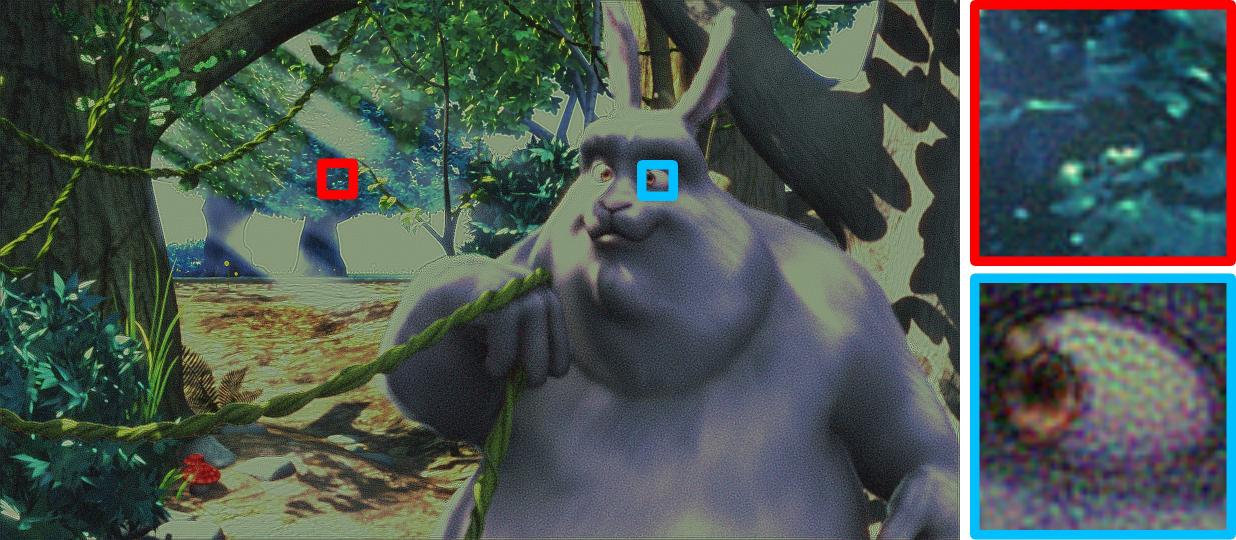}
\\
\raisebox{0.25\height}{\rotatebox{90}{\textbf{Near Focus}}}
&\includegraphics[width=5.7cm]{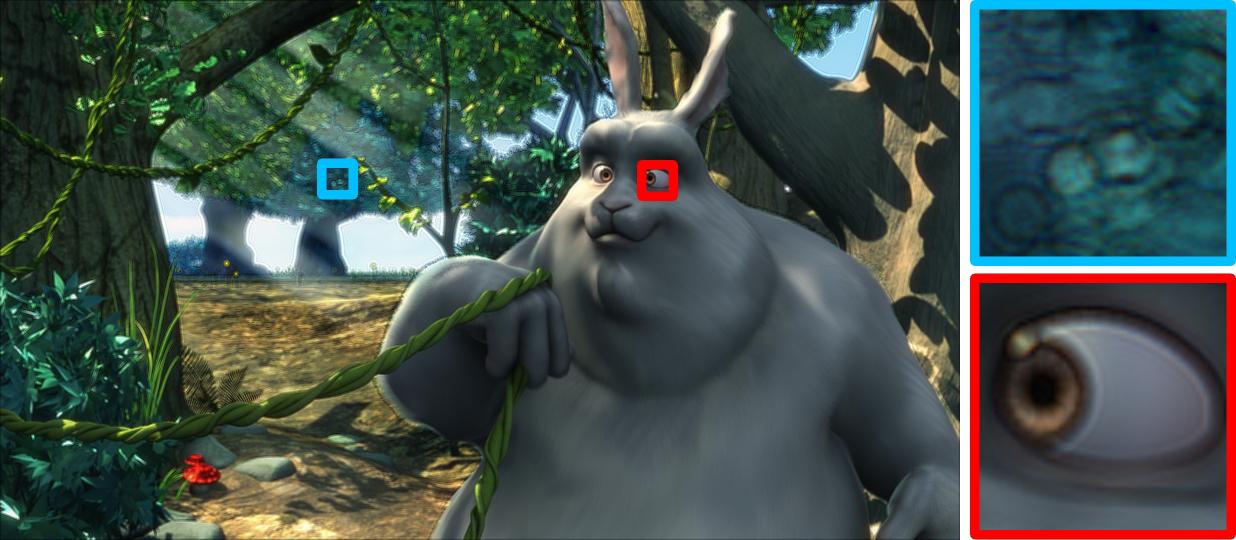}
&\includegraphics[width=5.7cm]{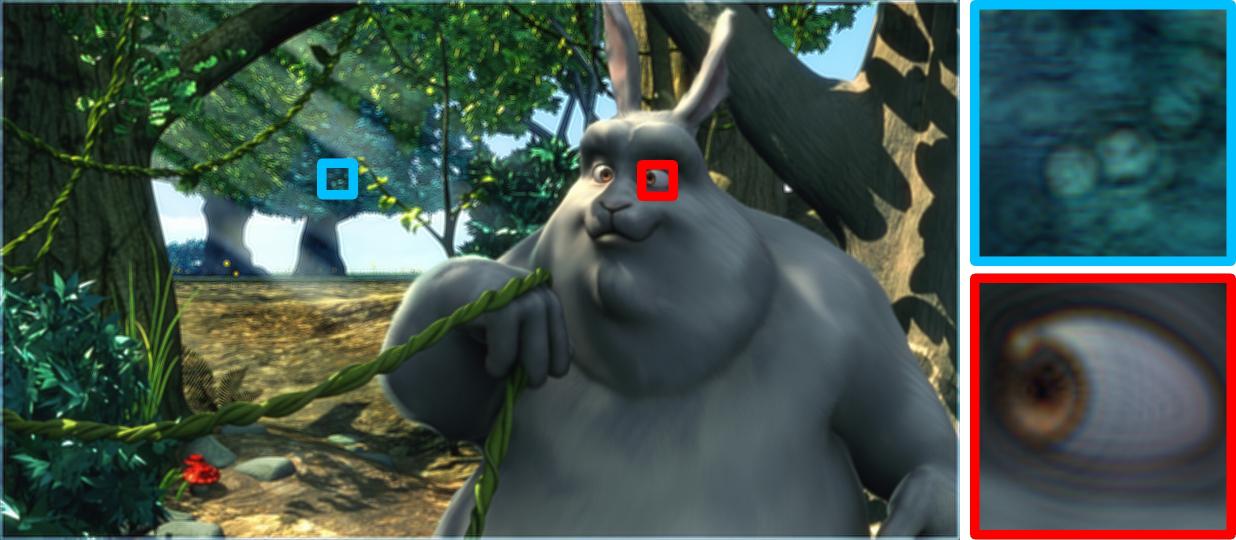}
&\includegraphics[width=5.7cm]{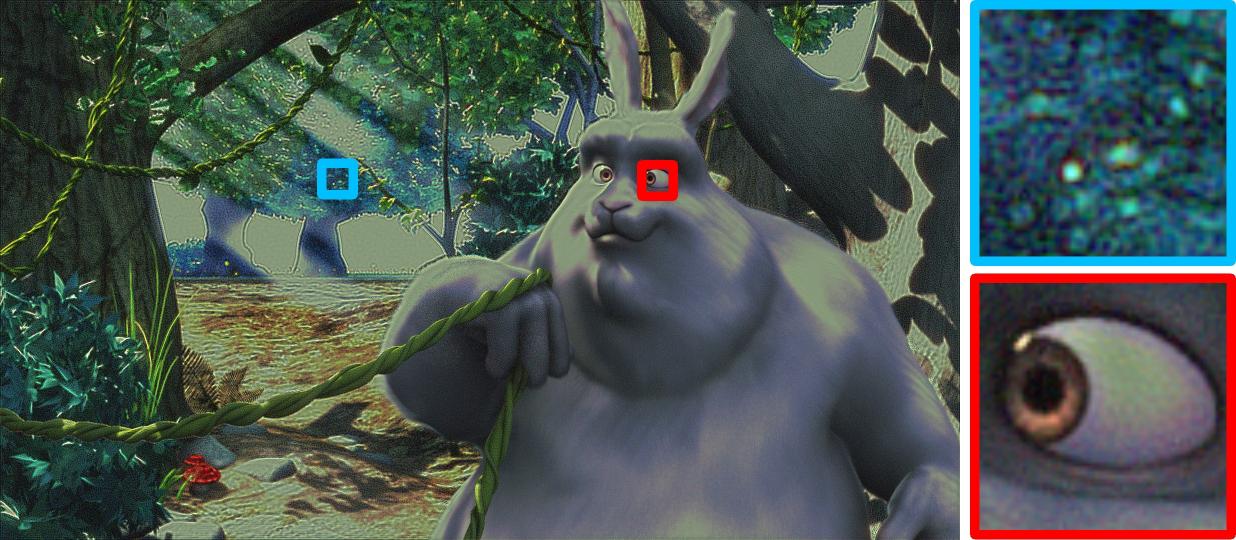}
\\
\end{tabular}
\end{center}
\caption{Simulated reconstruction from holograms obtained by different methods on in-the-wild data. In-focus regions are highlighted in red and out-of-focus regions are marked in blue. Only one group of results is provided for methods that do not consider varying pupil sizes and corresponding depth-of-field effects. As the highlighted regions (the distant leaves the closer eye area) show, our results present desirable defocus effects under various pupil conditions, while other methods produce ringing artifacts or noise in out-of-focus regions.}
\label{figure:comp_wild3}
\end{figure*}
\setlength{\tabcolsep}{3.15pt}
\renewcommand{\arraystretch}{1.3}
\begin{table*}[htbp]
\begin{center}
\caption{Performance of the proposed framework}\label{tab:performance}
\small
\begin{tabular}{|l|cccccccccccc|llllllllllll}
\cline{1-13}
\multirow{3}{*}{Variant\:\:\:\:\:\:\:\:\:\:\:\:\:\:\:\:\:\:\:\:} & \multicolumn{12}{c|}{Data$-512\times512$}                                                                                                                                                                                                                                                                          &  &  &  &  &  &  &  &  &  &  &  &  \\ \cline{2-13}
                         & \multicolumn{4}{c|}{$s=$2 mm}                                                                                         & \multicolumn{4}{c|}{$s=$3 mm}                                                                                         & \multicolumn{4}{c|}{$s=$4 mm}                                                                    &  &  &  &  &  &  &  &  &  &  &  &  \\ \cline{2-13}
                         & \multicolumn{1}{c|}{PSNR} $\uparrow$ & \multicolumn{1}{c|}{SSIM} $\uparrow$ & \multicolumn{1}{c|}{LPIPS (v)}$\downarrow$ & \multicolumn{1}{c|}{LPIPS (a)} $\downarrow$ & \multicolumn{1}{c|}{PSNR} $\uparrow$& \multicolumn{1}{c|}{SSIM} $\uparrow$ & \multicolumn{1}{c|}{LPIPS (v)} $\downarrow$ & \multicolumn{1}{c|}{LPIPS (a)} $\downarrow$& \multicolumn{1}{c|}{PSNR} $\uparrow$ & \multicolumn{1}{c|}{SSIM} $\uparrow$ & \multicolumn{1}{c|}{LPIPS (v)} $\downarrow$\textbf{} & LPIPS (a) $\downarrow$ &  &  &  &  &  &  &  &  &  &  &  &  \\ \cline{1-13}
Ours     & \multicolumn{1}{c|}{29.1123}    & \multicolumn{1}{c|}{0.9536}    & \multicolumn{1}{c|}{0.1176}     & \multicolumn{1}{c|}{0.0663}       & \multicolumn{1}{c|}{27.4647}    & \multicolumn{1}{c|}{0.9308}    & \multicolumn{1}{c|}{0.1627}     & \multicolumn{1}{c|}{0.0990}       & \multicolumn{1}{c|}{26.2582}    & \multicolumn{1}{c|}{0.9126}    & \multicolumn{1}{c|}{0.2039}     & 0.1311       &  &  &  &  &  &  &  &  &  &  &  &  \\\cline{1-13}
Ours (FS)    & \multicolumn{1}{c|}{\textbf{29.7798}}    & \multicolumn{1}{c|}{\textbf{0.9595}}    & \multicolumn{1}{c|}{\textbf{0.1067}}     & \multicolumn{1}{c|}{\textbf{0.0607}}       & \multicolumn{1}{c|}{\textbf{28.0163}}    & \multicolumn{1}{c|}{\textbf{0.9378}}    & \multicolumn{1}{c|}{\textbf{0.1525}}     & \multicolumn{1}{c|}{\textbf{0.0919}}       & \multicolumn{1}{c|}{\textbf{26.7930}}    & \multicolumn{1}{c|}{\textbf{0.9204}}    & \multicolumn{1}{c|}{\textbf{0.1918}}     & \textbf{0.1206}       &  &  &  &  &  &  &  &  &  &  &  &  \\ \cline{1-13}
\multicolumn{1}{|c}{}     & \multicolumn{12}{c|}{Data\emph{$-1920\times1080$}}                                                                                                                                                                                                                                                                                                    &  &  &  &  &  &  &  &  &  &  &  &  \\ \cline{1-13}
Ours        & \multicolumn{1}{c|}{34.0320}    & \multicolumn{1}{c|}{0.9813}    & \multicolumn{1}{c|}{0.0745}     & \multicolumn{1}{c|}{0.0372}       & \multicolumn{1}{c|}{32.1694}    & \multicolumn{1}{c|}{0.9710}    & \multicolumn{1}{c|}{0.1083}     & \multicolumn{1}{c|}{0.0585}       & \multicolumn{1}{c|}{30.7317}    & \multicolumn{1}{c|}{0.9615}    & \multicolumn{1}{c|}{0.1406}     & 0.0799      &  &  &  &  &  &  &  &  &  &  &  &  \\ \cline{1-13}
Ours (FS)        & \multicolumn{1}{c|}{\textbf{34.7735}}    & \multicolumn{1}{c|}{\textbf{0.9838}}    & \multicolumn{1}{c|}{\textbf{0.0654}}     & \multicolumn{1}{c|}{\textbf{0.0342}}       & \multicolumn{1}{c|}{\textbf{32.7494}}    & \multicolumn{1}{c|}{\textbf{0.9741}}    & \multicolumn{1}{c|}{\textbf{0.0981}}     & \multicolumn{1}{c|}{\textbf{0.0543}}       & \multicolumn{1}{c|}{\textbf{31.2619}}    & \multicolumn{1}{c|}{\textbf{0.9650}}    & \multicolumn{1}{c|}{\textbf{0.1287}}     & \textbf{0.0736}       &  &  &  &  &  &  &  &  &  &  &  &  \\ \cline{1-13}
\end{tabular}
\smallskip
\begin{tablenotes}
\footnotesize
\item[] * {LPIPS (v) and LPIPS (a) denote the perceptual metrics that compare the features extracted for results and ground truth images from VGG network and AlexNet respectively.}
\end{tablenotes}
\vspace{5mm}
\end{center}
\end{table*}

The results from Neural3D holography, on the other hand, exhibit noise in the out-of-focus areas. 
In contrast, our method yields smoother blur effects in both striped regions of far-focus planes and the borders of the gray cup in near-focus planes, approaching the quality seen in rendered data.
Furthermore, the results obtained using our approach demonstrate varying blur effects for out-of-focus areas with changes in pupil size.
In \Cref{figure:comp_1080}, we report results when the image resolution is increased to $1920 \times 1080$. Here, defocus effects generated by holograms from TensorHolo v1 and TensorHolo v2 become less prominent, especially in the case of TensorHolo v2. 
Notable noise reappears in the out-of-focus areas of Neural3D holography results. 
Conversely, the reconstructed images from our method exhibit more natural defocus blur, displaying a desirable, adaptable trend in response to changes in pupil size.

\paragraph{Discussion}
The pronounced ringing artifacts observed in results obtained from TensorHolo v1 stem from inaccuracies in occlusion modeling during the training data preparation.
Specifically, the ground truth holograms in TensorHolo v1 \cite{shi2021nature} are derived from RGB-D data using a point-based method with occlusion detection. This method inaccurately handles the wavefront from occluded parts, as discussed in Chakravarthula et al. \shortcite{chakravarthula2022hogel}.
%
The method TensorHolo v2, which is trained with ground truth holograms calculated from layered depth images (LDI), providing better occlusion consideration, exhibits reduced ringing artifacts. However, because LDIs are sampled discretely during hologram computation, inaccuracies in occlusion modeling in certain areas might contribute to the remaining ringing artifacts in TensorHolo v2 results.
%
The noise present in Neural3D holograms \cite{choi2021neural3d} results from the random object phase that stems from the optimization process. Although adopting a time-multiplexed solution can significantly reduce this noise, it comes at the cost of increased computation and time since it requires generating numerous holograms.
As shown in the Supplementary Material, the method by Lee et al. \shortcite{b-sgd} achieves smooth defocus blur when the number of multiplexed frames is increased. The solution presented in Kavaklı et al. \shortcite{realistic_blur} also produces reasonable defocus blur, but it exhibits less noticeable variations when the maximum blur kernel size changes.

In contrast to all the above described methods 
our framework ensures speckle mitigation and high image quality while realizing more realistic pupil-dependenet defocus effects. However, slight disparities still remain in our results due to the absence of occlusion information in the input RGB-D data. 
Furthermore, we also test our framework, trained purely on the synthetic dataset, on real world in-the-wild test data. Specifically, we use our trained networks on $1920 \times 1080$ in-the-wild data \emph{without any finetuning on real data} to assess its generalization ability, and present results in \Cref{figure:comp_wild} and \Cref{figure:comp_wild3}. As can be seen, the reconstructed images exhibit defocus effects on par with that obtained using synthetic test data, demonstrating our method's ability to \emph{generalize to cross-domain inference}. Notably, observe that ringing artifacts appear noticeably in results from TensorHolo v1 and the contrast between the in-focus and out-of-focus areas in the results from TensorHolo v2 are still inferior, which are significantly overcome with our method.

\subsection{Framework Design Choices and Analysis} \label{sec:ablation}
In this section, we quantitatively evaluate the proposed framework and investigate the efficacy of key design choices, such as the adjustable deformable convolution and the proposed training approach.
Specifically, our evaluation focuses on comparing the reconstructed images generated from holograms produced by different variants of our method against the rendered focal stack images.
To assess the results, we employ four widely used image metrics, including PSNR (peak signal-to-noise ratio), SSIM (structural similarity index measure), and two perceptual metrics based on neural features, LPIPS (v) and LPIPS (a), which rely on image features extracted by VGG \cite{vgg} and AlexNet \cite{AlexNet}, respectively. 
Additionally, we provide a statistical analysis of the learned offsets in the ADC layers in \Cref{sec:offsets} and investigate the impact of network capacity by comparing the performance of various variants with different capacities in \Cref{sec:capacity}. 
In \Cref{sec:runtime}, we report and discuss the runtime efficiency of different methods.
Further ablation studies on the loss terms are detailed in the Supplementary Material. 

\subsubsection{Quantitative Performance}
We conducted a quantitative evaluation of our frameworks, specifically \emph{Ours} and \emph{Ours(FS)}, and the results are summarized in \Cref{tab:performance}.
The table indicates that both \emph{Ours} and \emph{Ours(FS)} yield commendable numerical results, particularly on test data with a resolution of $1920\times1080$.
However, when dealing with data at a resolution of $512\times512$, the quantitative metrics show a degradation in performance. This can be attributed to the increased prominence of defocus blur relative to the image as the pupil size grows, occupying a larger portion of the entire image and affecting the reconstructed images.
%
Notice that the variant \emph{Ours(FS)} demonstrates noticeable improvements across all metrics. This enhancement is primarily due to the input focal images providing more information regarding occluded parts of the 3D scene, resulting in improved quality. These findings are further supported by the qualitative results presented in Section S7.1 of the Supplementary Material. Since the performance trends among different pupil sizes are similar for each variant, we focus on investigating their performance on data with a resolution of $512\times512$ hereon.

\setlength{\tabcolsep}{2.65pt}
\renewcommand{\arraystretch}{1.15}
\begin{table*}[htbp]
\begin{center}
\caption{Performance of different hologram generation network variants evaluated on reconstructed image quality.}\label{tab:ablation}
\small
\begin{tabular}{|l|cccccccccccc|llllllllllll}
\cline{1-13}
\multirow{3}{*}{Variant} & \multicolumn{12}{c|}{Data$-512\times512$}                                                                                                                                                                                                                                                                          &  &  &  &  &  &  &  &  &  &  &  &  \\ \cline{2-13}
                         & \multicolumn{4}{c|}{$s=2$ mm}                                                                                         & \multicolumn{4}{c|}{$s=3$ mm}                                                                                         & \multicolumn{4}{c|}{$s=4$ mm}                                                                    &  &  &  &  &  &  &  &  &  &  &  &  \\ \cline{2-13}
                         & \multicolumn{1}{c|}{PSNR} $\uparrow$ & \multicolumn{1}{c|}{SSIM} $\uparrow$ & \multicolumn{1}{c|}{LPIPS (v)}$\downarrow$ & \multicolumn{1}{c|}{LPIPS (a)} $\downarrow$ & \multicolumn{1}{c|}{PSNR} $\uparrow$& \multicolumn{1}{c|}{SSIM} $\uparrow$ & \multicolumn{1}{c|}{LPIPS (v)} $\downarrow$ & \multicolumn{1}{c|}{LPIPS (a)} $\downarrow$& \multicolumn{1}{c|}{PSNR} $\uparrow$ & \multicolumn{1}{c|}{SSIM} $\uparrow$ & \multicolumn{1}{c|}{LPIPS (v)} $\downarrow$\textbf{} & LPIPS (a) $\downarrow$ &  &  &  &  &  &  &  &  &  &  &  &  \\ \cline{1-13}
Ours (3 nets (conv.))        & \multicolumn{1}{c|}{28.8226}    & \multicolumn{1}{c|}{0.9494}    & \multicolumn{1}{c|}{0.1259}     & \multicolumn{1}{c|}{0.0700}       & \multicolumn{1}{c|}{26.9639}    & \multicolumn{1}{c|}{0.9238}    & \multicolumn{1}{c|}{0.1805}     & \multicolumn{1}{c|}{0.1096}       & \multicolumn{1}{c|}{25.7832}    & \multicolumn{1}{c|}{0.9045}    & \multicolumn{1}{c|}{0.2243}     & 0.1448       &  &  &  &  &  &  &  &  &  &  &  &  \\ \cline{1-13}
Ours (3 nets)        & \multicolumn{1}{c|}{28.9137}    & \multicolumn{1}{c|}{0.9517}    & \multicolumn{1}{c|}{0.1207}     & \multicolumn{1}{c|}{0.0682}       & \multicolumn{1}{c|}{27.2649}    & \multicolumn{1}{c|}{0.9280}    & \multicolumn{1}{c|}{0.1700}     & \multicolumn{1}{c|}{0.1041}       & \multicolumn{1}{c|}{26.0877}    & \multicolumn{1}{c|}{0.9109}    & \multicolumn{1}{c|}{0.2124}     & 0.1365      &  &  &  &  &  &  &  &  &  &  &  &  \\ \cline{1-13}
Ours (modulated conv)    & \multicolumn{1}{c|}{28.4165}    & \multicolumn{1}{c|}{0.9476}    & \multicolumn{1}{c|}{0.1330}     & \multicolumn{1}{c|}{0.0752}       & \multicolumn{1}{c|}{27.0237}    & \multicolumn{1}{c|}{0.9246}    & \multicolumn{1}{c|}{0.1790}     & \multicolumn{1}{c|}{0.1088}       & \multicolumn{1}{c|}{25.8820}    & \multicolumn{1}{c|}{0.9054}    & \multicolumn{1}{c|}{0.2217}     &   0.1414   &  &  &  &  &  &  &  &  &  &  &  &  \\ \cline{1-13}
Ours (DEH)    & \multicolumn{1}{c|}{28.3288}    & \multicolumn{1}{c|}{0.9457}    & \multicolumn{1}{c|}{0.1341}   & \multicolumn{1}{c|}{0.0763}       & \multicolumn{1}{c|}{26.7164}    & \multicolumn{1}{c|}{0.9208}   & \multicolumn{1}{c|}{0.1863}     & \multicolumn{1}{c|}{0.1135}       & \multicolumn{1}{c|}{25.5121}    & \multicolumn{1}{c|}{0.9016}    & \multicolumn{1}{c|}{0.2303}   & {0.1485}      &  &  &  &  &  &  &  &  &  &  &  &  \\ \cline{1-13}
Ours     & \multicolumn{1}{c|}{\textbf{29.1123}}    & \multicolumn{1}{c|}{\textbf{0.9536}}    & \multicolumn{1}{c|}{\textbf{0.1176}}     & \multicolumn{1}{c|}{\textbf{0.0663}}       & \multicolumn{1}{c|}{\textbf{27.4647}}    & \multicolumn{1}{c|}{\textbf{0.9308}}    & \multicolumn{1}{c|}{\textbf{0.1627}}     & \multicolumn{1}{c|}{\textbf{0.0990}}       & \multicolumn{1}{c|}{\textbf{26.2582}}    & \multicolumn{1}{c|}{\textbf{0.9126}}    & \multicolumn{1}{c|}{\textbf{0.2039}}     & \textbf{0.1311}       &  &  &  &  &  &  &  &  &  &  &  &  \\ \cline{1-13}
\end{tabular}
\smallskip
\begin{tablenotes}
\footnotesize
\item[] * {All reported variants in this table receive RGB-D data as input.}
\end{tablenotes}
\vspace{2mm}
\end{center}
\end{table*}
\subsubsection{Architectural Effectiveness} \label{sec:analysis_arch}
Our framework, which incorporates the ADC layer, demonstrates promising results in accommodating defocus effects due to variations in pupil size. This adaptability minimizes the need for training separate networks for distinct pupil conditions, improving the framework's overall efficiency and flexibility.
We examine the effectiveness of our architectural design by comparing it to three variants, each of which is described below.

\paragraph{Ours vs. Separate Networks} 
To assess the effectiveness of our proposed unified framework, we conducted a comparison against a straightforward alternative, which involves training three distinct networks for the three different pupil sizes.
Specifically, we evaluated our framework against two variants.  
The first variant, labeled \emph{Ours (3 nets (conv.))}, employs three separate networks with standard convolutional layers, as adjustability is unnecessary for each individual network.
The second variant, labeled \emph{Ours (3 nets)}, employs three separate sub-networks implemented with the same architecture as our complete framework.
In both of these variants, each sub-network is trained independently on data with a single pupil size setting.
\begin{figure}[t!]
\includegraphics[width=0.48\textwidth]{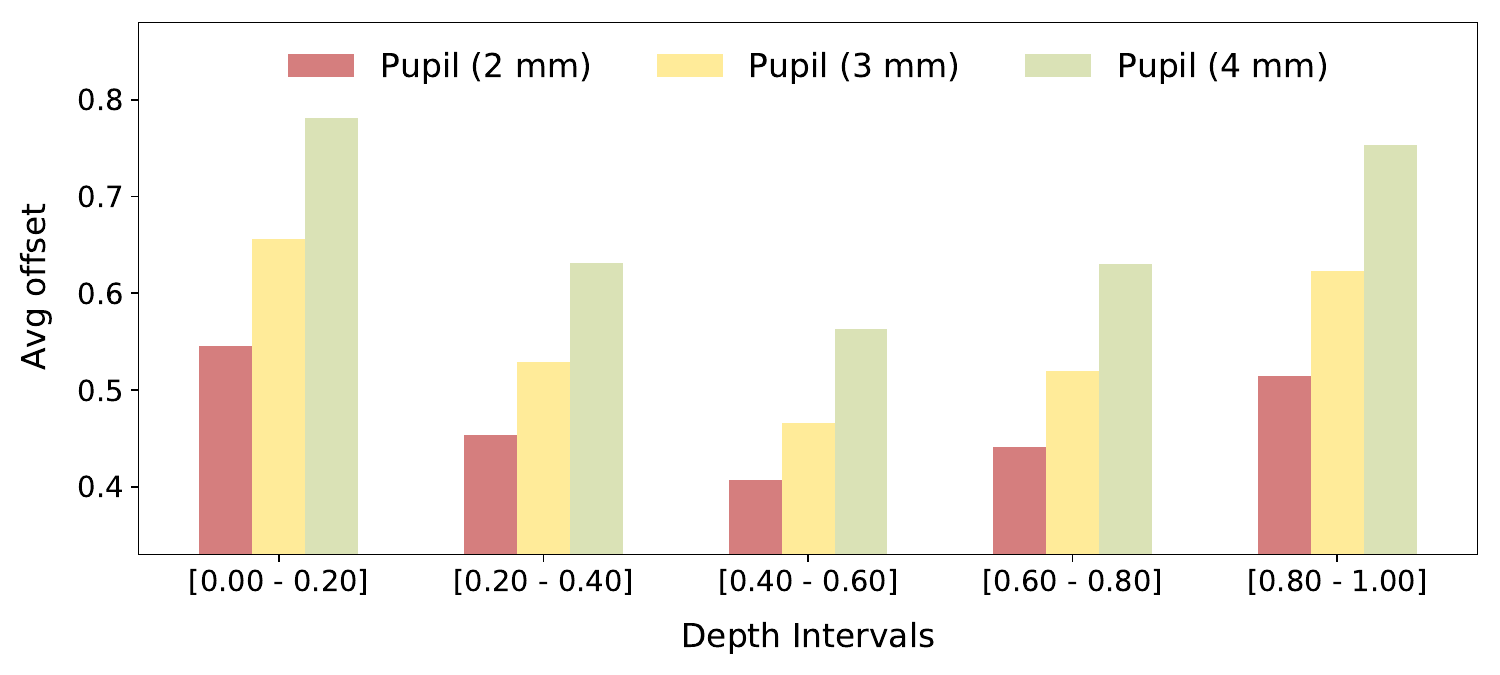}
\caption{Statistics of the learned offsets of the adjustable deformable convolution in the proposed framework. X-axis shows $5$ depth intervals and the values are normalized. Notably, the learned offsets are larger for near/far positions and smaller for middle positions, following the defocus trend.}
\label{fig:offset}
\end{figure}
To ensure fairness in the comparison, we trained these sub-networks for $10$ epochs, as our complete framework averaged $10$ epochs for each pupil parameter.

\Cref{tab:ablation} underscores a significant performance gap between the \emph{Ours (3 nets (conv.))} variant and our full model.
Notably, the LPIPS perceptual error \cite{LPIPS} values indicate a noticeable increase, highlighting perceptual dissimilarities in the images generated by \emph{Ours (3 nets (conv.))}.
Additionally, the performance gaps become more pronounced with larger pupil-size settings as can be seen in \Cref{tab:ablation}.
This is due to the restricted receptive field resulting from a fixed number of standard convolutional layers.

For \emph{Ours (3 nets)}, the performance surpasses that of \emph{Ours (3 nets (conv.))} variant.
Even when trained on three parameter settings together, our full model outperforms \emph{Ours (3 nets)}, demonstrating the effectiveness of the proposed ADC layer in utilizing priors present in data with several pupil settings.
Furthermore, our framework is more lightweight when compared to the scenario of employing three specialized sub-networks, overall indicating that the learnable offsets within the deformable convolutional layer are well-suited for the task of synthesizing 3D holograms to achieve spatially-varying defocus effects.

Furthermore, the variant \emph{Ours (DEH)} trains three sub-networks using the architecture proposed in Yang et al. \shortcite{deh2022}. This variant is akin to \emph{Ours (3 nets (conv.)} but with a reduced network capacity and is trained for $60$ epochs, following the specifications in Yang et al. \shortcite{deh2022}. As indicated in \Cref{tab:ablation}, the performance significantly falls behind the proposed method, primarily due to the constraints imposed by the limited network capacity.

\paragraph{Effectiveness of Deformable Convolution}
To assess the effectiveness of the newly introduced adjustable deformable convolutional layer, we conducted a comparison with a variant named \emph{Ours (modulated conv.)}, which employs the modulation layer introduced in a previous work \cite{he_conv}.
This modulation layer adjusts the per-channel feature scales in response to the convolutional layer's output. The scale is predicted using a Multi-Layer Perceptron (MLP) based on input control parameters.
The results, as presented in \cref{tab:ablation}, demonstrate that this variant, \emph{Ours (modulated conv.)}, yields a noticeable drop in performance across all four evaluation metrics, with a particularly significant decline in the LPIPS metrics. 
\setcounter{figure}{14}
\begin{figure}[t!]
\includegraphics[width=0.48\textwidth]{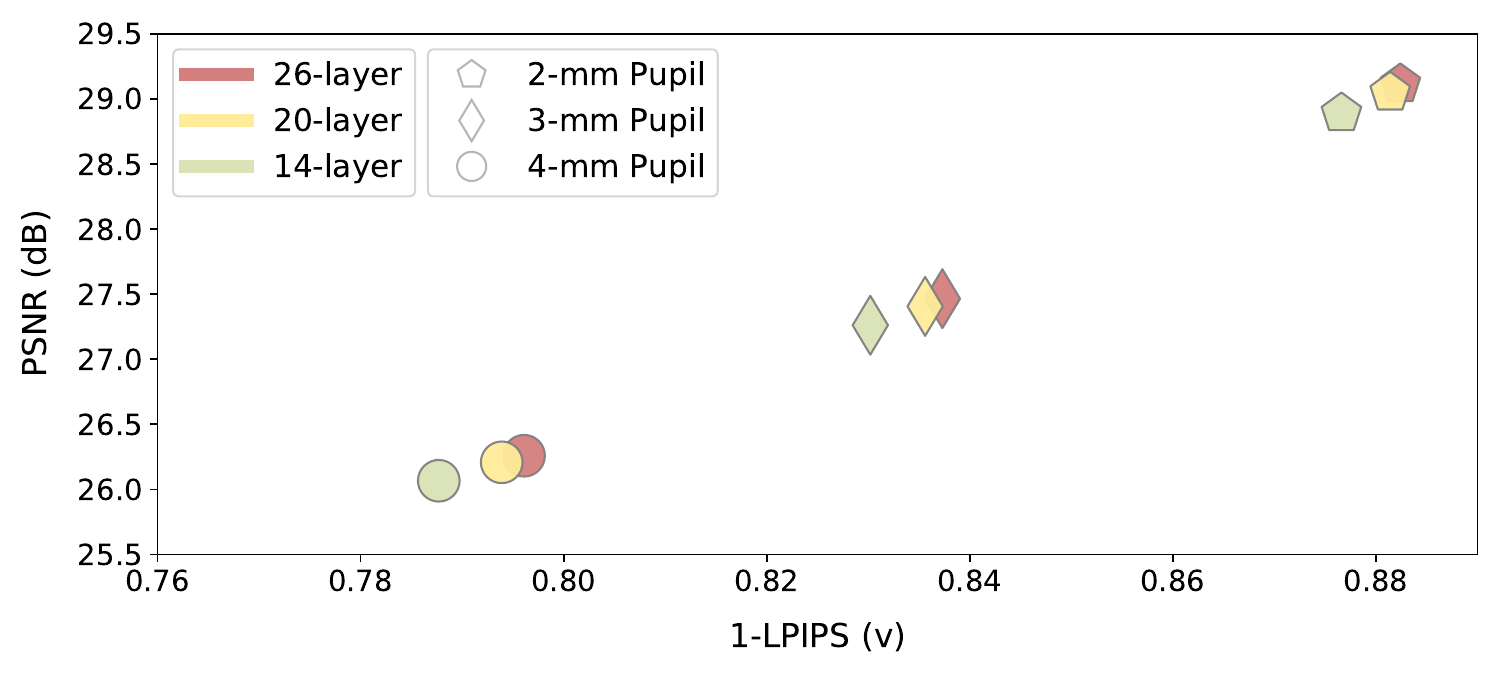}
\caption{Influence of the network capacity of the proposed framework. X-axis shows values for `1-LPIPS (v)', the larger the better. The size of the bubbles denotes the pupil condition for clarity. While the 26-layer version performs the best, two shallower variants achieve similar performance with acceptable performance degradation. 
}
\label{fig:capacity}
\end{figure}
Surprisingly, this variant's performance is even worse than the variant \emph{Ours (3 nets (conv.))} on data with a pupil size of $2$ mm.
This suggests that the modulated layer struggles to fully leverage the advantages of joint training with multiple settings.
This could be attributed to the modulation layer's global scaling of intermediate features on a per-channel basis, which may not effectively capture the spatially varying characteristics of each position's contribution area.

\setlength{\tabcolsep}{0.5pt}
\setcounter{figure}{15}
\renewcommand{\arraystretch}{0.6}
\begin{figure*}[htbp!]
\begin{center}
\small
\begin{tabular}{ccccccccccl}
&\:& s=1.0 & s=1.5 & s=2.0 & s=2.5 & s=3.0 & s=3.5 & s=4.0 & s=4.5 & \:\:\:\:\:\:s=5.0 \\[0.5ex]
\raisebox{0.45\height}{\rotatebox{90}{\textbf{Far Focus}}}
&\:&\includegraphics[height=2.73cm]{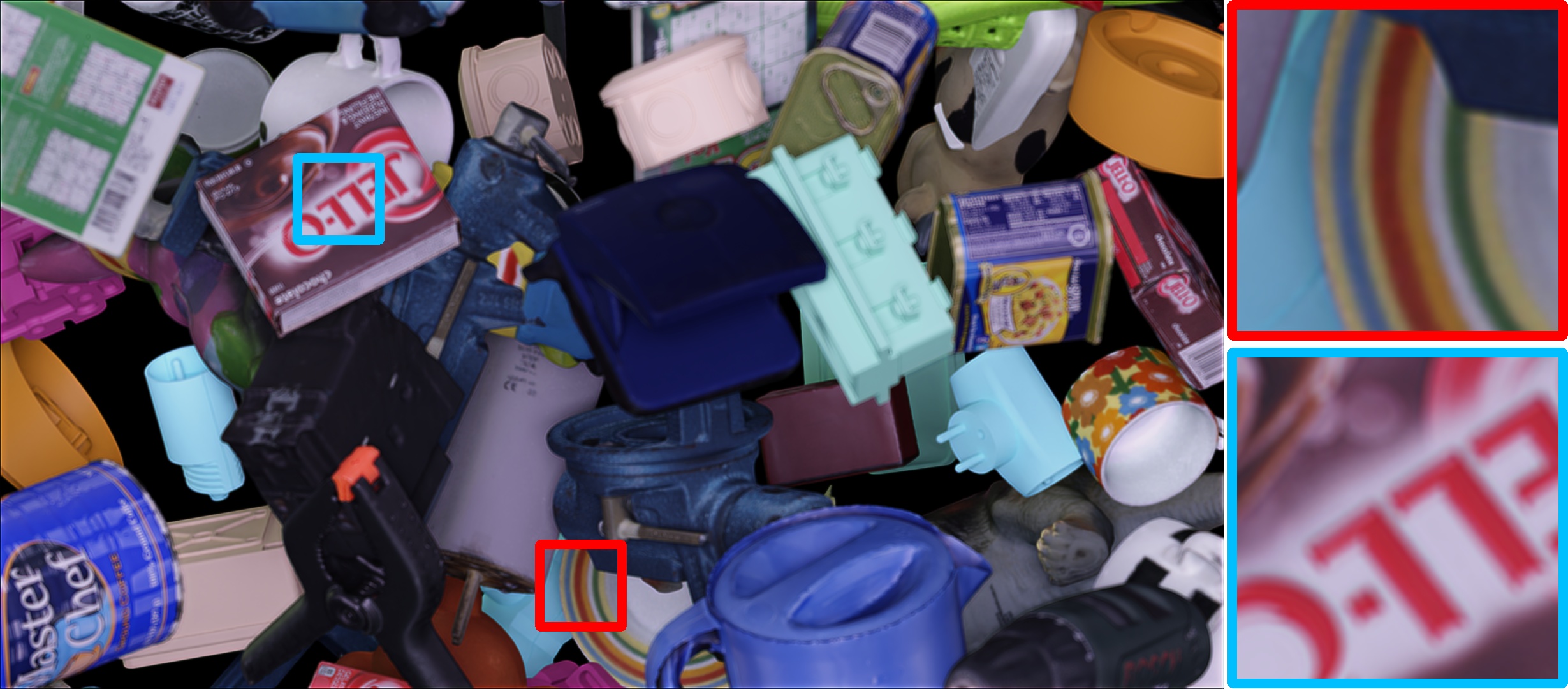}
&\includegraphics[height=2.73cm]{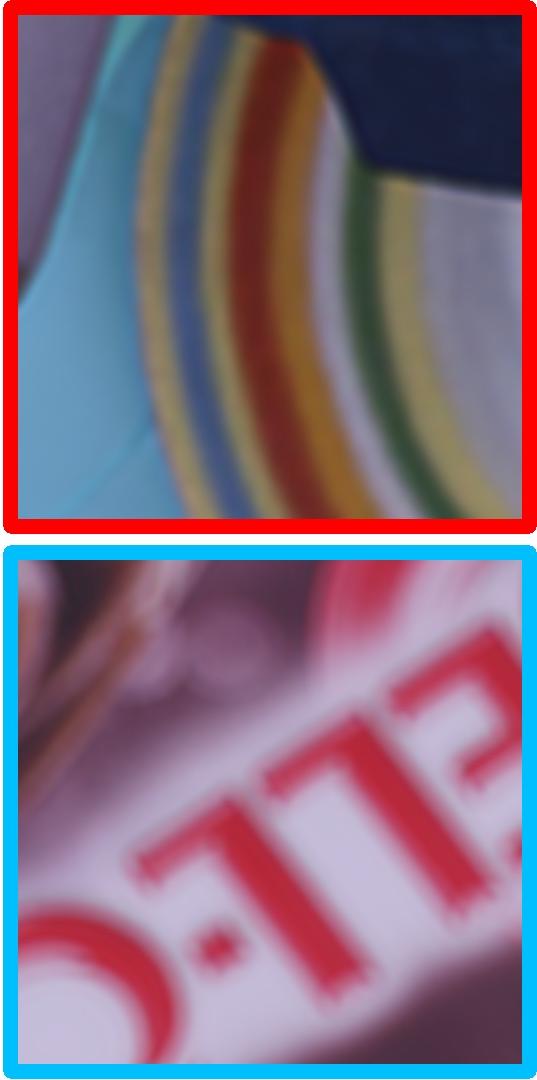}
&\includegraphics[height=2.73cm]{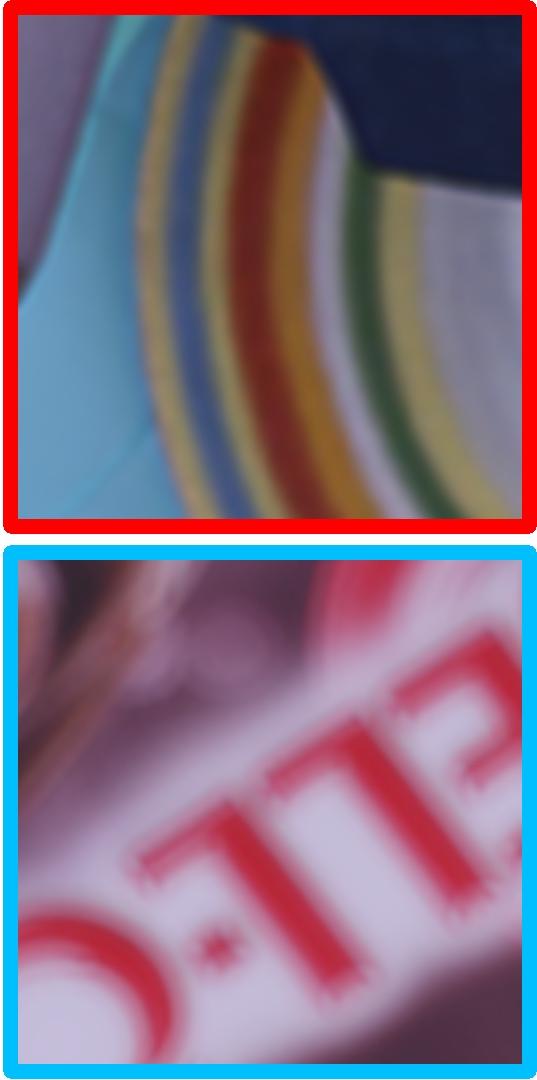}
&\includegraphics[height=2.73cm]{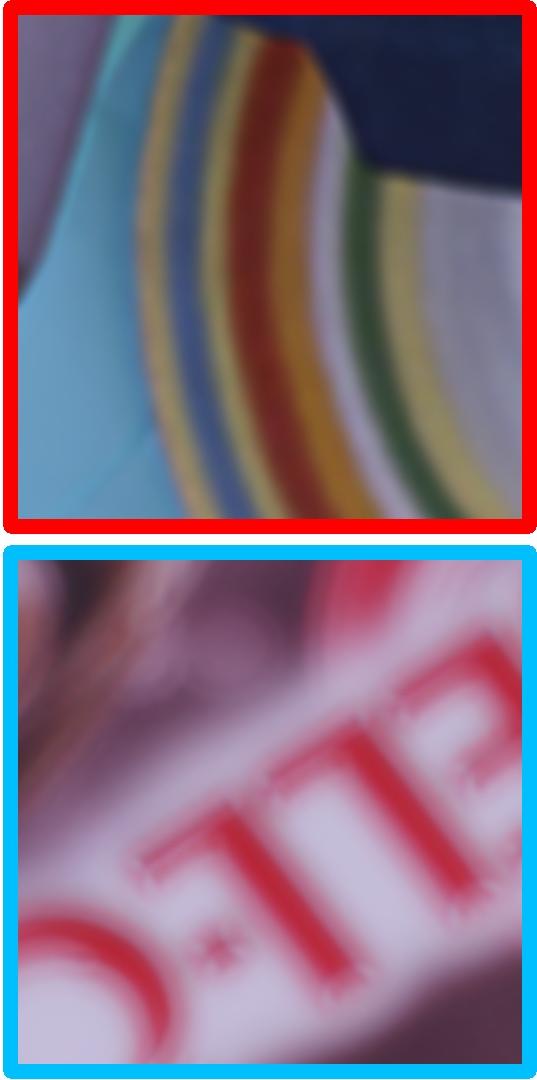}
&\includegraphics[height=2.73cm]{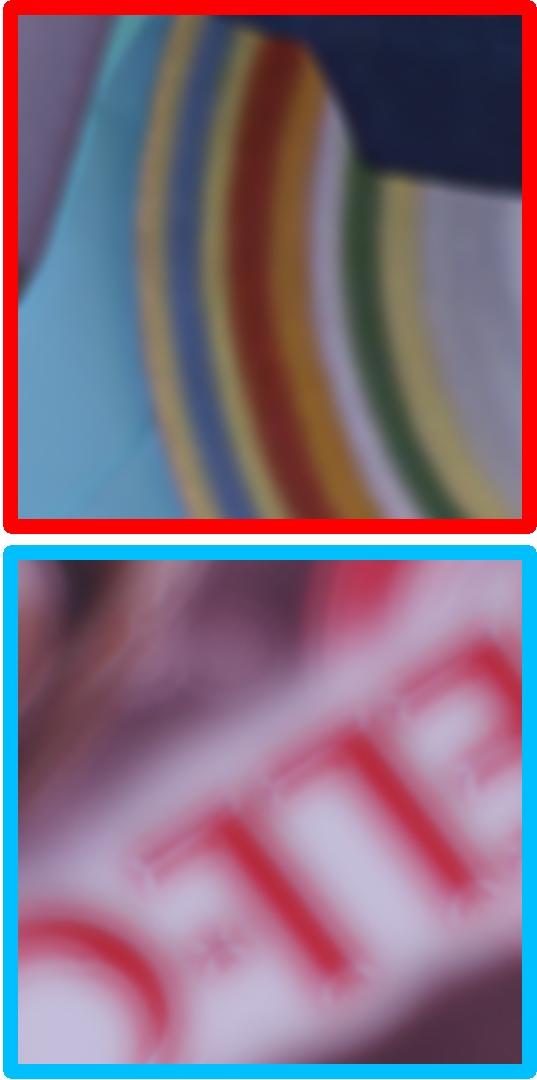}
&\includegraphics[height=2.73cm]{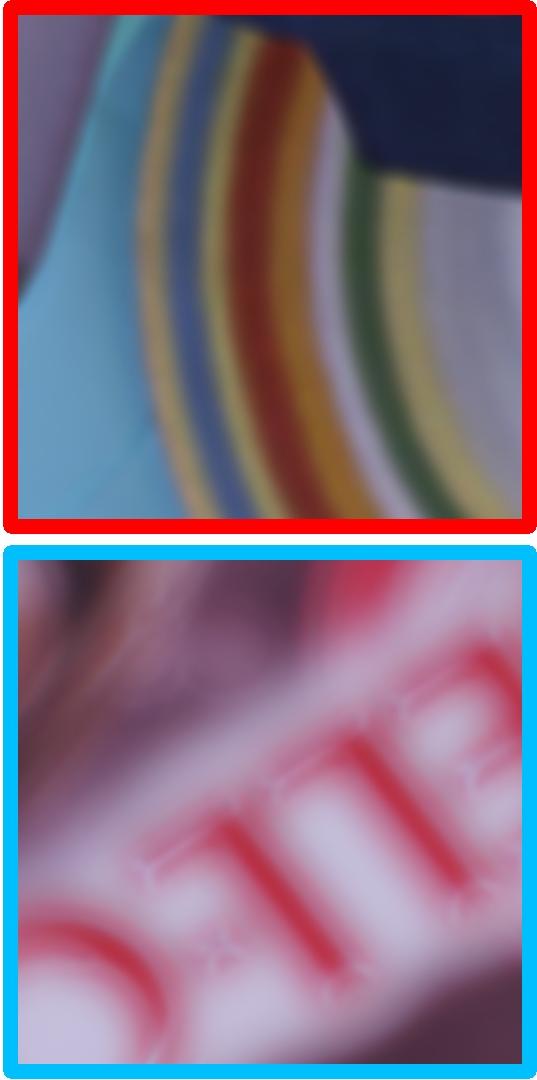}
&\includegraphics[height=2.73cm]{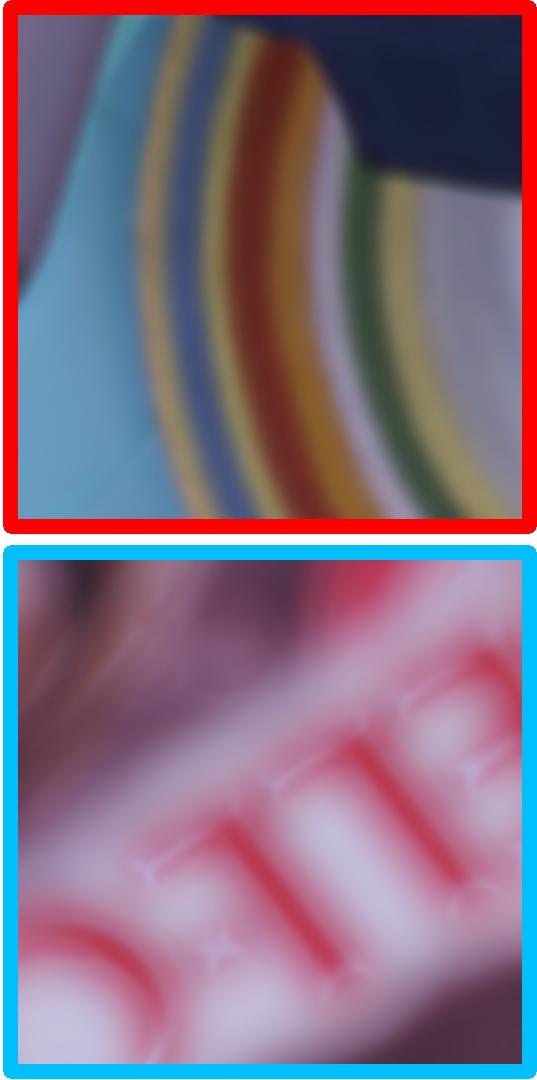}
&\includegraphics[height=2.73cm]{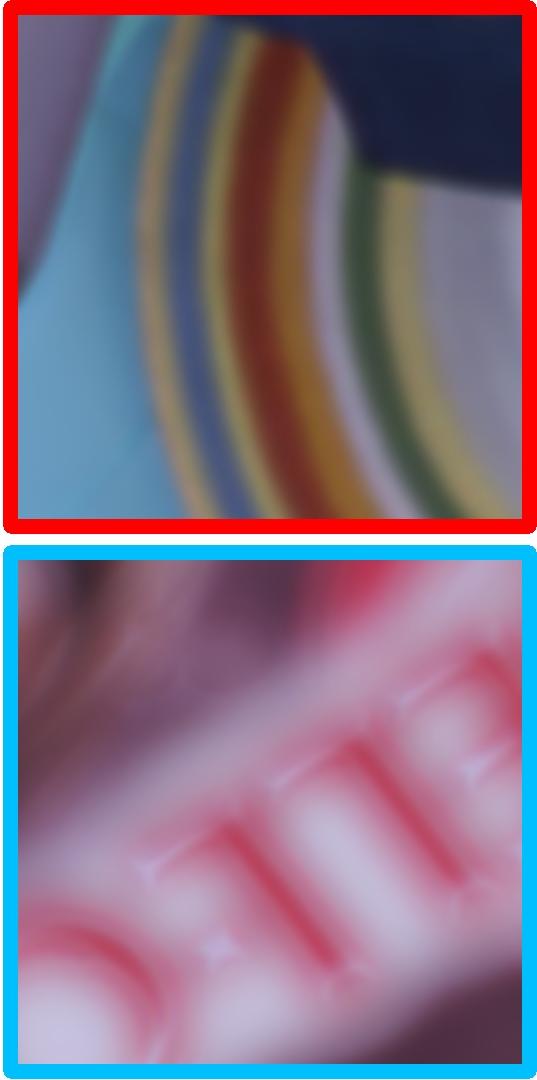}
&\includegraphics[height=2.73cm]{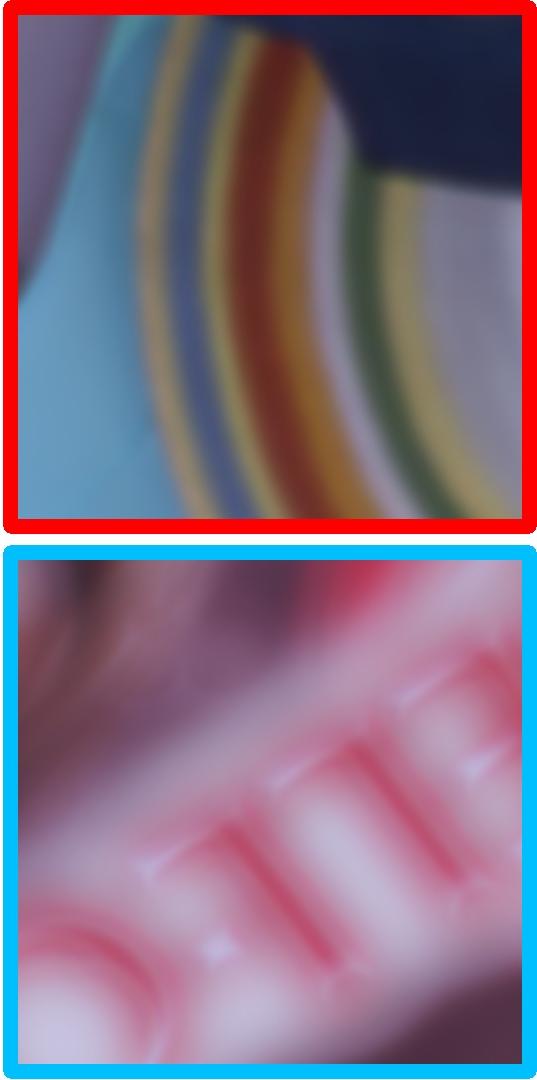}
\\
\raisebox{0.25\height}{\rotatebox{90}{\textbf{Near Focus}}}
&\:&\includegraphics[height=2.73cm]{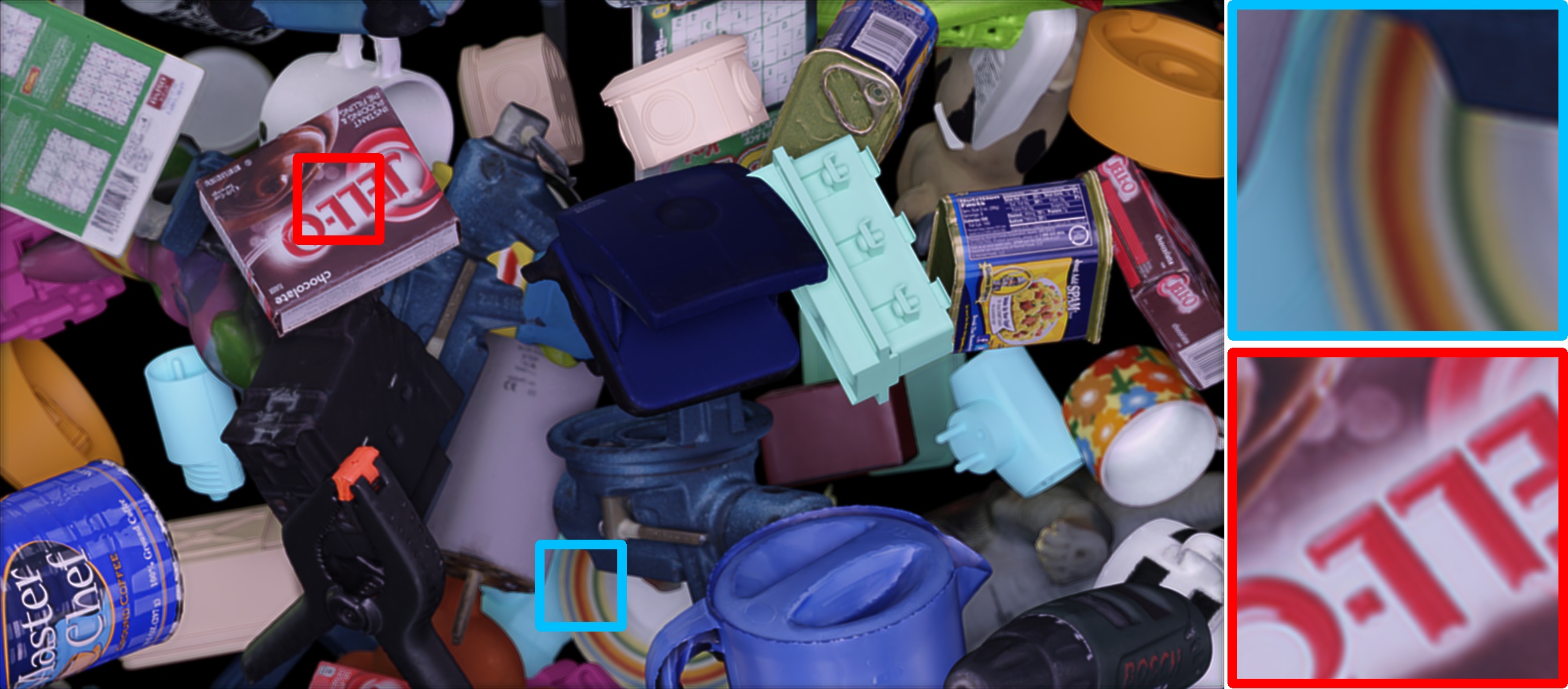}
&\includegraphics[height=2.73cm]{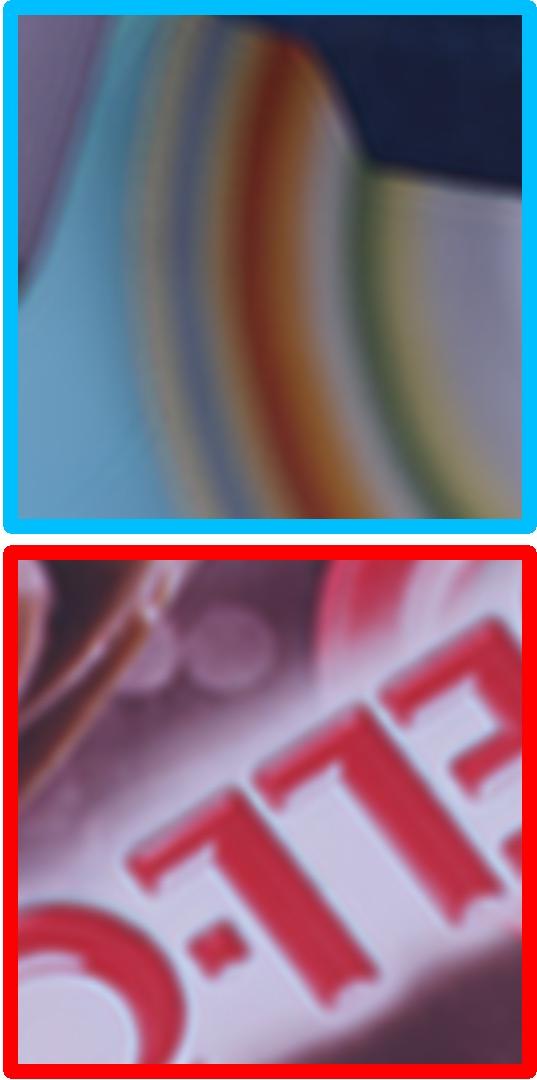}
&\includegraphics[height=2.73cm]{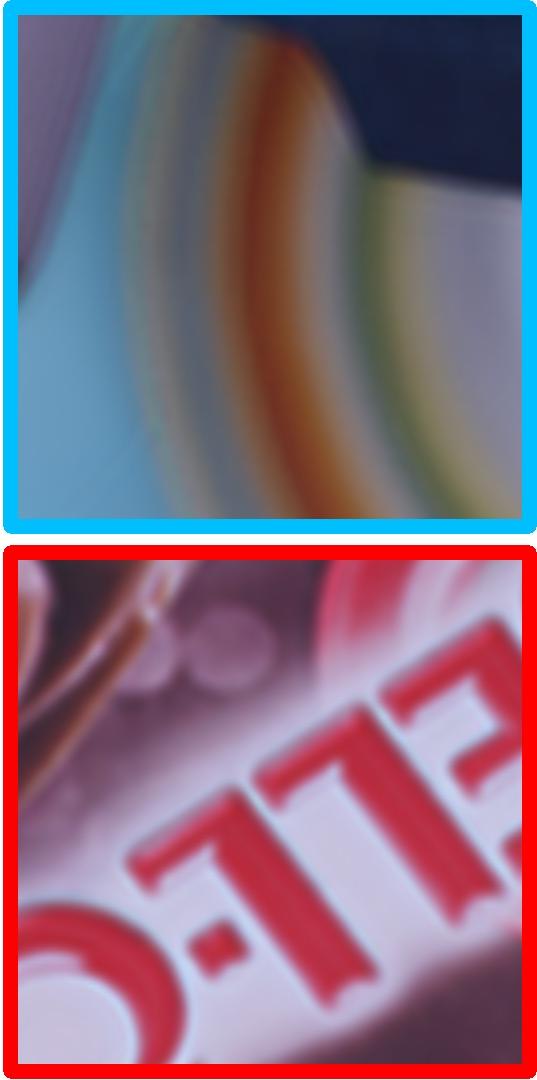}
&\includegraphics[height=2.73cm]{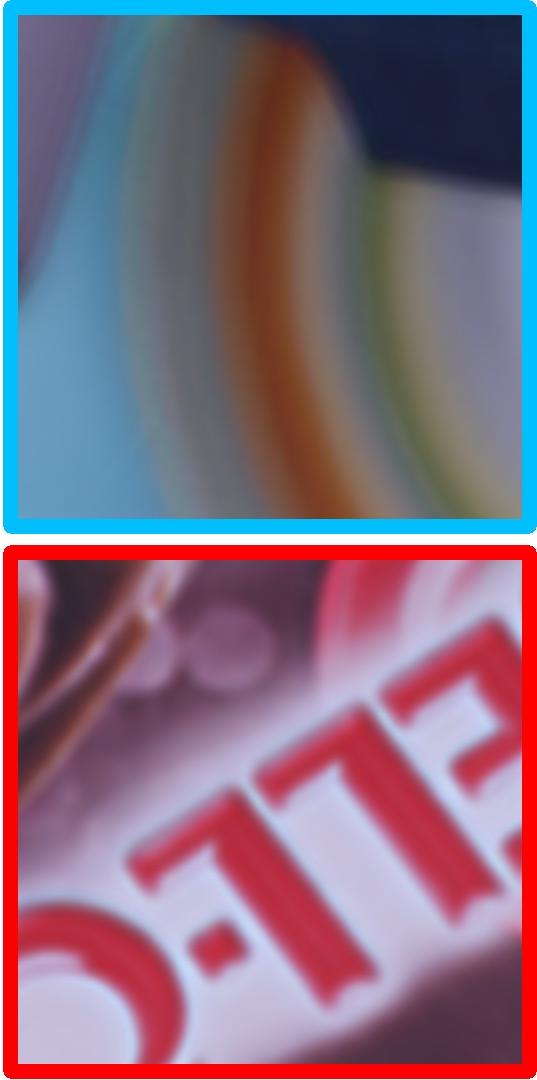}
&\includegraphics[height=2.73cm]{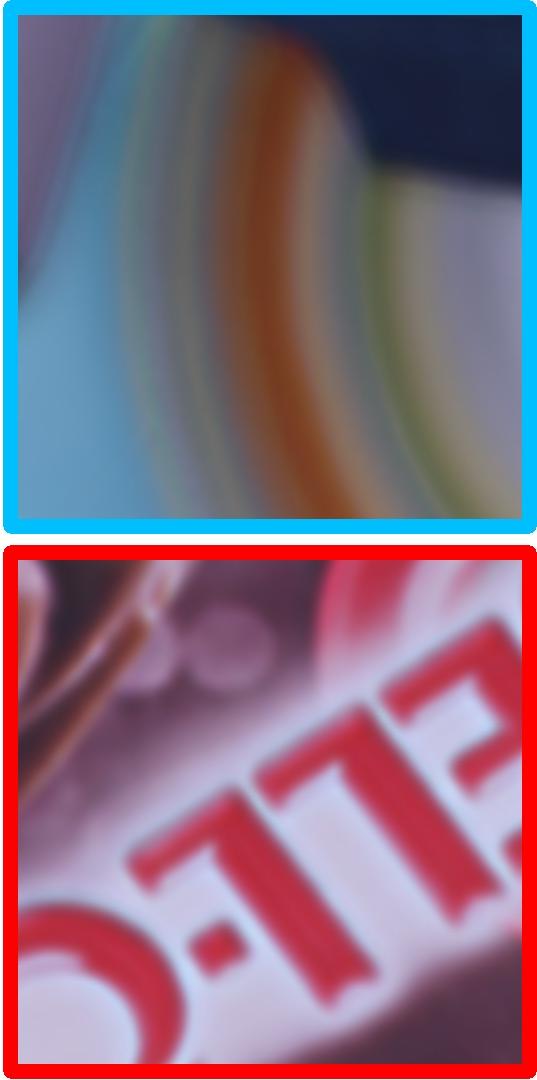}
&\includegraphics[height=2.73cm]{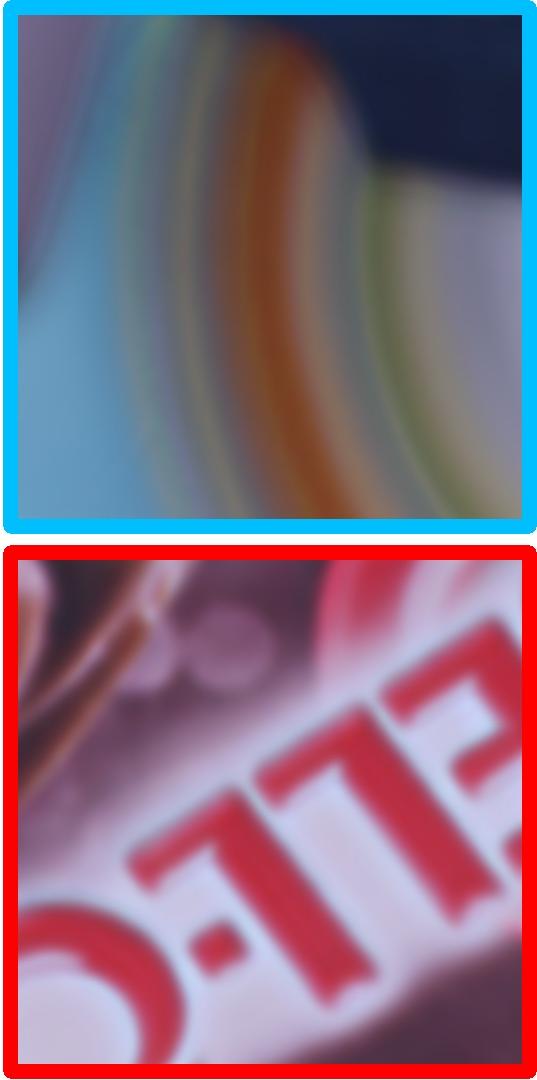}
&\includegraphics[height=2.73cm]{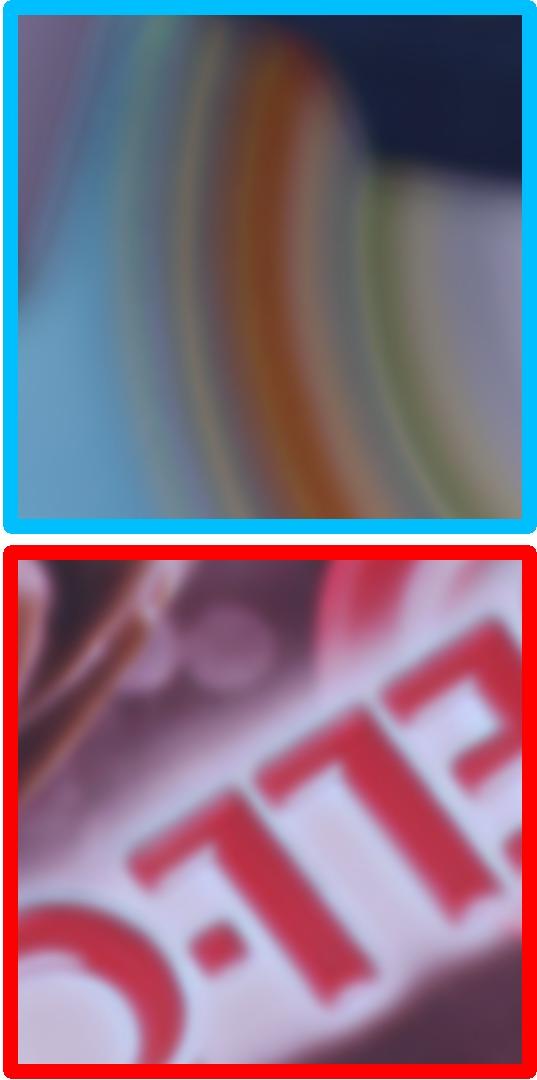}
&\includegraphics[height=2.73cm]{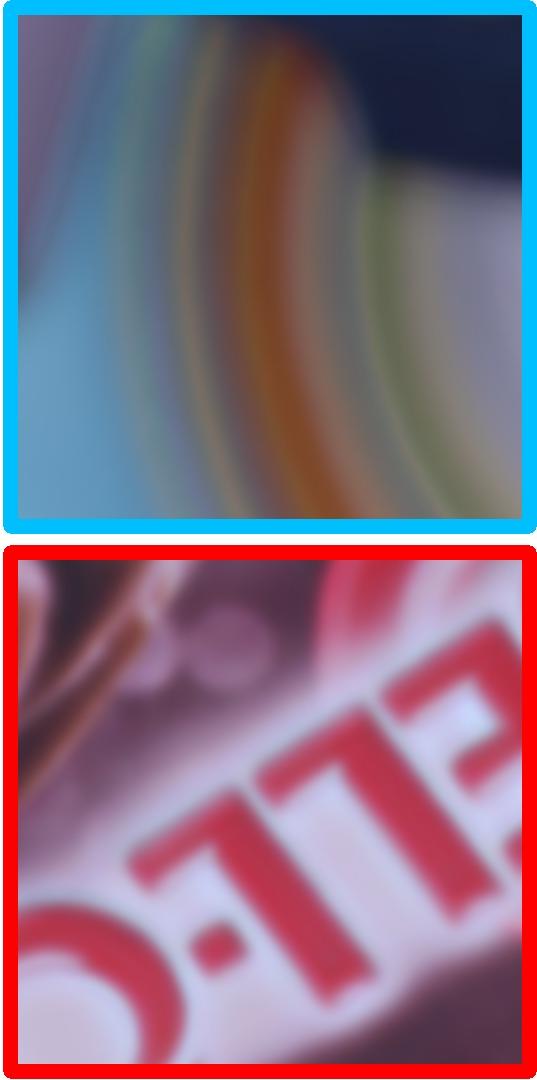}
&\includegraphics[height=2.73cm]{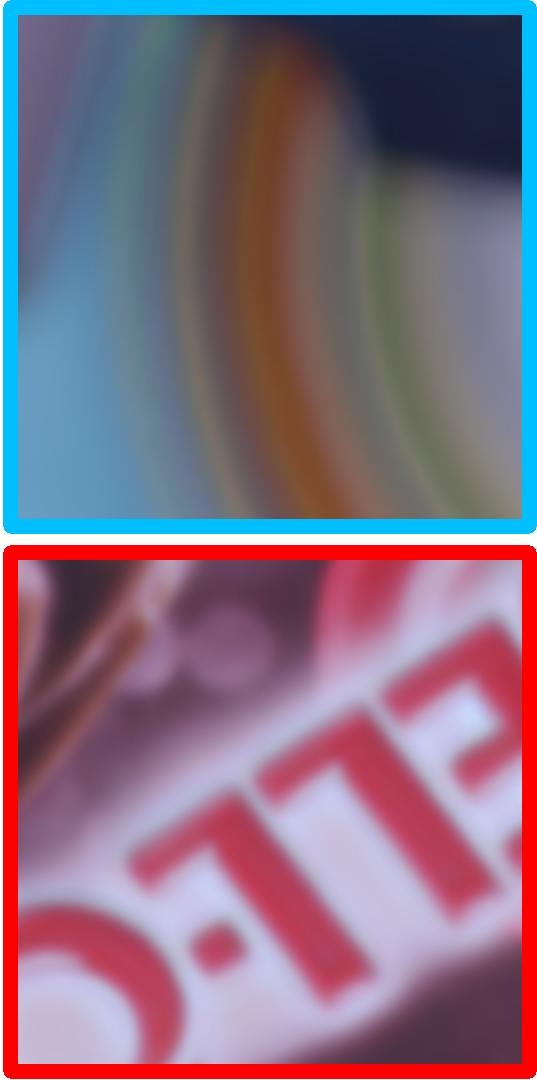}
\\
\end{tabular}
\end{center}
\caption{Simulated reconstruction of holograms obtained from our method with different pupil conditions. The regions highlighted by red bounding boxes are closer to the focal plane while the regions in blue boxes are distant to the focal plane. As shown in the insets, the defocus effects in out-of-focus areas are reconstructed even under unseen pupil sizes including conditions $s=(1.0, 1.5, 4.5, 5.0)$ that exceed the training range and a series of intermediate sizes. 
}
\label{figure:interp}
\end{figure*}
\subsubsection{Analysis of the Predicted ADC Offsets} \label{sec:offsets}
In this section, we aim to gain a deeper understanding of the adjustable deformable convolutional (ADC) layer's functionality within our framework. Specifically, we perform a statistical analysis of the predicted offsets $O^{(s)}$ for various spatial positions. As previously discussed in \cref{sec:approach}, the design of the ADC layer is rooted in the spatially-varying nature of defocus blur and its relationship with pupil size. To explore the connection between these offsets and depth information, we leverage the available ground truth depth and pupil size parameters in our rendered test data. To this end, we collect the predicted offsets for the ADC layers across 100 test examples.

To analyze these offsets, we categorize the input depth into five distinct intervals. We then calculate the mean offset values for positions falling within each depth interval. The resulting statistical findings are presented in \Cref{fig:offset}. The data reveals that, for each pupil size, predicted offsets are more substantial at positions within closer or further depth intervals, whereas they are smaller for positions in the middle-depth range. This aligns with the principle that the maximum blur size for objects at the closest or most distant points is larger across an entire focal stack. It is worth noting that the offsets are also influenced by the color of the input images. Nevertheless, the statistical results indicate that the adjustable deformable convolutions effectively learn to adapt the receptive field for different positions based on the input depth cues.
Additionally, because the offset prediction layer $g_o$ is influenced by the input pupil size, the offsets are proportionally adjusted according to the pupil parameter. This feature contributes to the adaptability of the ADC layer, ensuring that receptive fields are appropriately tuned to different spatial positions, accounting for both depth and pupil size variations.


\subsubsection{Network Capacity Analysis} \label{sec:capacity}
We assess the impact of varying network capacities by evaluating three variants with different numbers of layers: 14, 20, and 26 layers respectively, and report the results in \Cref{fig:capacity}. 
The results in \Cref{fig:capacity} demonstrate that our full model, employing $26$ layers by default, achieves the best performance. However, the variant \emph{Ours (20-layer)} only exhibits a slight drop in quality as reflected by LPIPS (v) and PSNR metrics. Reducing the number of layers further to $14$ leads to a more noticeable performance decline, though it remains acceptable. In summary, our framework can be scaled down without significant performance degradation.

\setlength{\tabcolsep}{2pt}
\renewcommand{\arraystretch}{0.6}
\begin{figure*}[t]
\begin{center}
\small
\begin{tabular}{cccccc}

& \textbf{Ours (2mm Pupil)} & \textbf{Ours (3mm Pupil)} & \textbf{Ours (4mm Pupil)} & \textbf{Shi et al. 2021} & \textbf{Shi et al. 2022} \\ [0.5ex]
\raisebox{0.7\height}{\rotatebox{90}{\textbf{Far Focus}}}
&\includegraphics[width=3.35cm]{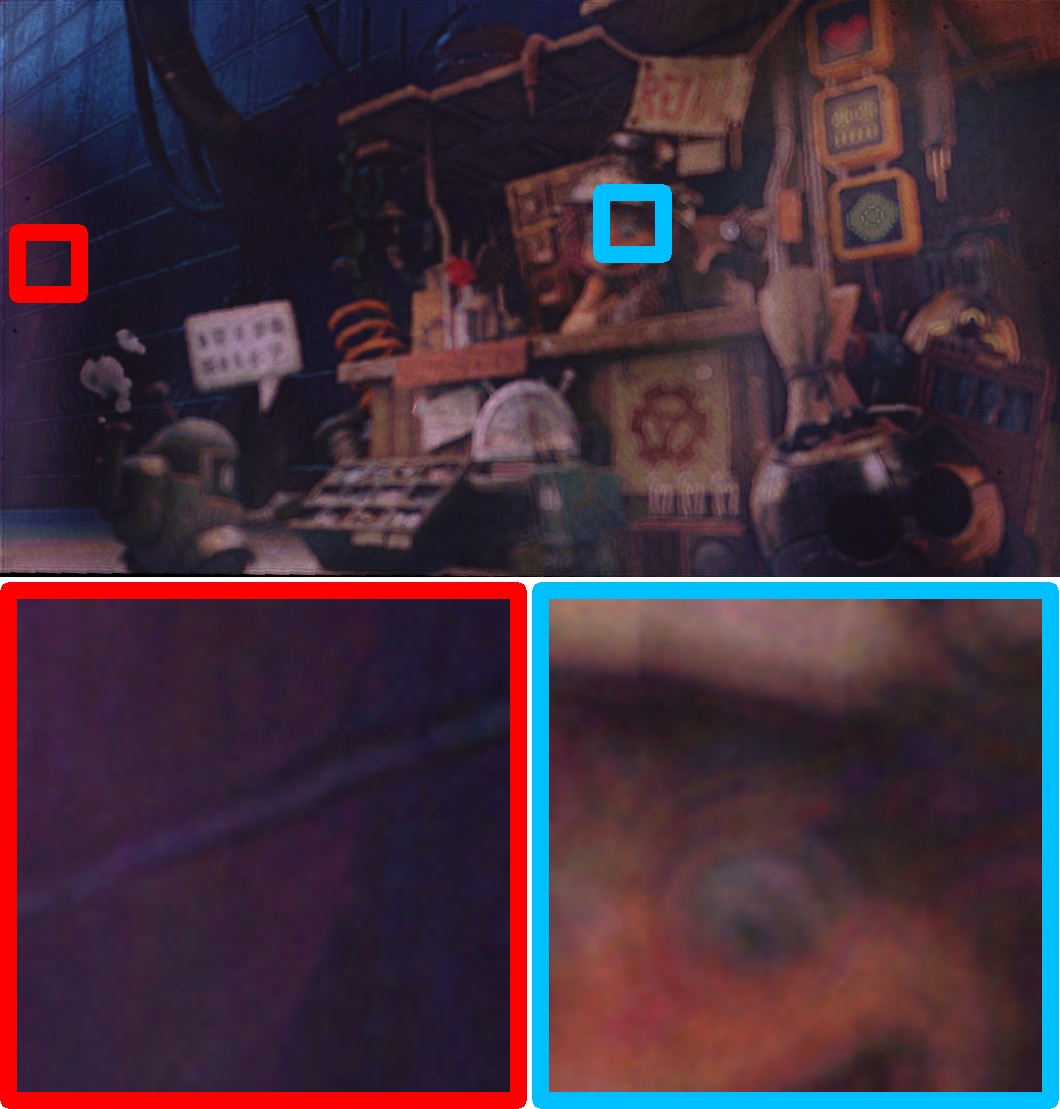}
&\includegraphics[width=3.35cm]{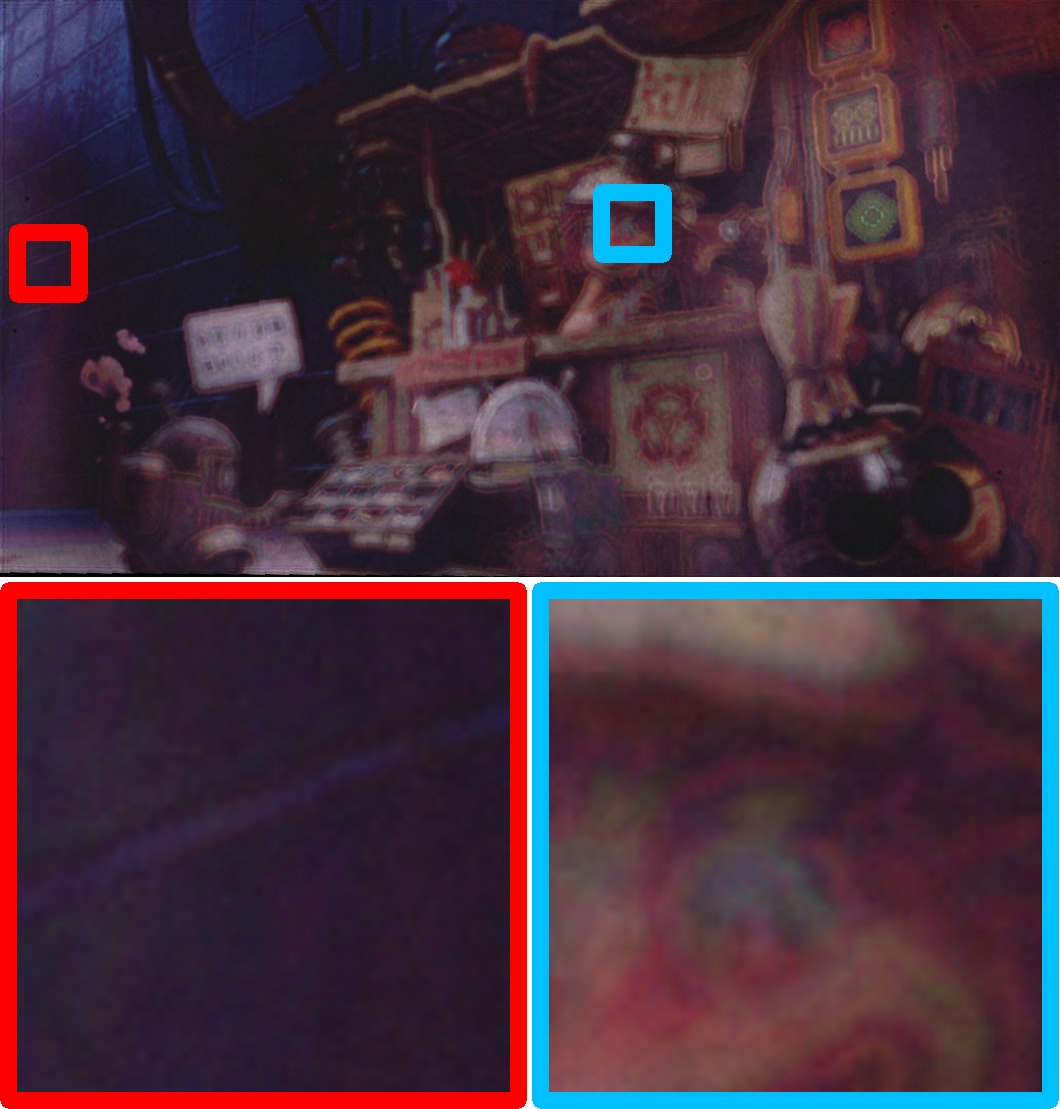}
&\includegraphics[width=3.35cm]{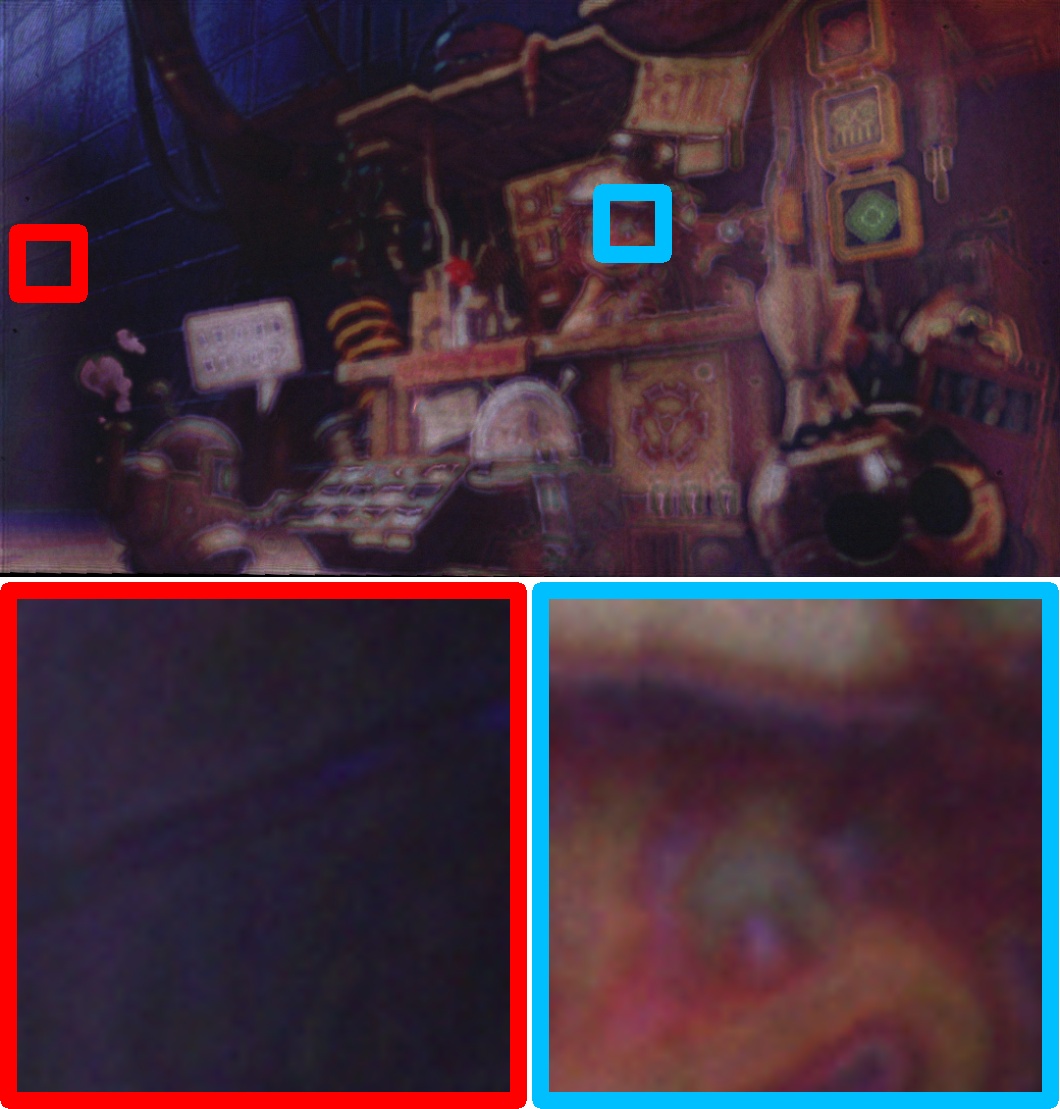}
&\includegraphics[width=3.35cm]{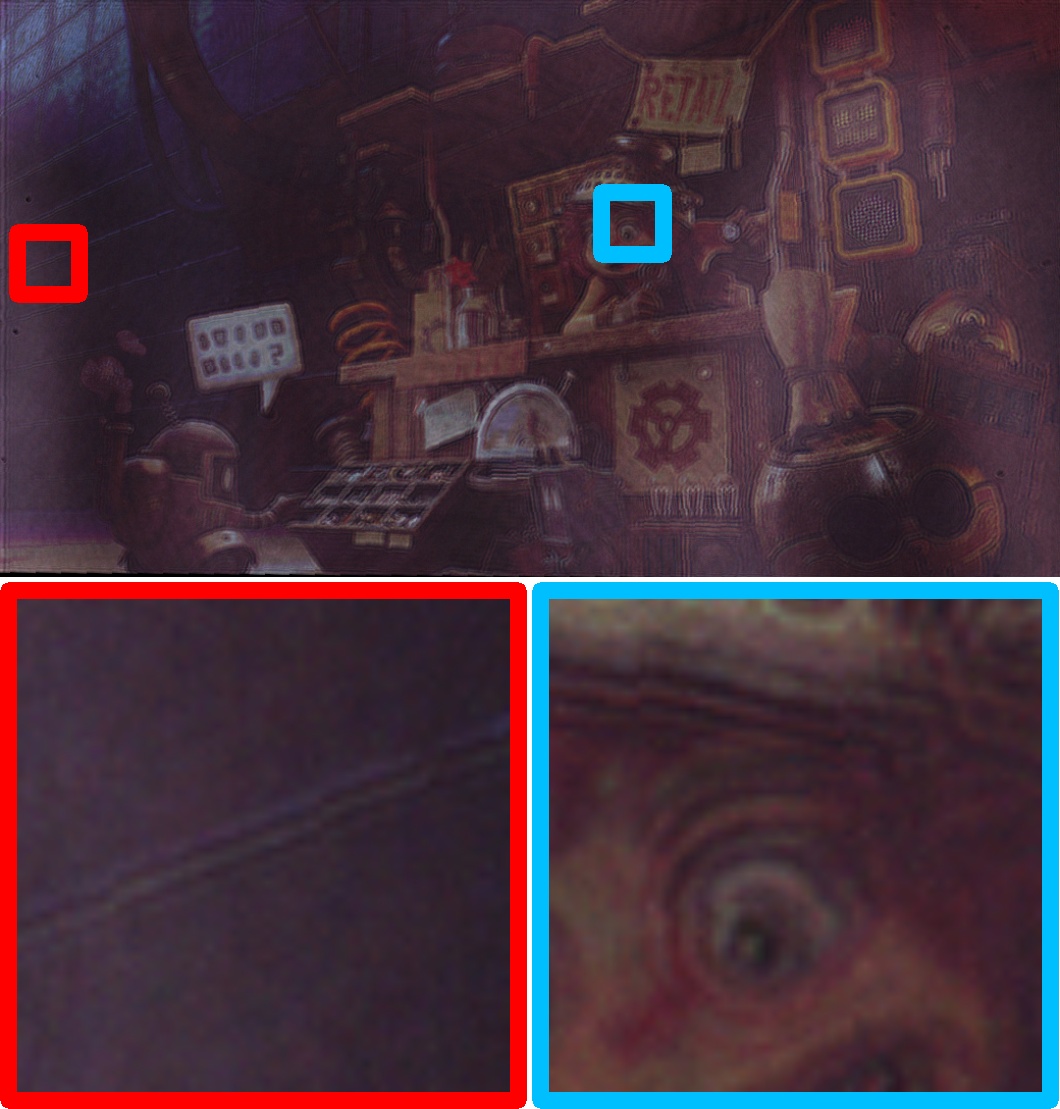}
&\includegraphics[width=3.35cm]{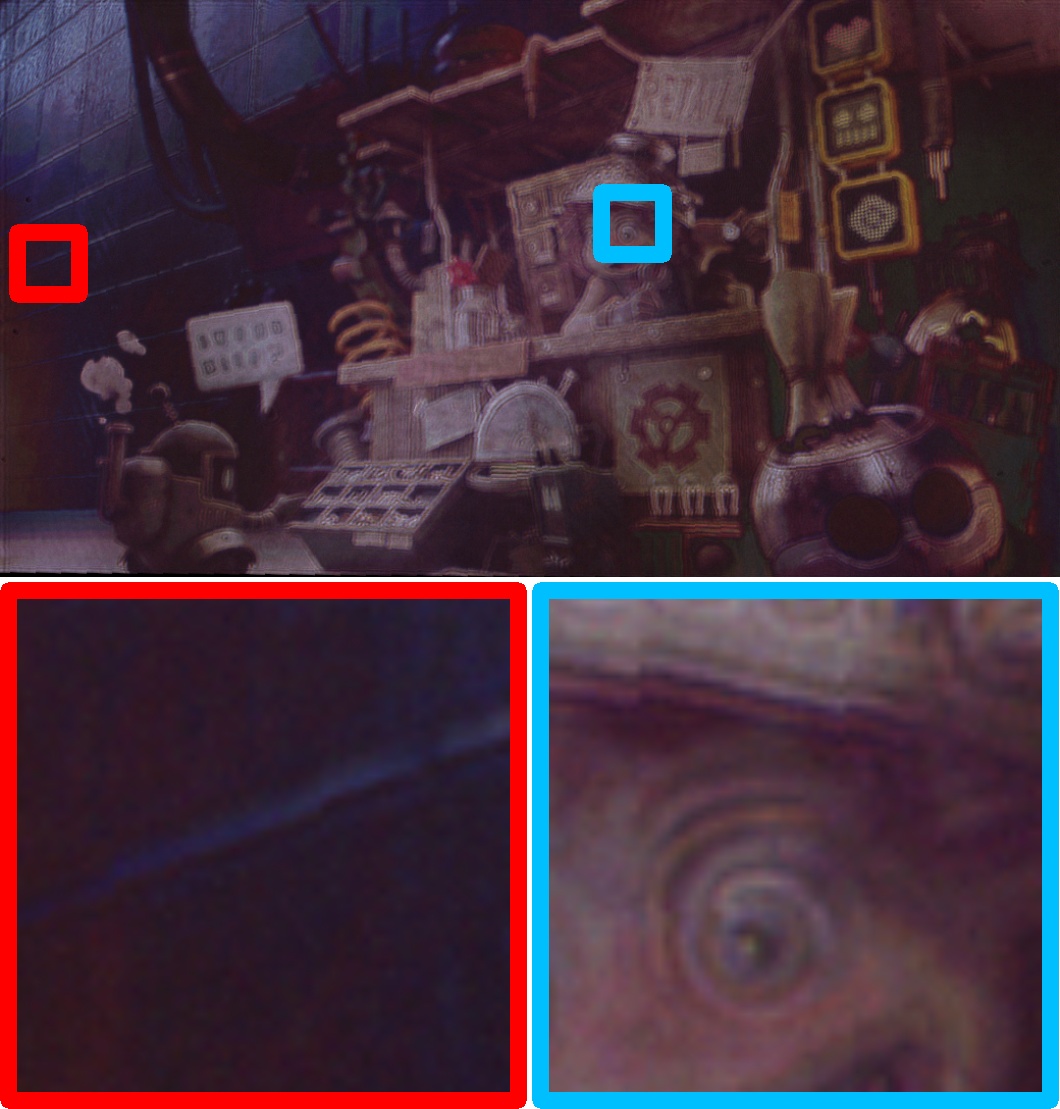}
\\ [1.2ex]
\raisebox{0.7\height}{\rotatebox{90}{\textbf{Near Focus}}}
&\includegraphics[width=3.35cm]{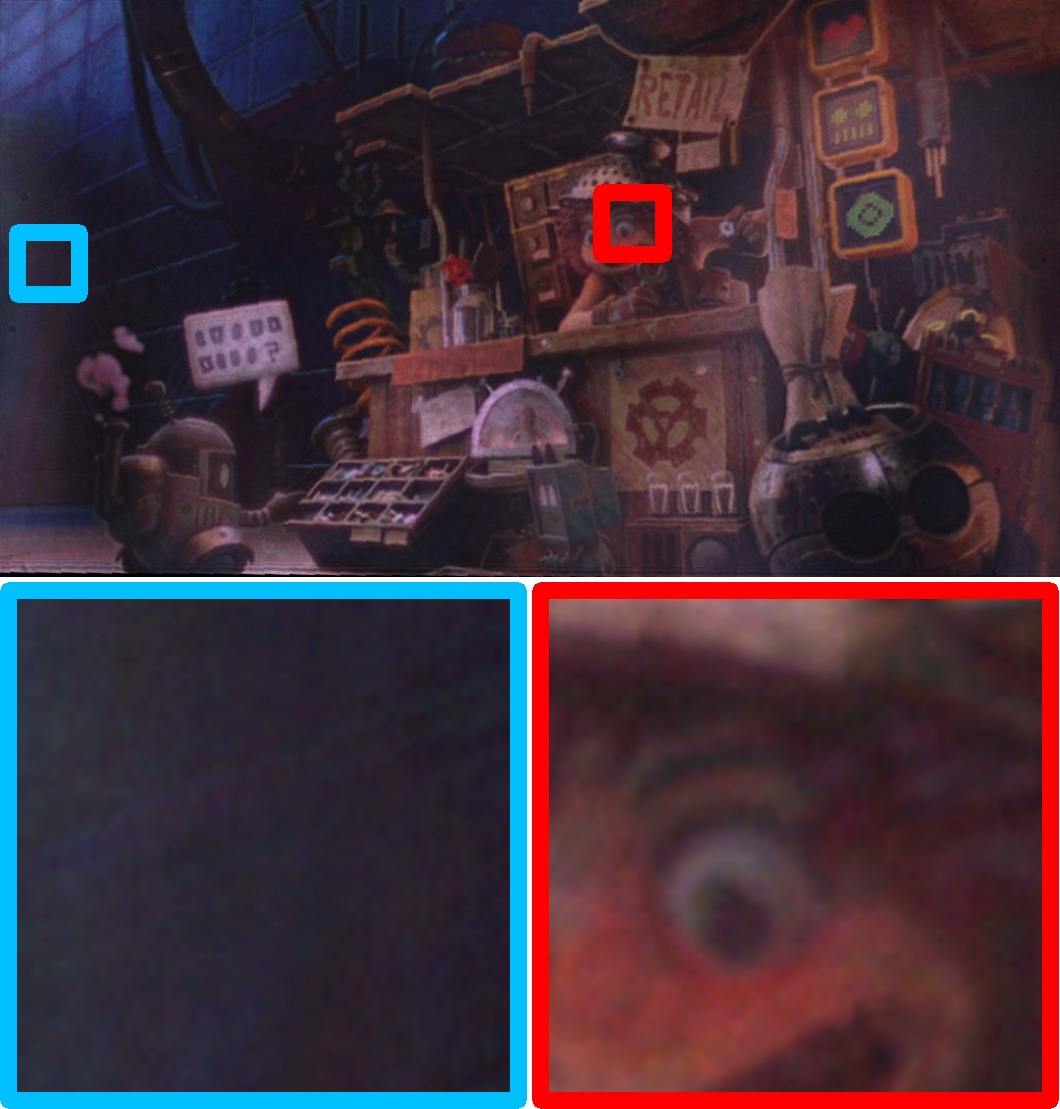}
&\includegraphics[width=3.35cm]{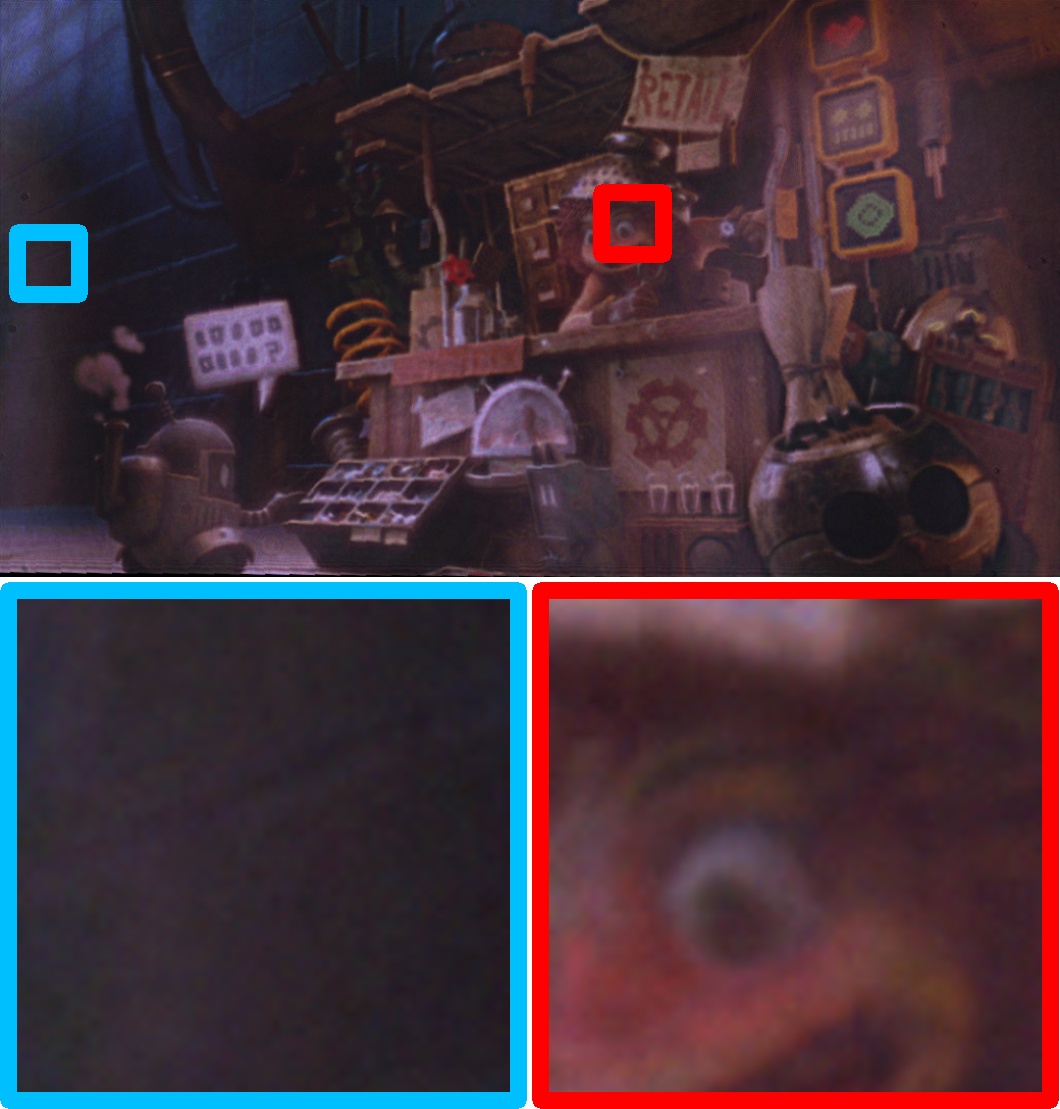}
&\includegraphics[width=3.35cm]{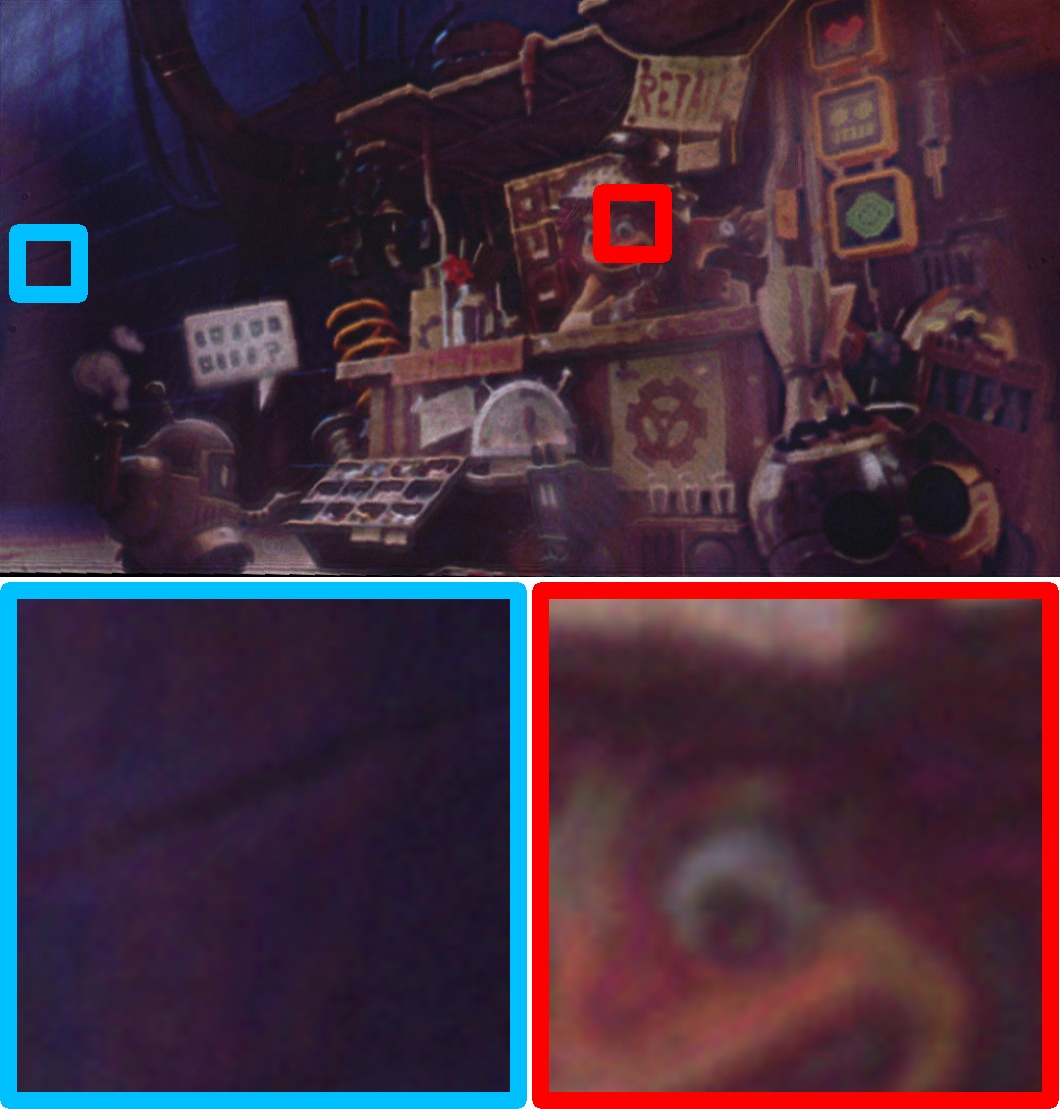}
&\includegraphics[width=3.35cm]{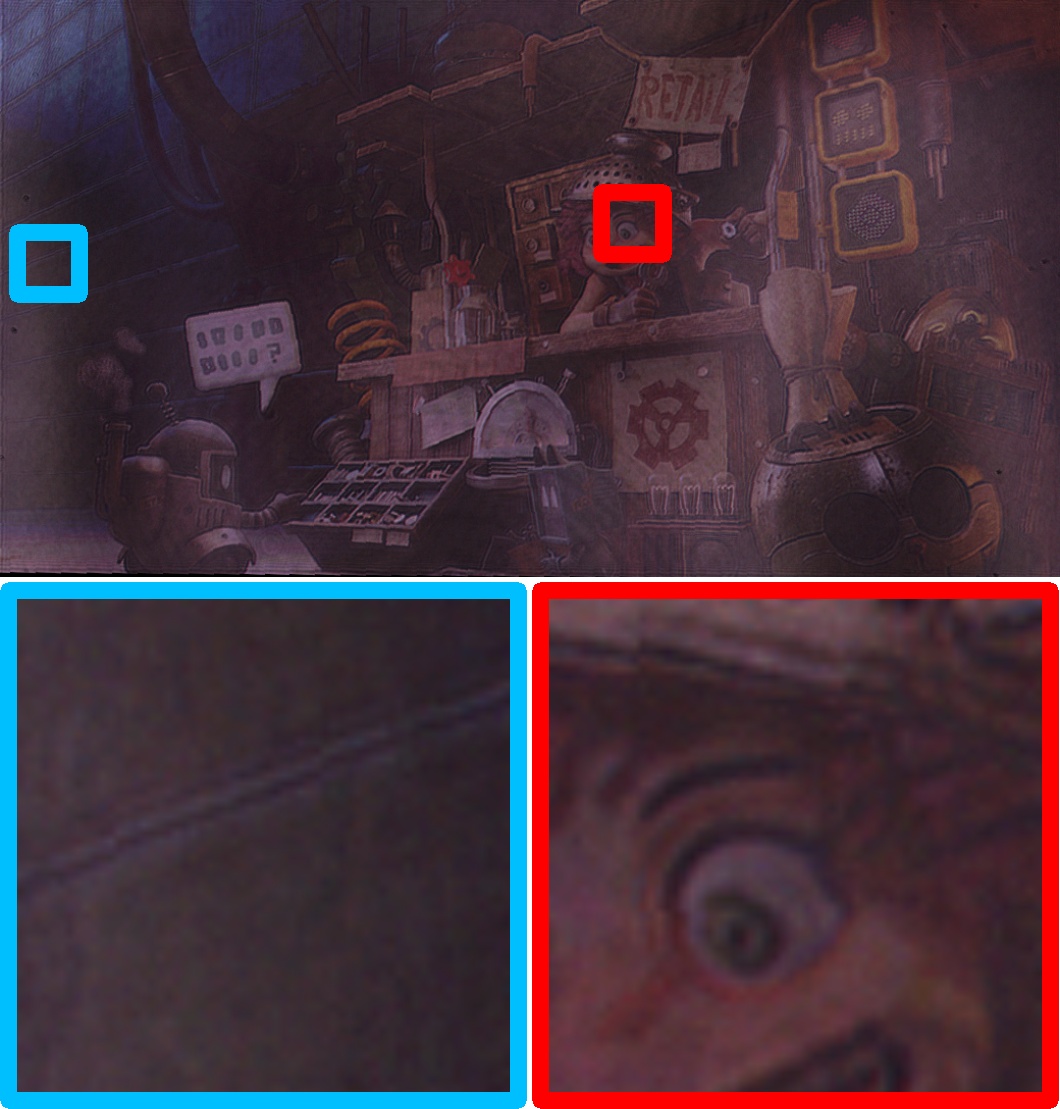}
&\includegraphics[width=3.35cm]{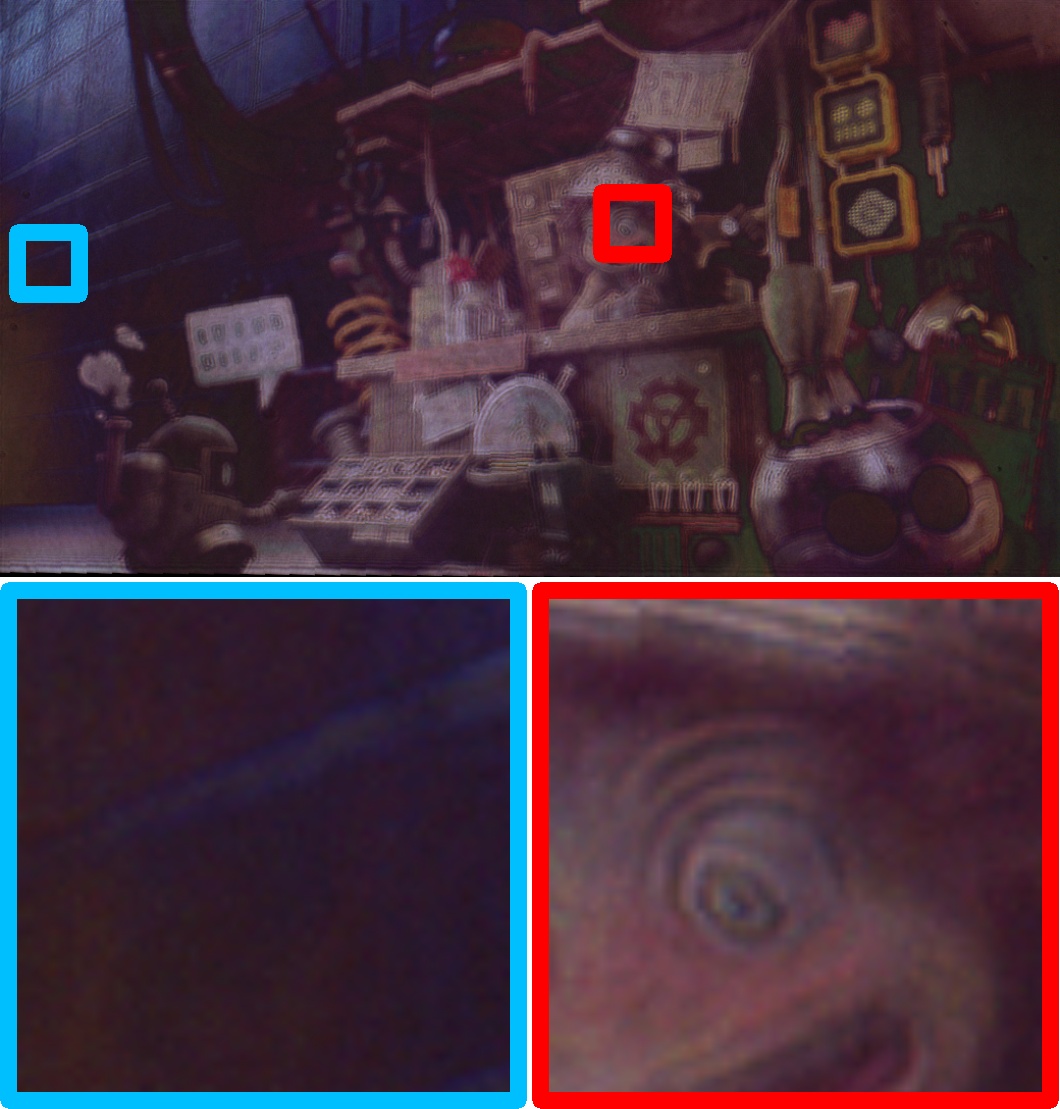}
\\
\end{tabular}
\end{center}
\caption{Experimental evaluation of various hologram methods. In-focus regions are highlighted in red and out-of-focus regions are marked in blue. Only one group of results is provided for methods that do not consider varying pupil sizes and corresponding depth-of-field effects. As shown in the insets, \emph{i.e.}, the distant wall and the closer eye region, our results exhibit appropriate defocus effects for a variety of pupil sizes while the results of other methods show ringing artifacts or less conspicuous defocus effects.}
\label{figure:hardware2}
\end{figure*}



\setlength{\tabcolsep}{2pt}
\renewcommand{\arraystretch}{1.3}
\begin{table}[]
\small
\caption{Runtime performance of different hologram generation methods}\label{tab:running_time}
\begin{tabular}{|m{2.6cm}|l||m{3cm}|l|l}
\cline{1-4}
Method & Time (s) & Method & Time (s) &  \\ \cline{1-4} \cline{1-4}
Shi et al. 2021   &   $0.11$          & Yang et al. 2022      &      $0.13$        &  \\
Shi et al. 2022  &   {$0.12$}          & Lee et al. 2022 (TM $=8$)      &     $490.65$         &  \\
Choi et al. 2021       &  $99.52$   & Lee et al. 2022 (TM $=24$)     &     $1462.52$          &  \\ 
Kavakl{\i} et al. 2023   &   $5.59$           &  Ours ($26$-layer)  &     {$0.52$}  &  \\ 
Ours ($20$-layer)   &     {$0.43$}    & Ours ($14$-layer)   &     {$0.29$}        &  \\ \cline{1-4}
\end{tabular}
\vspace{2mm}
\end{table}
\rerevise{
\subsubsection{Efficiency Analysis} \label{sec:runtime}
We assessed the computational time required to generate $1920\times1080$ holograms using different methods and summarized the results in \Cref{tab:running_time}.
Notably, the solutions by Choi et al. \shortcite{choi2021neural3d}, Kavakl\i et al. \shortcite{realistic_blur}, and Lee et al. \shortcite{b-sgd} exhibit relatively longer execution times due to their iterative optimization procedures. The impact is most pronounced for Lee et al. \shortcite{b-sgd} as it optimizes a substantial number of frames in each run for time-multiplexing. The time cost of the method \cite{b-sgd} in \Cref{tab:running_time} is evaluated with setting the number of depth planes to $10$.
While our proposed framework does introduce a slight computational overhead due to the flexible deformable convolutional layers, it remains competitive with existing neural network-based methods such as those by Shi et al. \shortcite{shi2021nature, shi2022light} and Yang et al. \shortcite{deh2022}.
Note that our tests for Yang et al. \shortcite{deh2022} were conducted on a re-implemented framework due to inaccessibility of the code. We also note that further enhancements in computational efficiency and inference speeds can be achieved through techniques like quantization and memory access optimizations for deformable convolutional layers as suggested in Ahn et al. \shortcite{Ahn2020}.
}

\setcounter{figure}{17}
\setlength{\tabcolsep}{2pt}
\renewcommand{\arraystretch}{0.6}
\begin{figure*}[htbp!]
\begin{center}
\small
\begin{tabular}{cccccc}

& \textbf{Ours (2mm Pupil)} & \textbf{Ours (3mm Pupil)} & \textbf{Ours (4mm Pupil)} & \textbf{Shi et al. 2021} & \textbf{Shi et al. 2022} \\ [0.5ex]
\raisebox{0.7\height}{\rotatebox{90}{\textbf{Far Focus}}}
&\includegraphics[width=3.35cm]{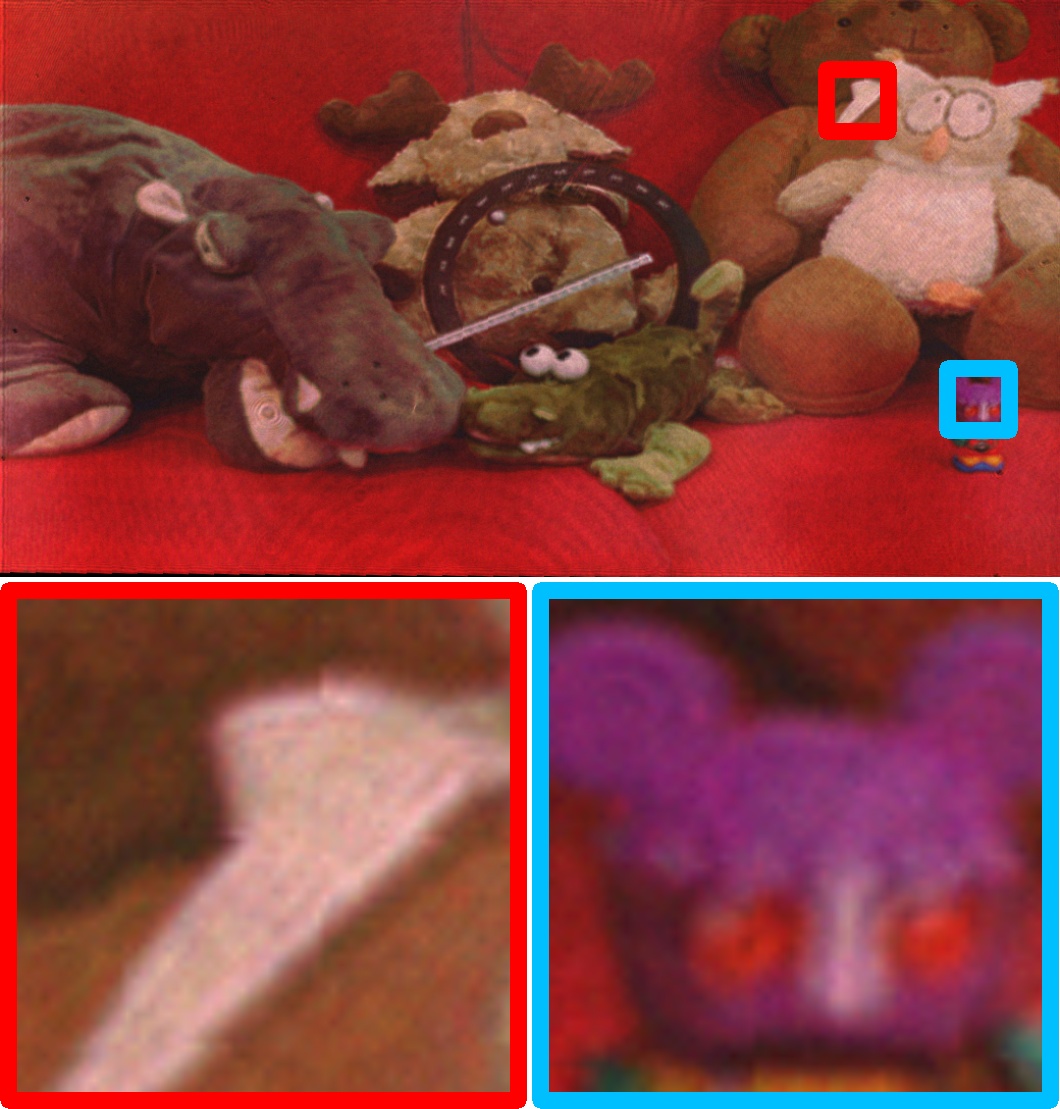}
&\includegraphics[width=3.35cm]{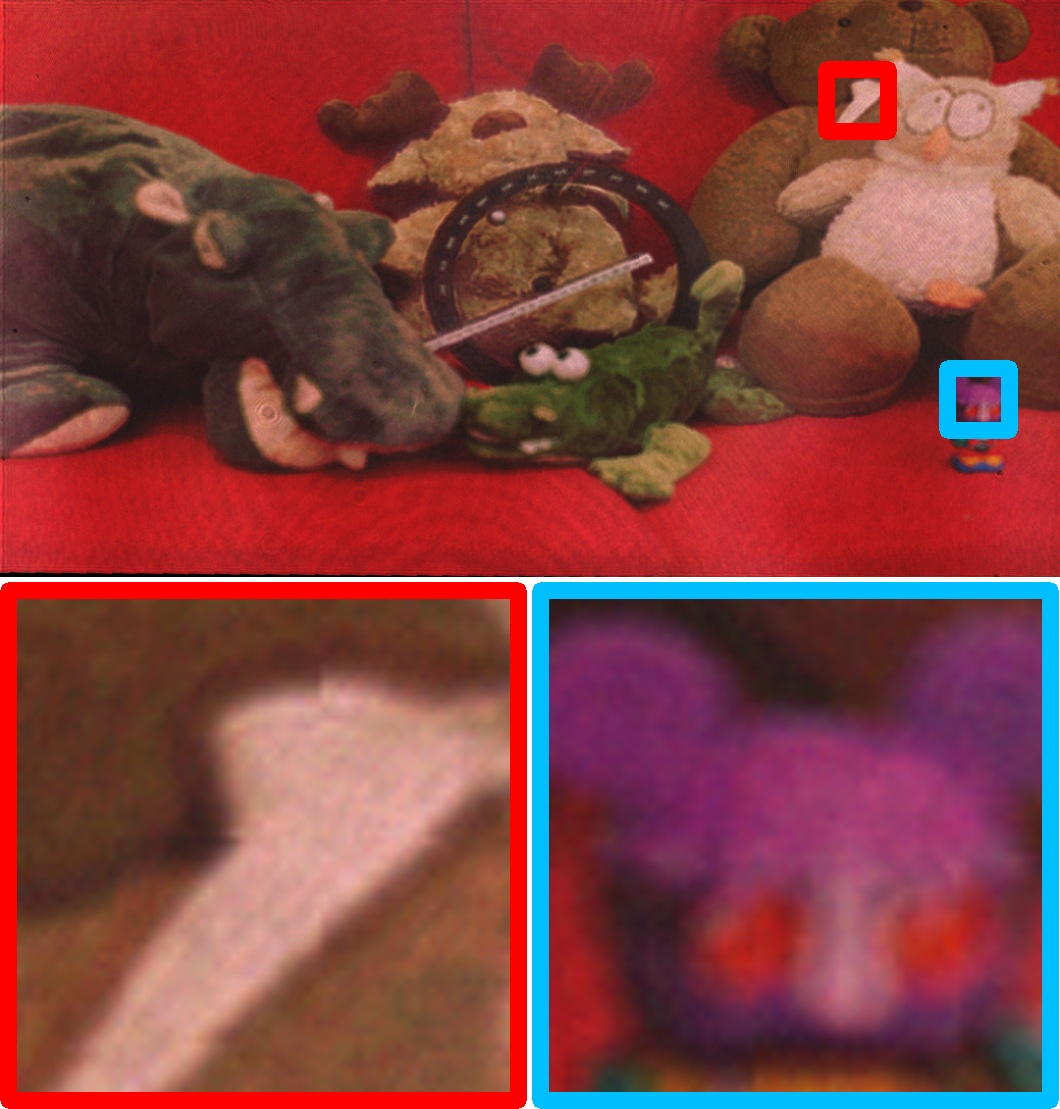}
&\includegraphics[width=3.35cm]{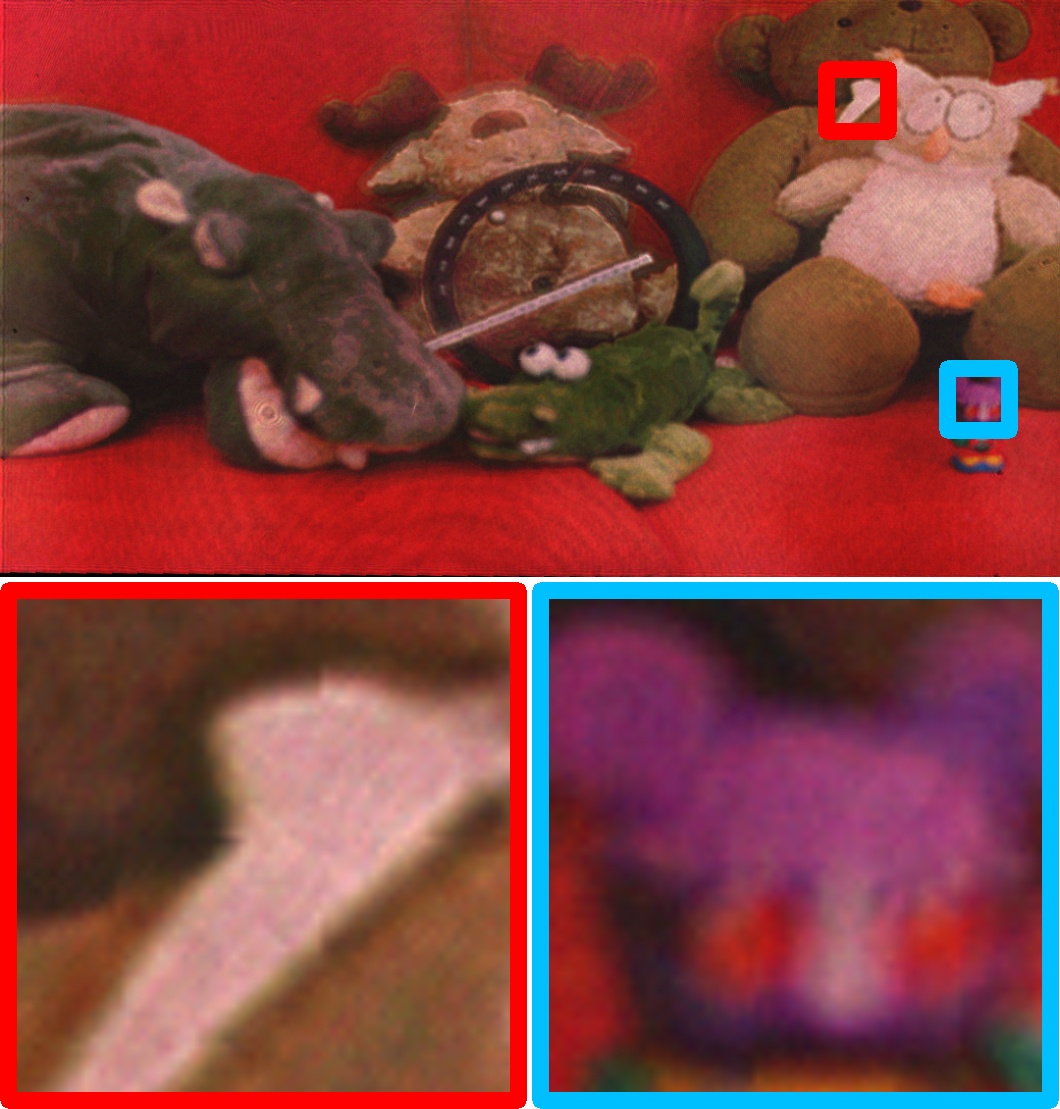}
&\includegraphics[width=3.35cm]{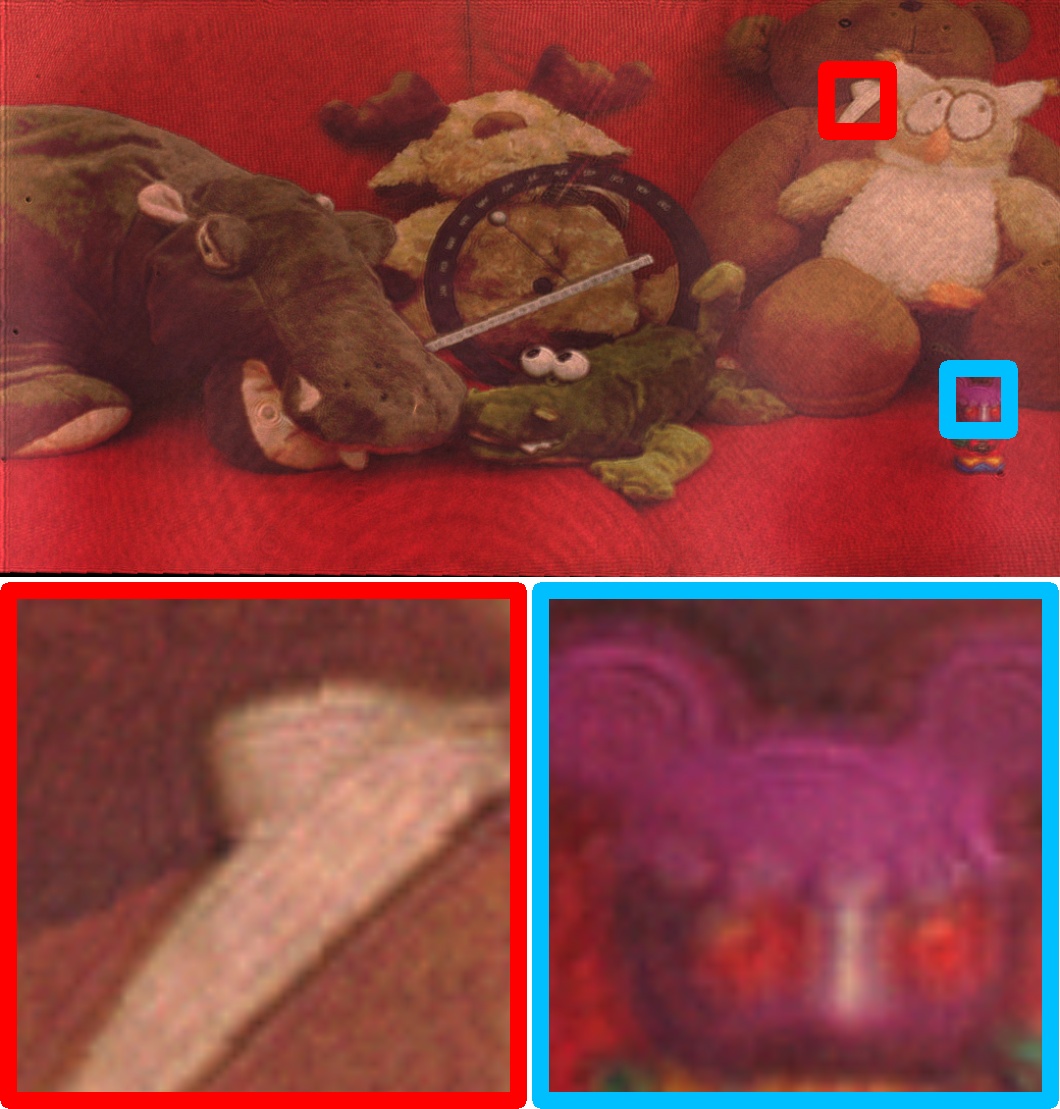}
&\includegraphics[width=3.35cm]{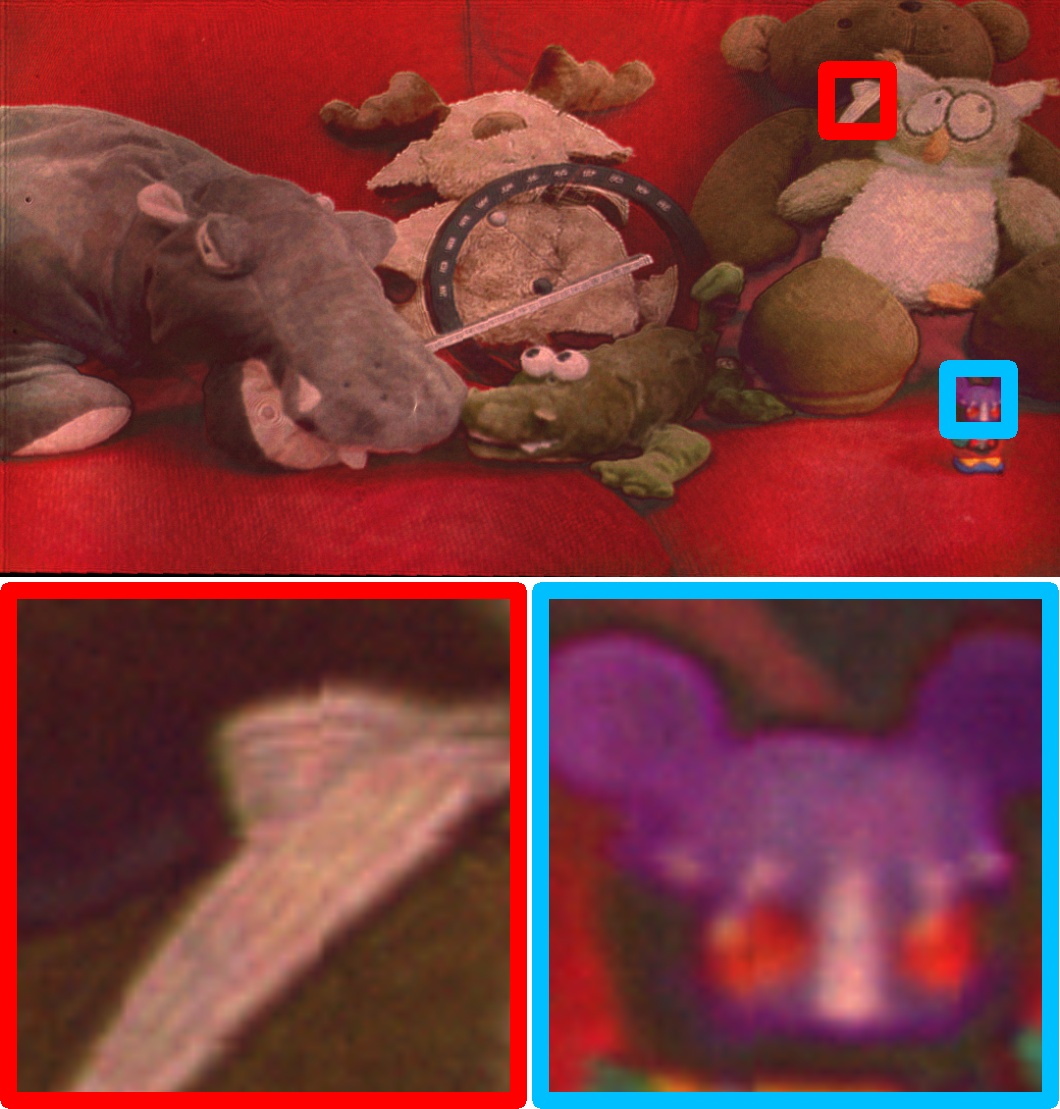}
\\ [1.2ex]
\raisebox{0.7\height}{\rotatebox{90}{\textbf{Near Focus}}}
&\includegraphics[width=3.35cm]{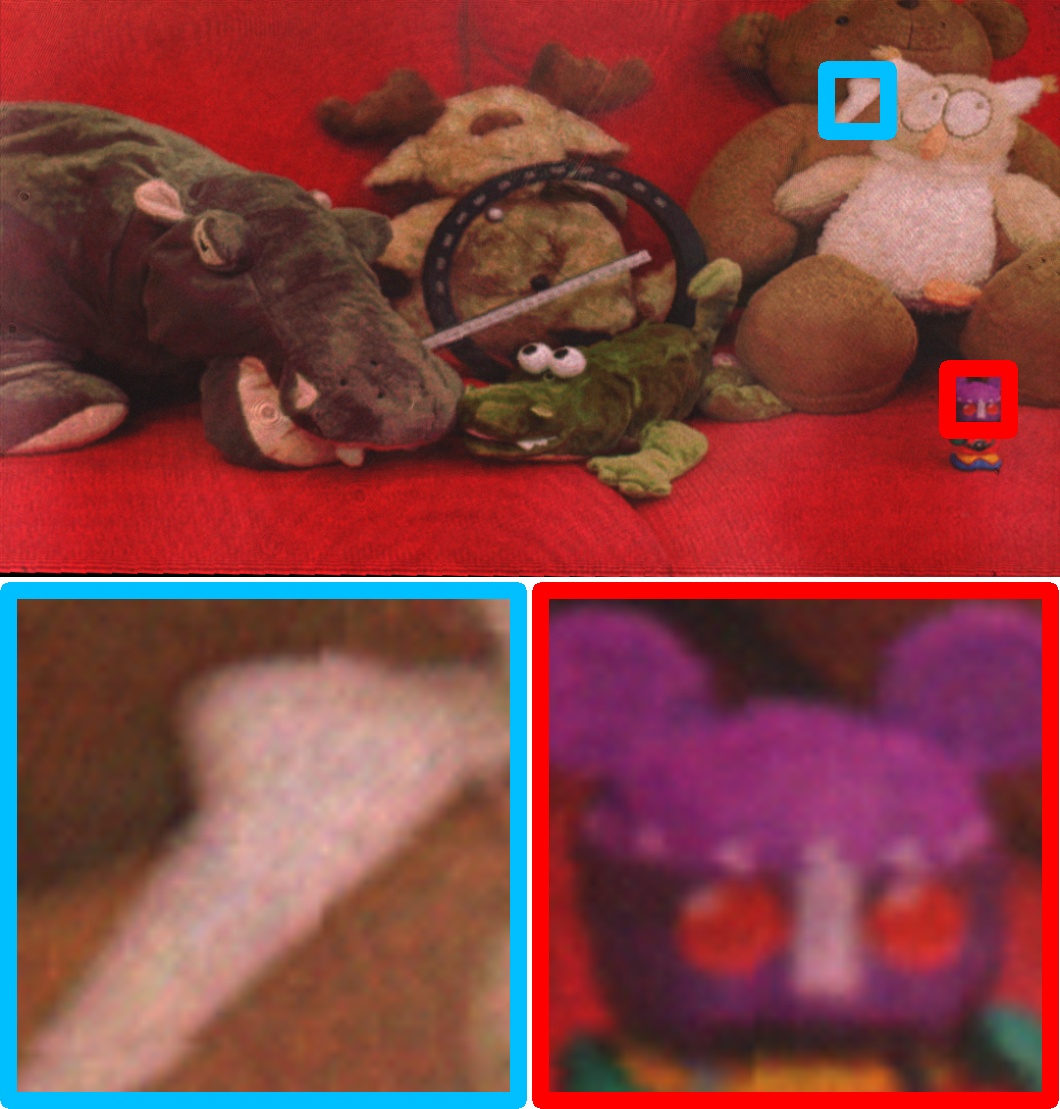}
&\includegraphics[width=3.35cm]{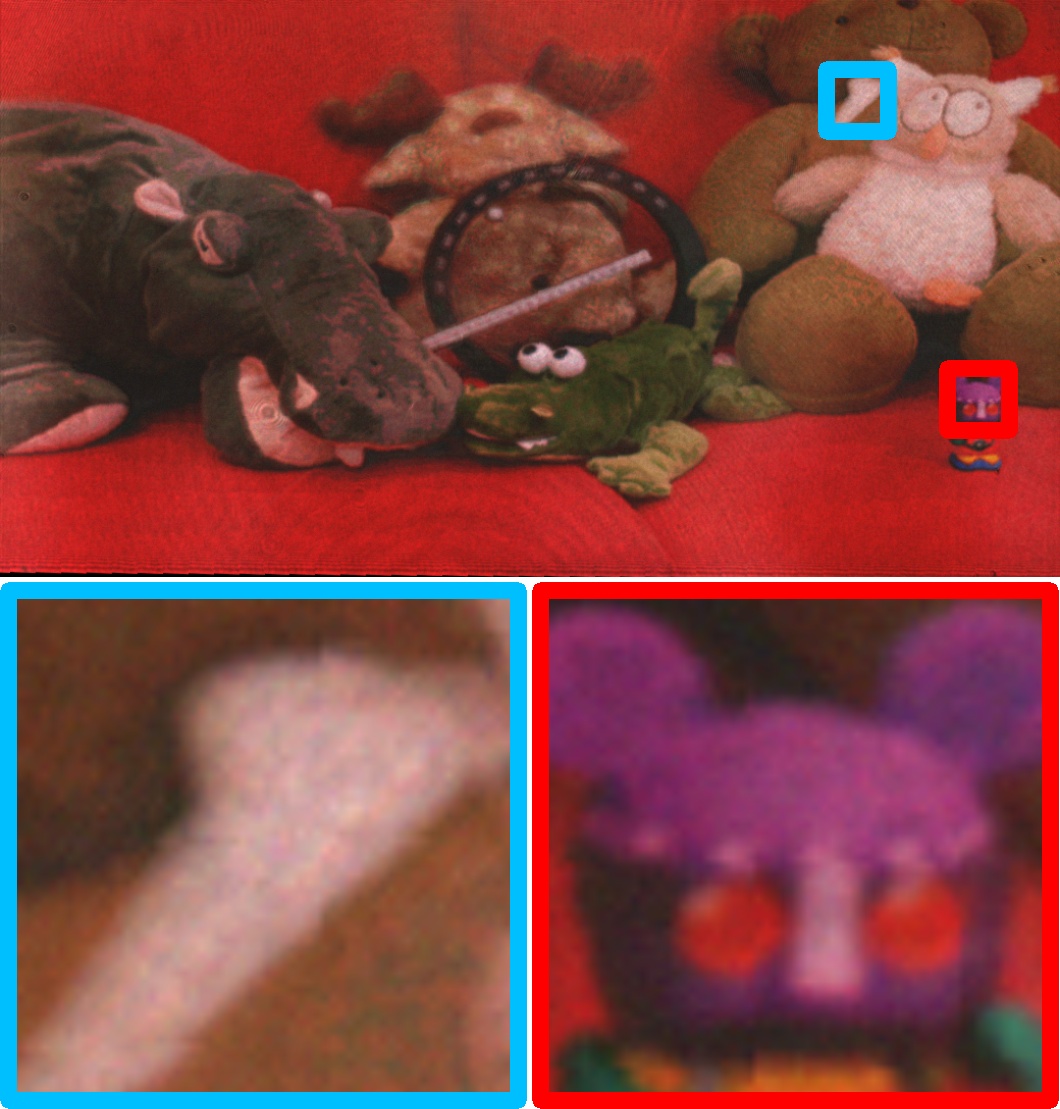}
&\includegraphics[width=3.35cm]{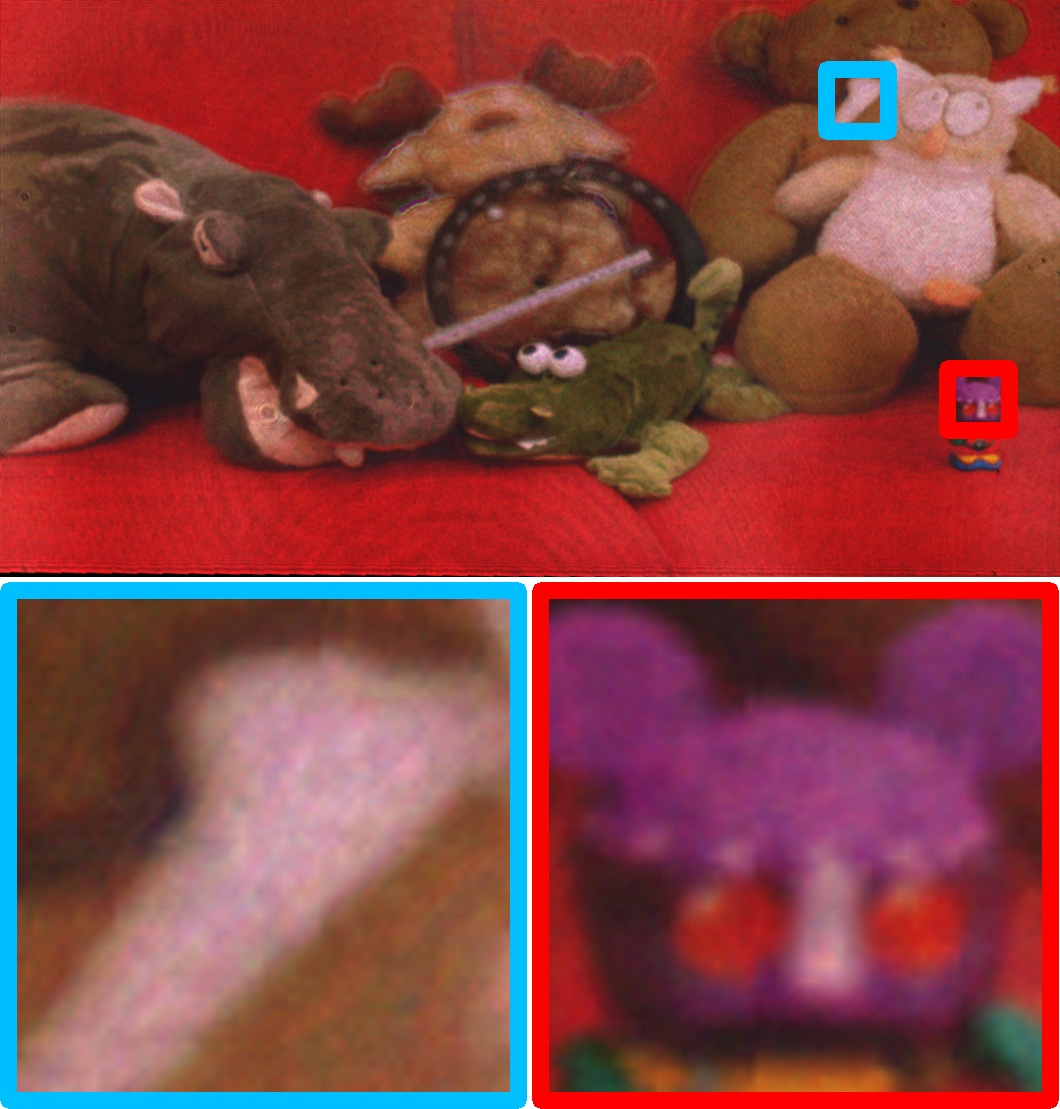}
&\includegraphics[width=3.35cm]{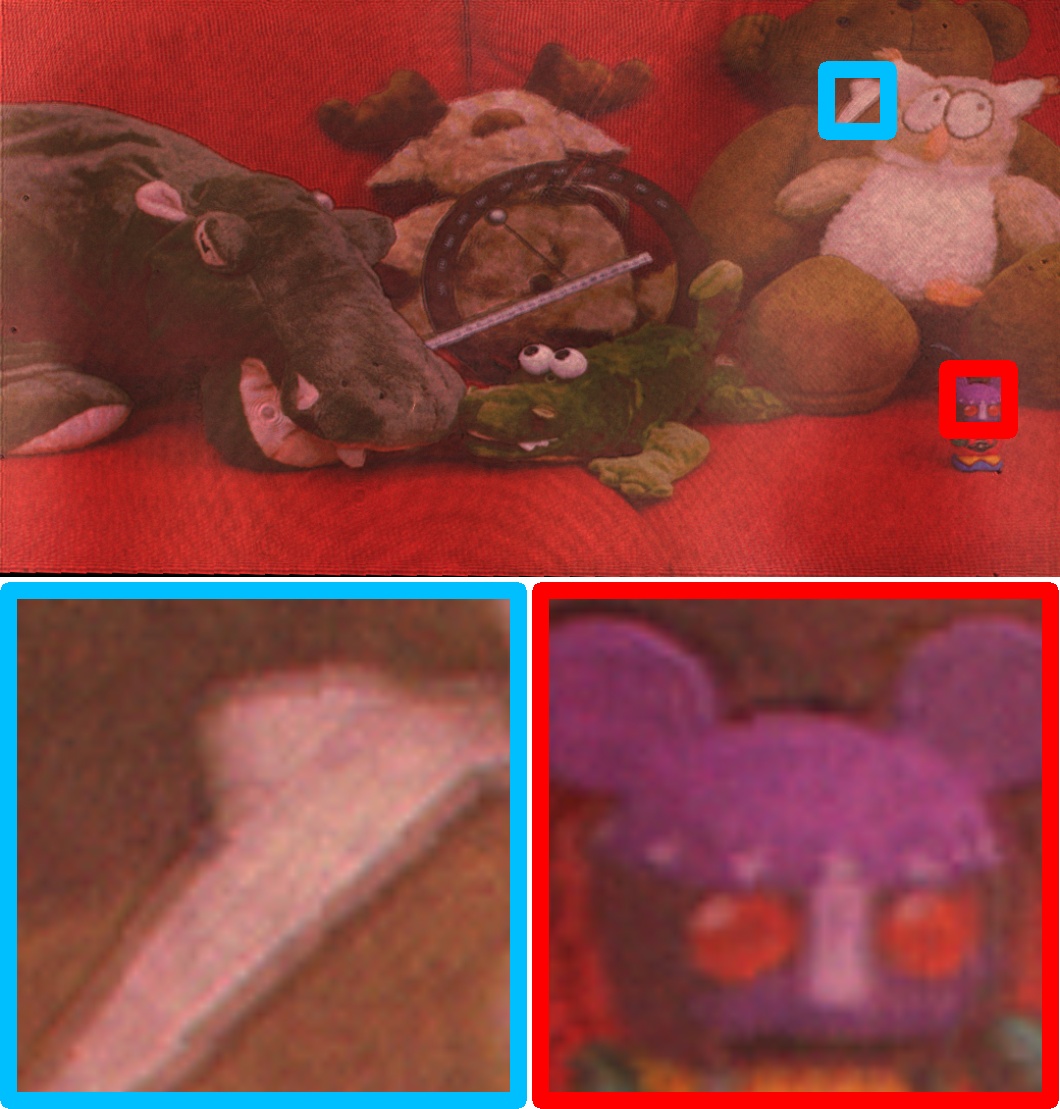}
&\includegraphics[width=3.35cm]{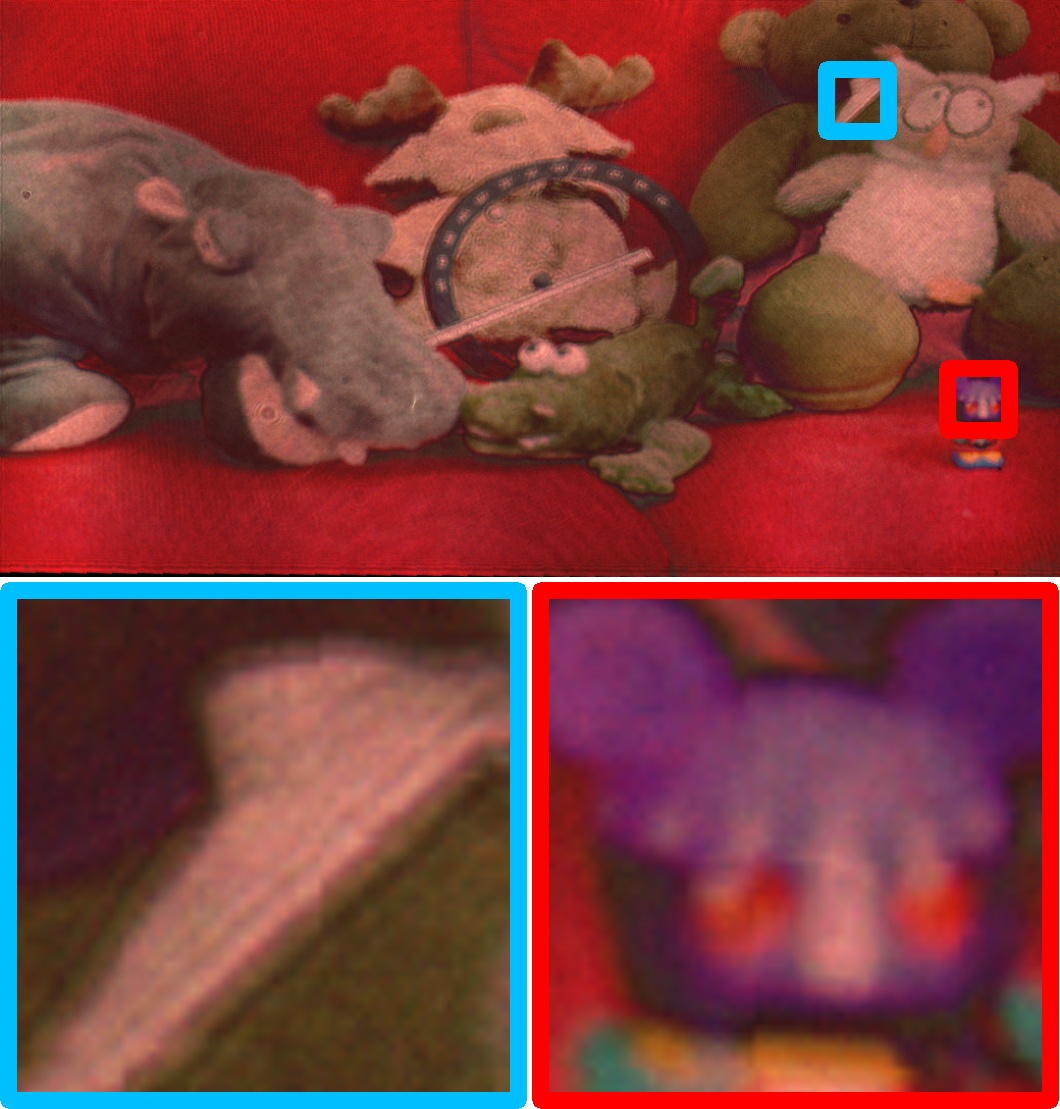}
\\[1.8ex]
\raisebox{0.7\height}{\rotatebox{90}{\textbf{Far Focus}}}
&\includegraphics[width=3.35cm]{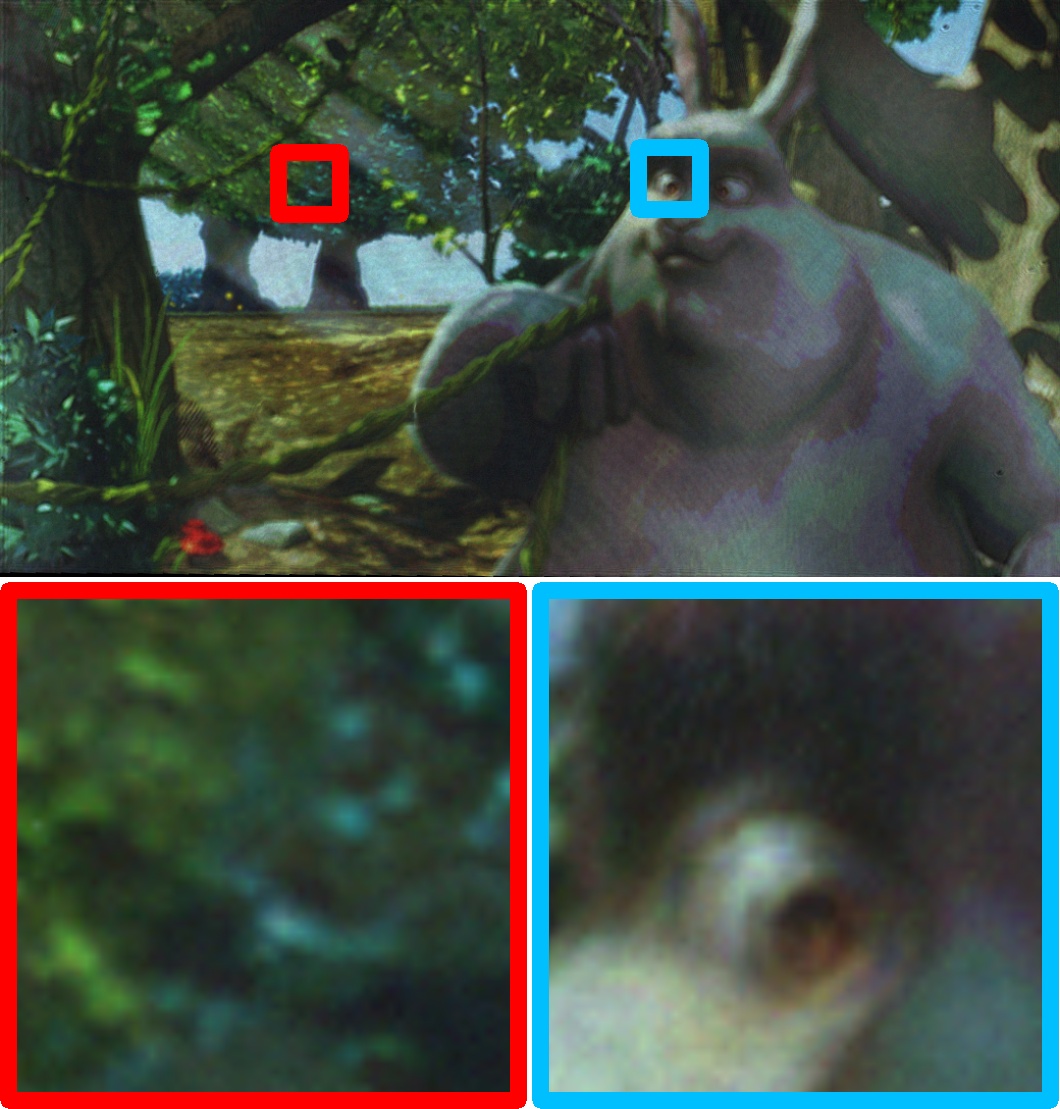}
&\includegraphics[width=3.35cm]{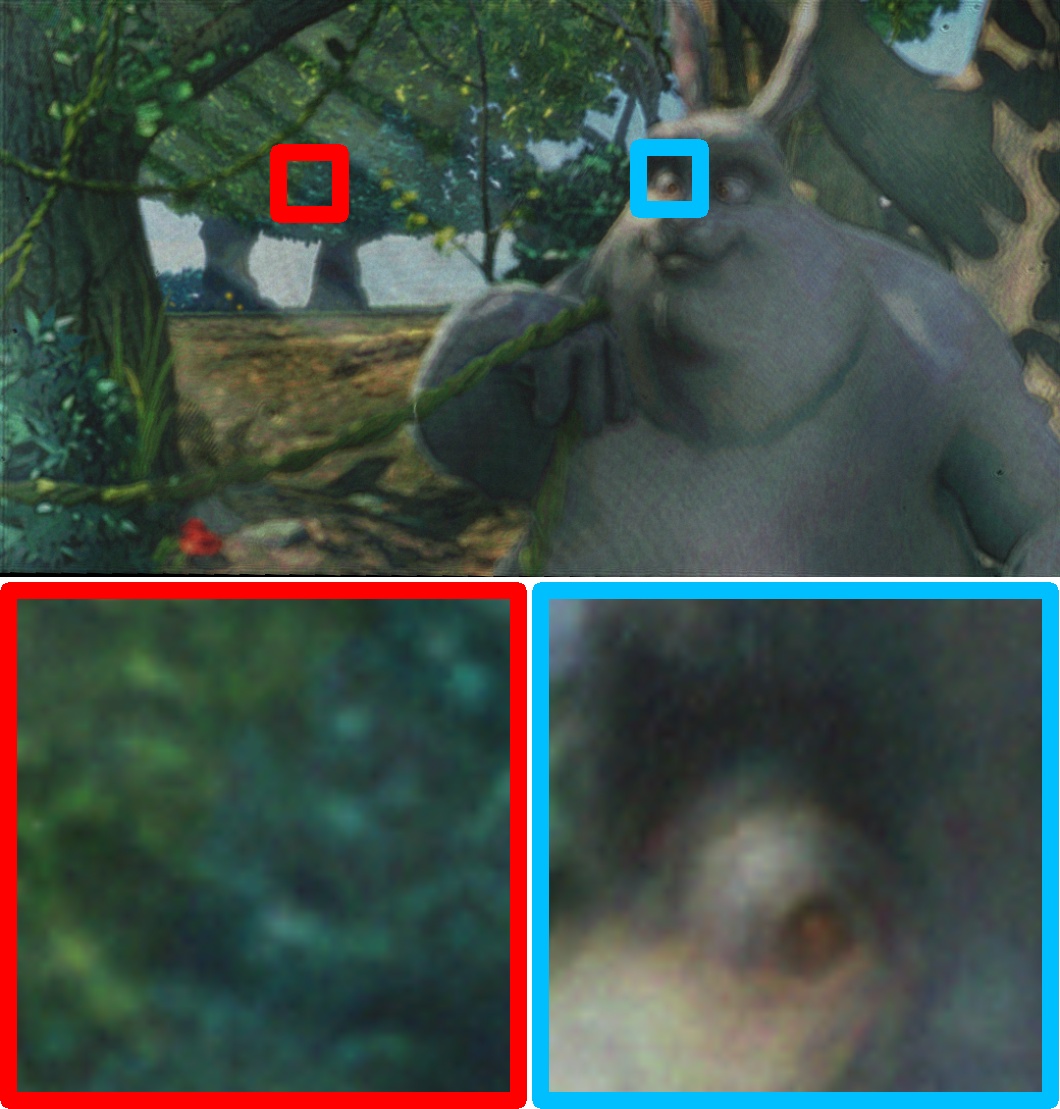}
&\includegraphics[width=3.35cm]{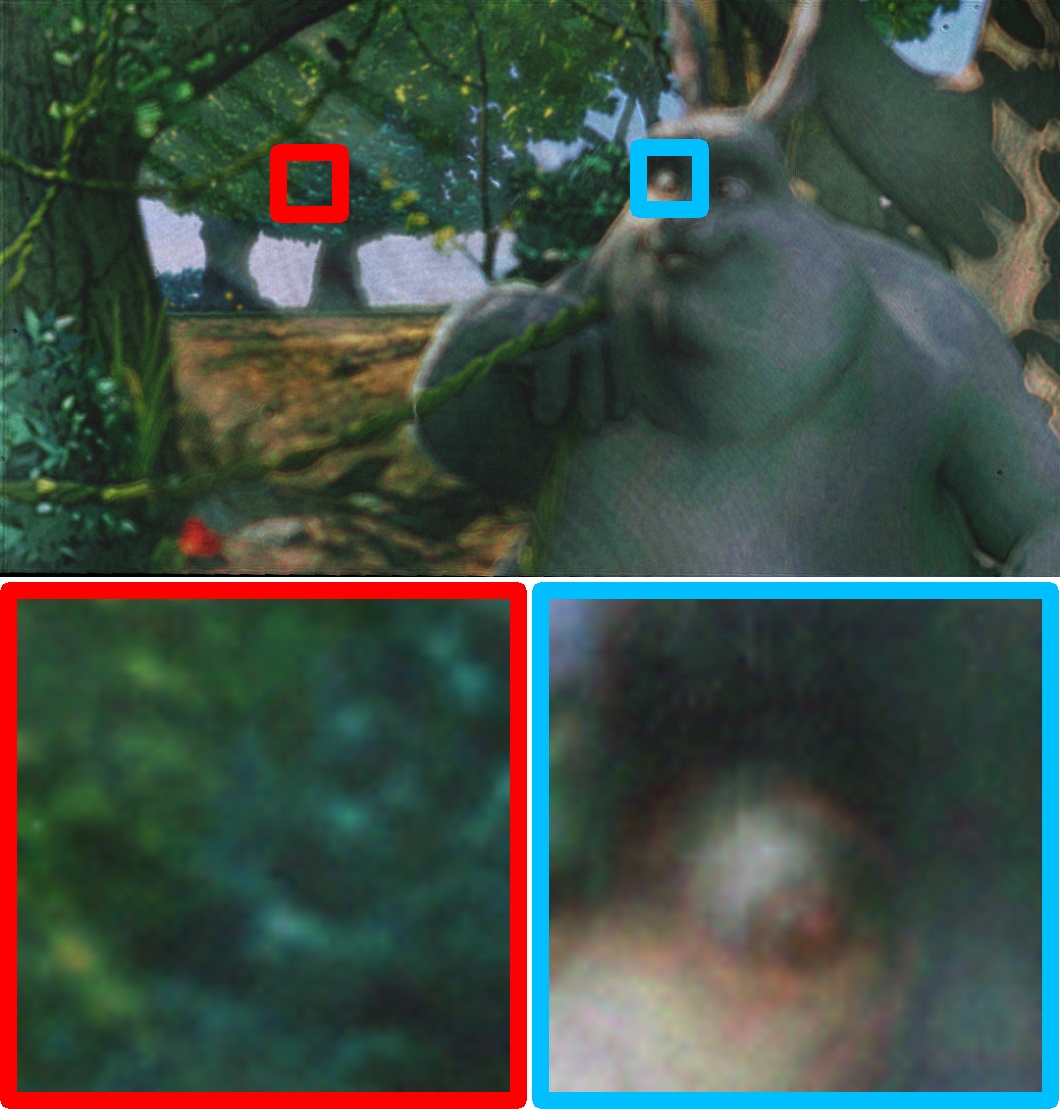}
&\includegraphics[width=3.35cm]{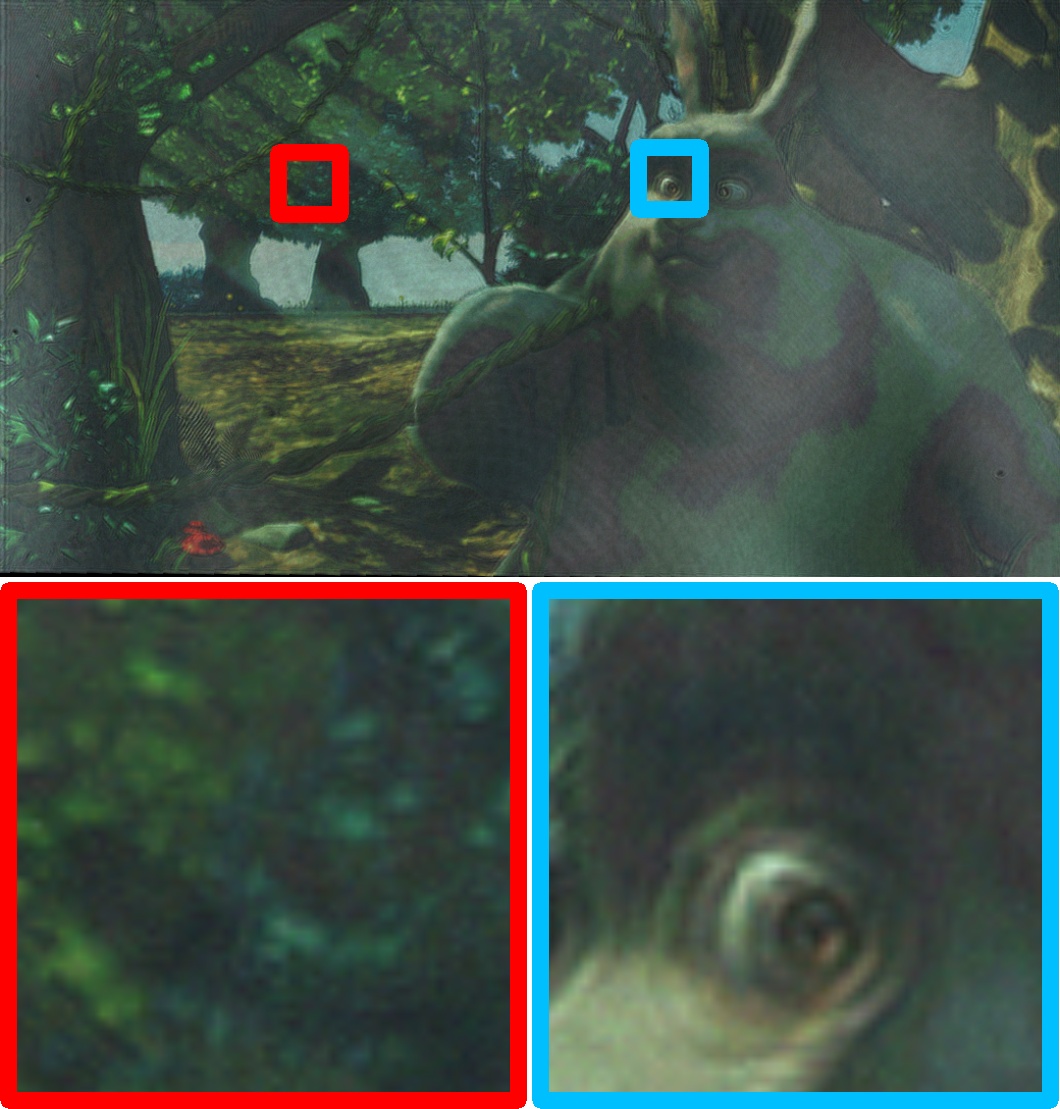}
&\includegraphics[width=3.35cm]{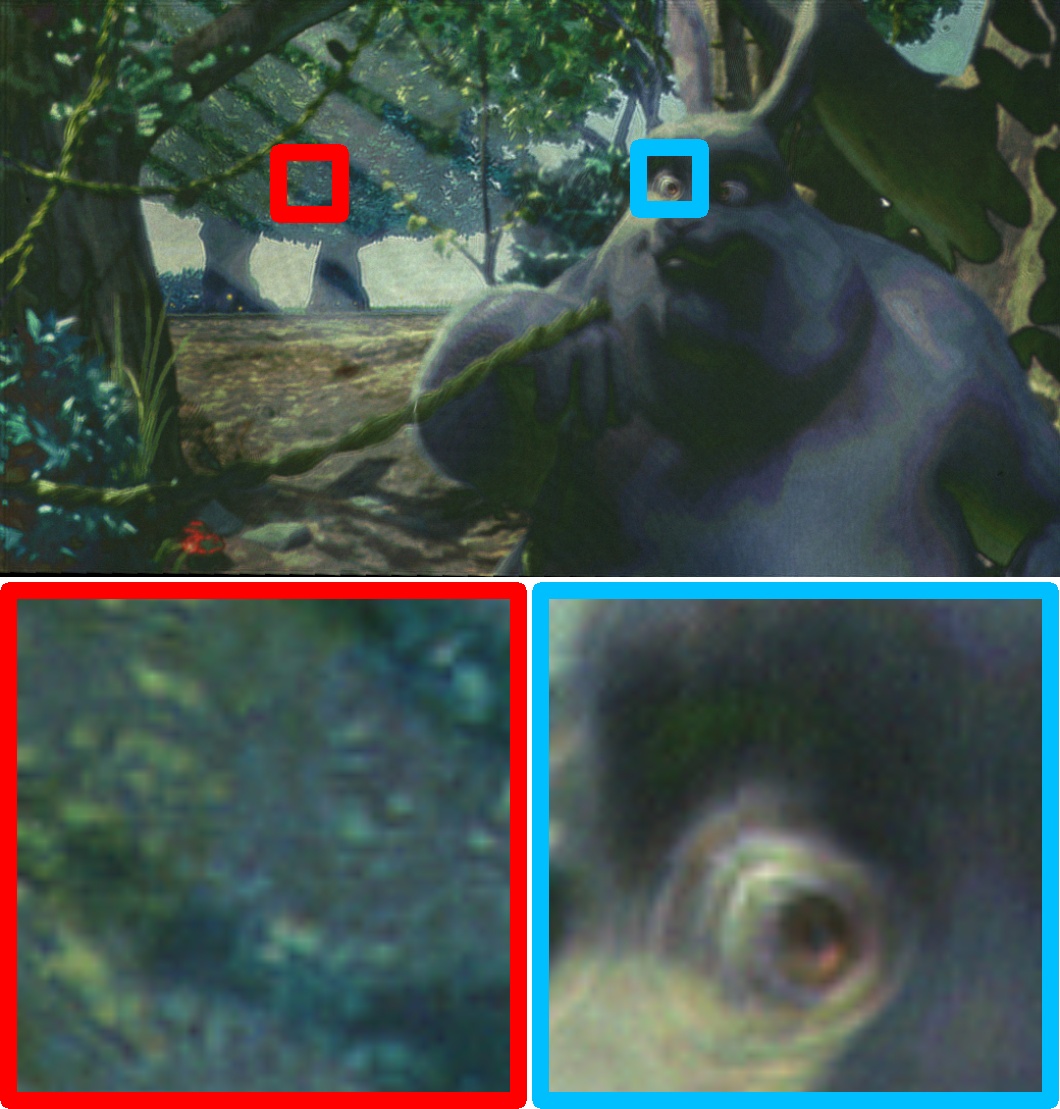}
\\ [1.2ex]
\raisebox{0.7\height}{\rotatebox{90}{\textbf{Near Focus}}}
&\includegraphics[width=3.35cm]{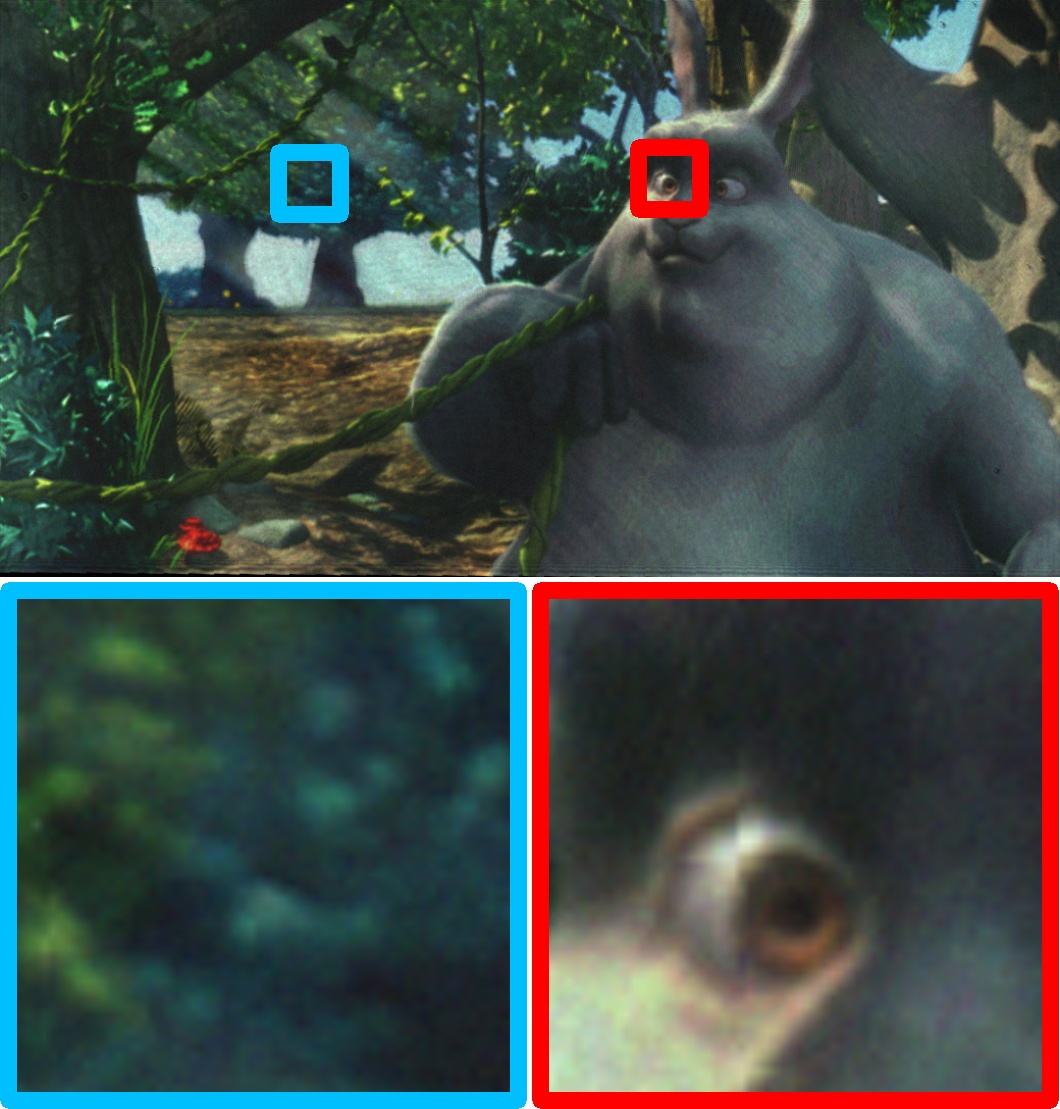}
&\includegraphics[width=3.35cm]{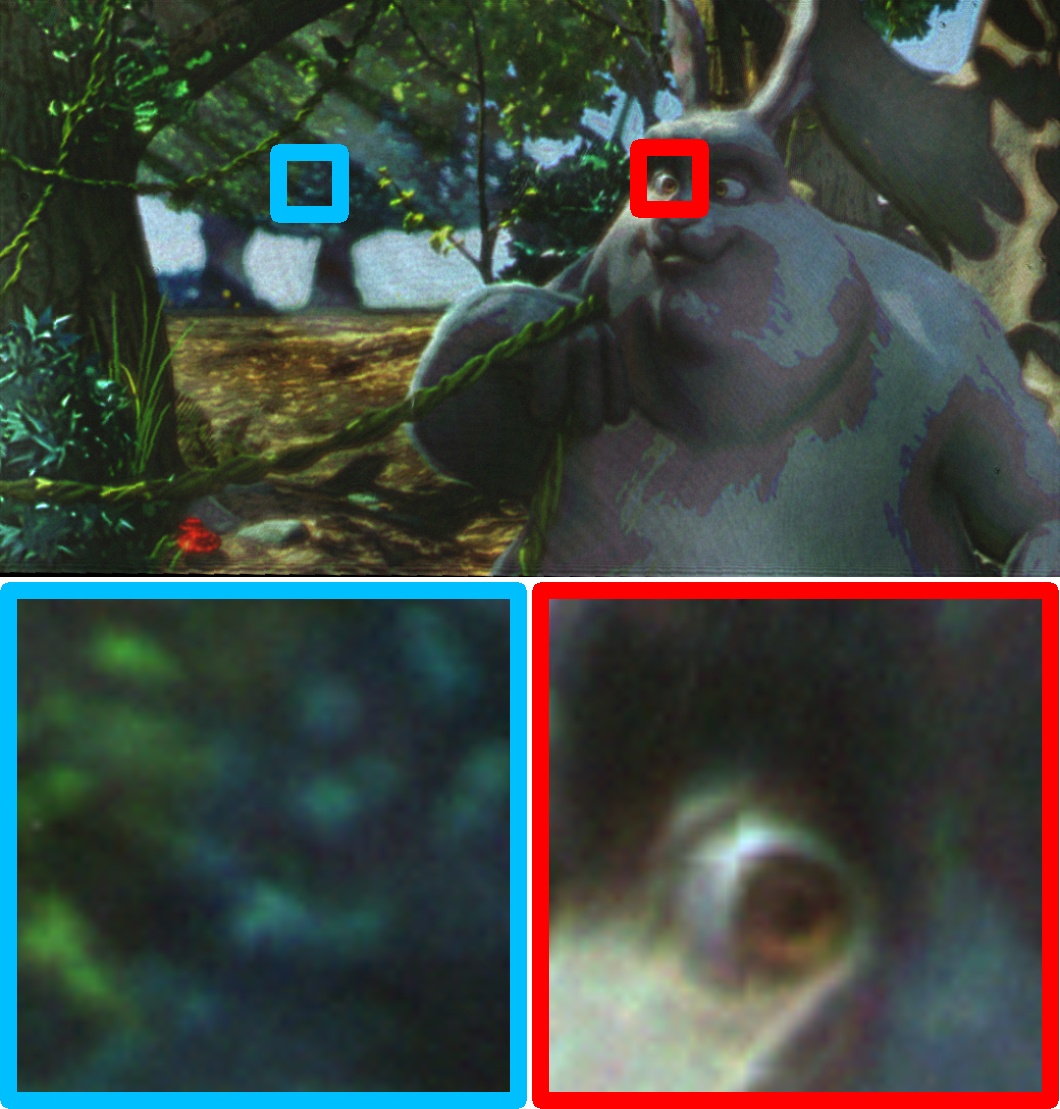}
&\includegraphics[width=3.35cm]{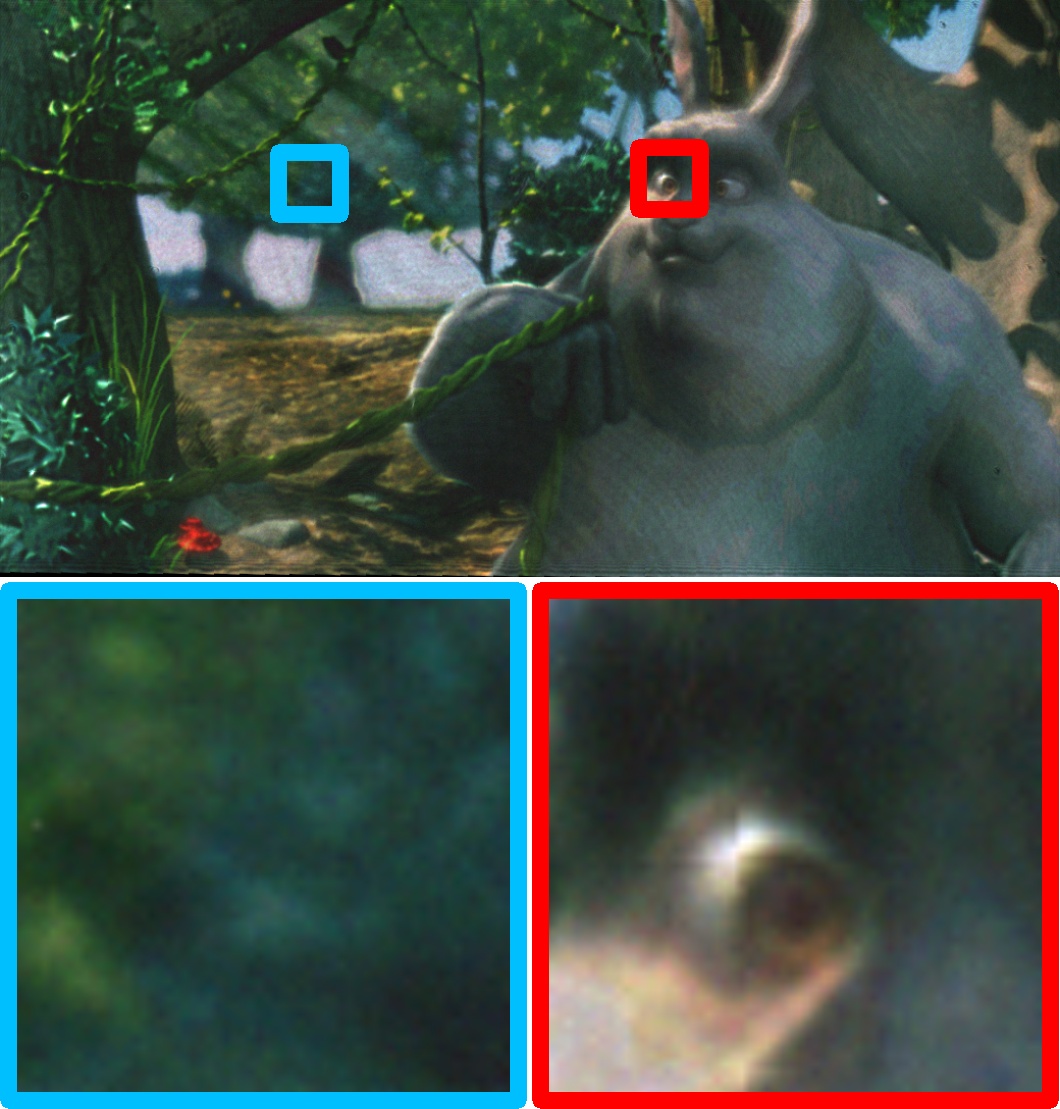}
&\includegraphics[width=3.35cm]{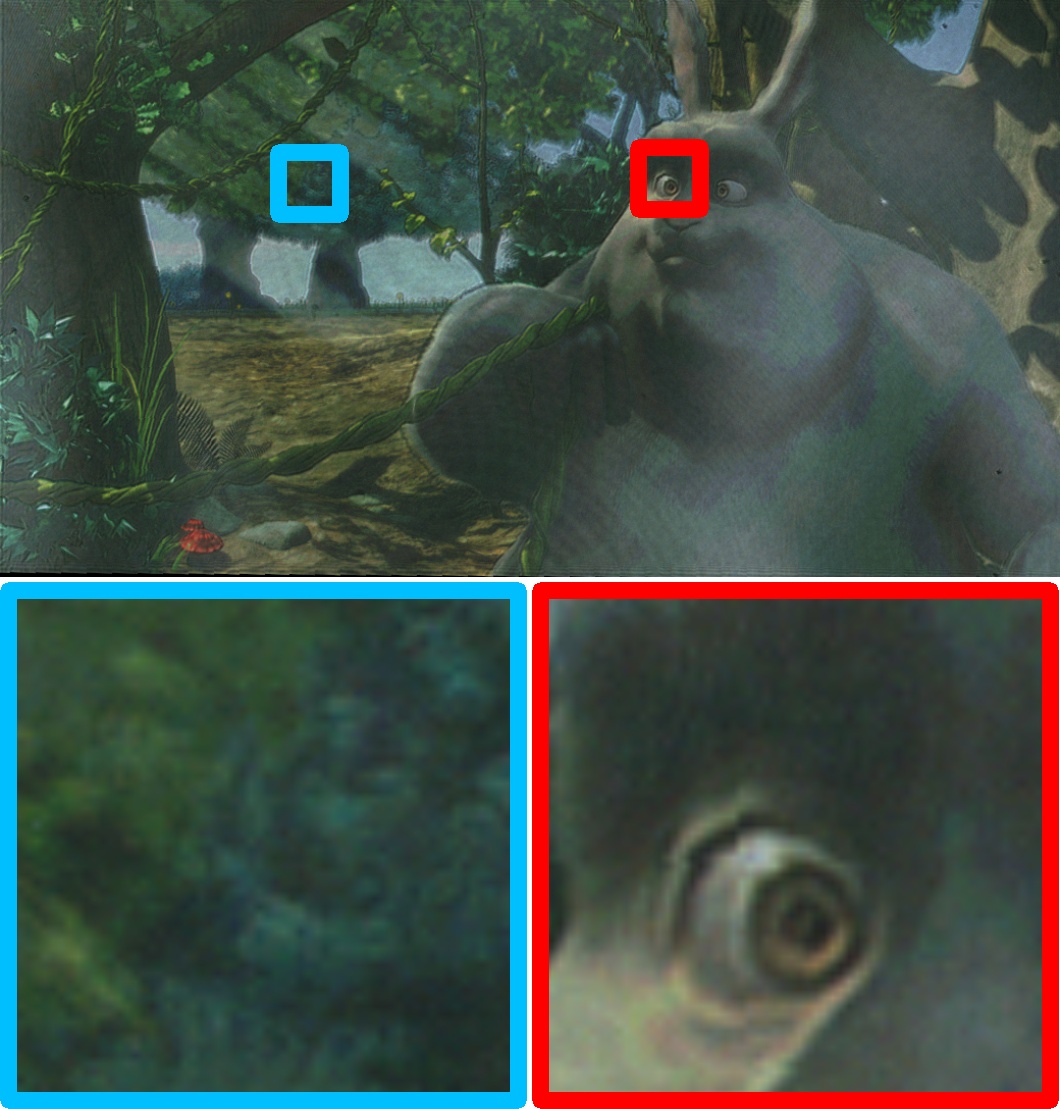}
&\includegraphics[width=3.35cm]{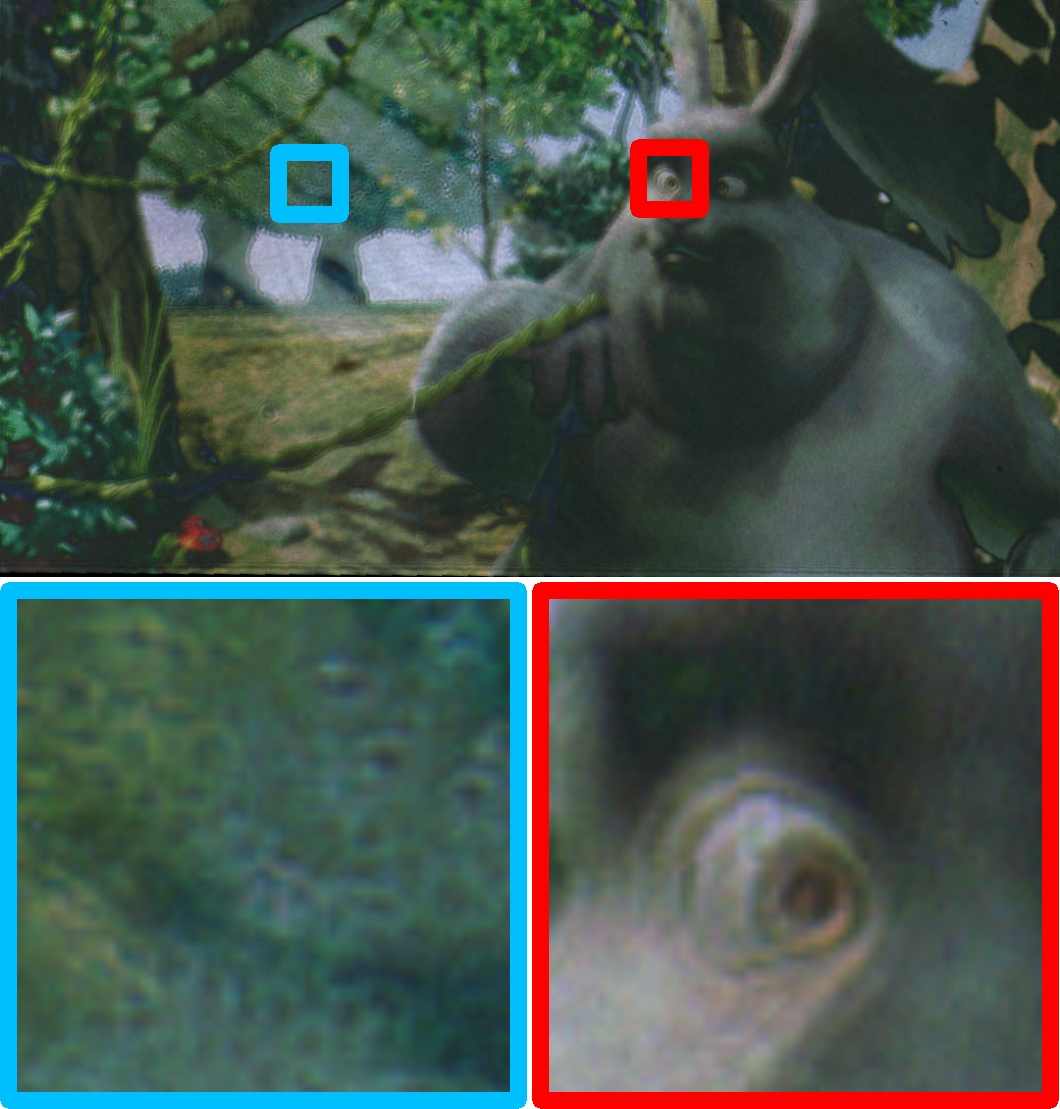}
\\
\end{tabular}
\end{center}
\caption{Experimental evaluation of various hologram methods. In-focus regions are highlighted in red and out-of-focus regions are marked in blue. Only one group of results is provided for methods that do not consider varying pupil sizes and corresponding depth-of-field effects. As shown in the insets, \emph{e.g.}, the ear of a distant plush toy and the head of the closer toy in the first example, distant leaves and closer eye area in the second example, our results exhibit appropriate defocus effects for various pupil sizes while existing approaches support a fixed pupil size and show ringing artifacts or less apparent defocus effects.}
\label{figure:hardware1}
\end{figure*}
\setlength{\tabcolsep}{2pt}
\renewcommand{\arraystretch}{0.6}
\begin{figure}[t]
\begin{center}
\small
\begin{tabular}{cccccc}

& RGB & Estimated Depth  \\ [0.5ex]
&\includegraphics[width=4.0cm]{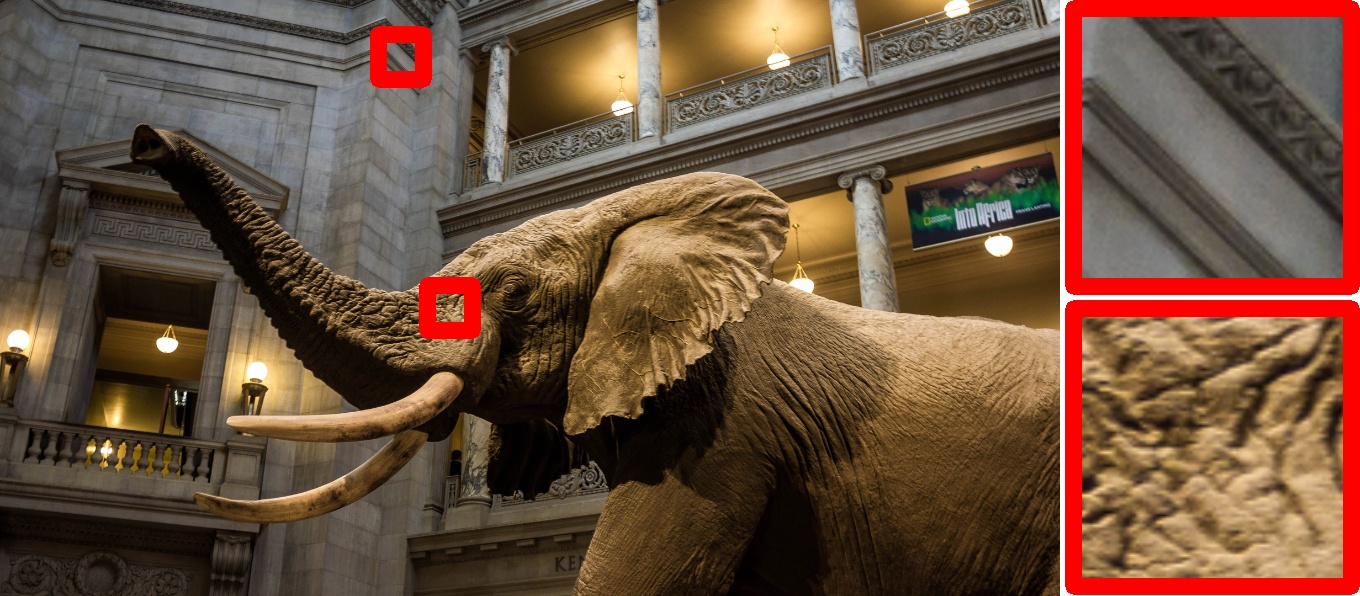}
&\includegraphics[width=4.0cm]{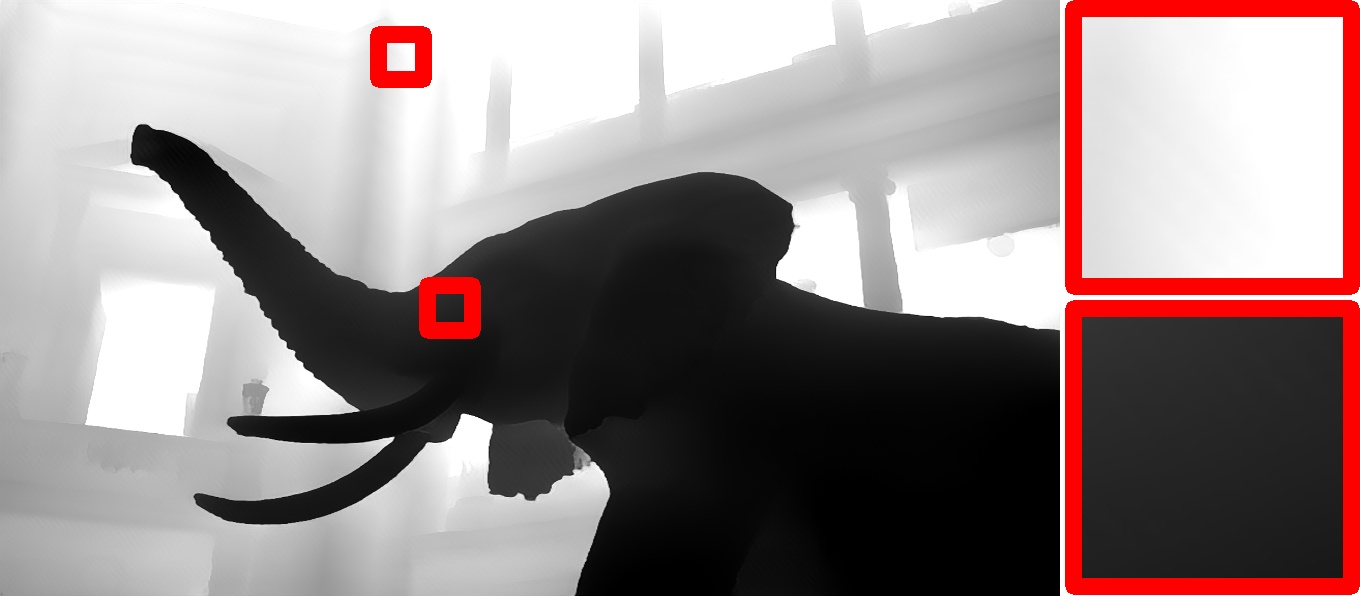}
\\[0.3ex]
& Near Focus & Far Focus  \\ [0.5ex]
&\includegraphics[width=4.0cm]{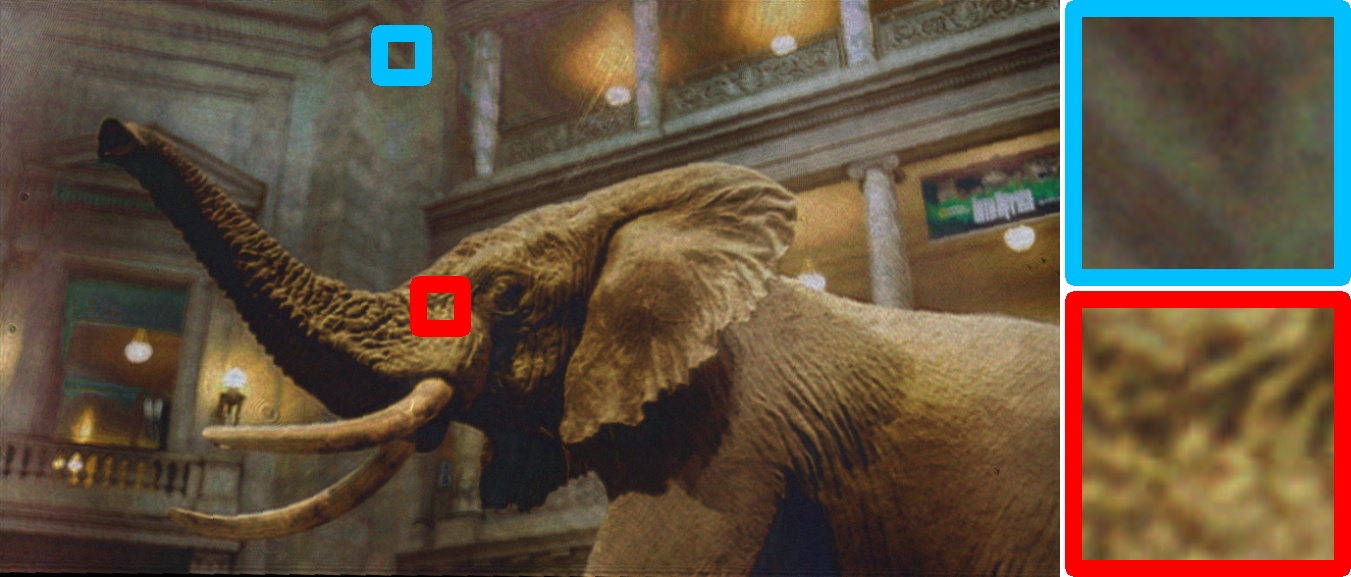}
&\includegraphics[width=4.0cm]{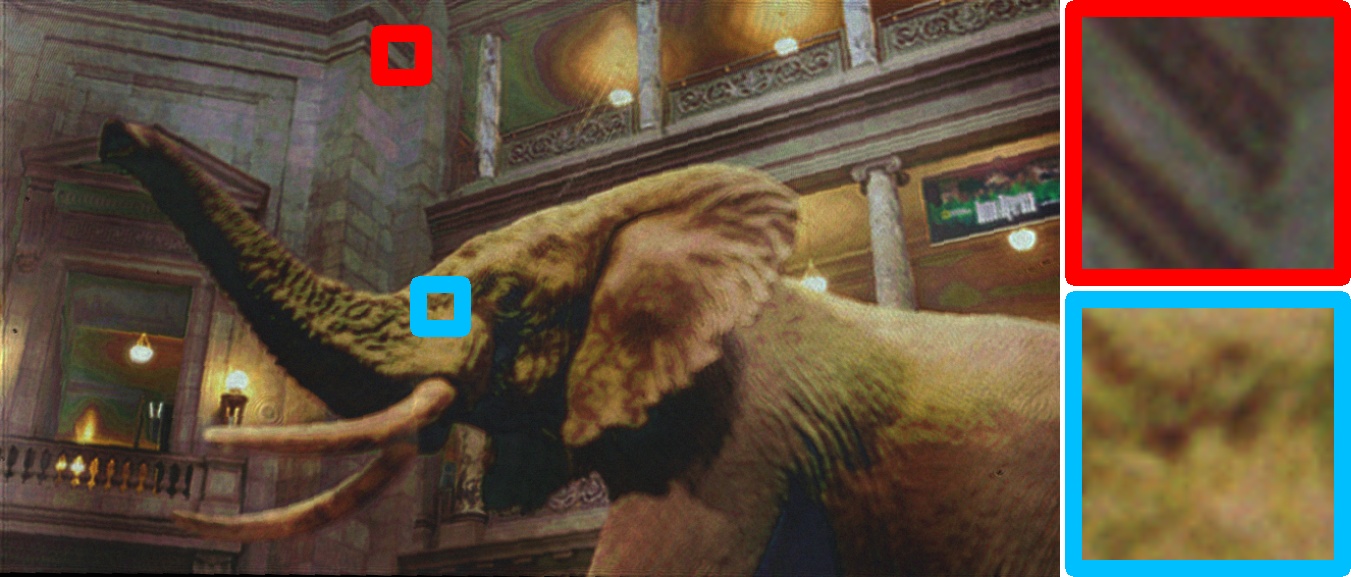}
\\
\end{tabular}
\end{center}
\caption{Experimental results on prototype display. The hologram is synthesized from a single RGB image paired with monocular estimated depth. As shown in the insets, our method is successful in producing natural defocus effects for single input RGB images with monocular depth priors.}
\label{figure:hardware_rgb}
\end{figure}


\revise{
\subsection{Adaptability to Continuous Pupil Variations} \label{sec:pupil-generalization}
In practical scenarios, the pupil size is often a continuous variable. Therefore, we tested the proposed framework with a range of incrementally increasing pupil sizes. 
As depicted in \Cref{figure:interp}, our framework consistently delivered satisfactory results across this spectrum of pupil parameters.
It is also noteworthy that the framework exhibited the ability to adapt the depth-of-field effects for pupil sizes falling beyond the training range, such as pupil sizes of ${1.0\:}\text{mm}, {1.5\:}\text{mm}, {4.5\:}\text{mm}$, and ${5.0\:}\text{mm}$. This adaptability underscores the framework's effectiveness in dynamically adjusting the receptive field to accommodate varying depth-of-field effects.
}

\section{Experimental Assessment}
\revise{
We validate our method on a hardware prototype display and compare with state-of-the-art real-time CGH frameworks, TensorHolo V1 \cite{shi2021nature} and V2 \cite{shi2022light}. 
We provide the experimental results in  \Cref{figure:hardware1}, \Cref{figure:hardware2} and additional results can be found in the Supplementary Material.
As \Cref{figure:hardware2} shows, the proposed approach is able to synthesize 3D holograms that can mimic defocus effects caused due to incoherent light, mitigating the severe artifacts occurring in existing 3D CGH methods. 
For example, holograms produced by the methods from Shi et al.~\cite{shi2021nature,shi2022light} show ringing artifacts, especially around the edges, in the out-of-focus regions. For instance, noticeable ringing artifacts can be observed around the foreground toy within the far-focus image of the first scene or the eyes of the bunny in the second scene as shown in \Cref{figure:hardware1}. 
In contrast, the captured results for the proposed framework show more apparent and natural focus cues and demonstrate varying depth-of-fields, mimicking the effects of incoherent light on a coherent laser-based holographic display.
In \Cref{figure:hardware_rgb}, we also demonstrate the generalizability of our framework to other image-based modalities by providing the experimentally captured results for a hologram synthesized from a single RGB image. As can be seen, the generated holograms produce plausible effects, further demonstrating the feasibility of our framework adapting to other modalities such as single RGB images.} 

\section{Conclusion and Future Work}
We propose a holographic display method that adapts to the viewer's pupil size and renders correct and photorealistic defocus effects, beyond the coherent depth-of-field of holographic offered by today's holographic display methods. Presenting defocus effects offered by incoherent light, as seen in the real world, on coherent light based holographic displays is a severely under-investigated problem. 
This is due to the limited {\'etendue} of spatial light modulators which limits holographic reconstructions with coherent properties such as speckle noise and limited depth-of-field resulting in unnatural defocus blur. To add to this, modern deep learning based computer-generated holography approaches are limited to small depth ranges and do not generalize to larger depths due to the limited receptive fields of convolutional neural networks.

\rerevise{In this work, we provide a glimpse into the performance of existing methods in achieving pupil-dependent depth-of-field effects. Considering the revealed trade-off, }
we propose a learning based approach for generating holograms beyond the coherent depth-of-field via dyanmic receptive fields, in a pupil-adaptive manner. 
We employ a rendering framework to generate photorealistic reconstructions of complex scenes as seen by an eye in different pupil settings, and use this model to shape the diffraction pattern of our holographic display to mimic incoherent light, thereby offering photorealistic defocus blur that goes beyond typical holographic displays. 
Overall, the proposed method not only renders realistic focus cues but also ensures high image fidelity as validated in our synthetic and real hardware experiments. 

\rerevise{
\paragraph{Limitations and future works} 
Despite achieving an interactive speed, the proposed framework does not achieve a real-time speed due to the increased computational complexity resulting from designs that aim for adaptive receptive fields.
Simultaneously improving the speed of the framework and maintaining display quality through a combination of algorithmic and hardware-specific strategies is a promising direction for future works.
The pupil-adaptive holographic display introduced in this study requires a real-time eyetracker, meaning the performance of the eyetracker will influence the practical display experience.
While the current scope is centered on achieving pupil-dependent depth-of-fields, future endeavors will focus on conducting comprehensive subjective studies to assess the practical impacts of the eye tracker, 
by constructing hardware display prototypes with eye trackers and systematically crafting research tests.
Recent research \cite{blurImpact2022} shows that artificial defocus blur affects the perceived realism in actual displays.
It is important to note that our framework is not tied to a particular defocus blur model. Instead, it can be viewed as a versatile framwork that can be easily adapted to various defocus blur models.
We plan to conduct identify the most suitable defocus blur models for human vision through subjective studies in future works, which can then be integrated into our framework.
With this work, however, we make a first step towards pupil-adaptive 3D holography for eyewear displays of the future, which may very likely be holographic, with the aim of rendering virtual imagery indistinguishable from the real world.
}


%



\bibliographystyle{acmart}
{
\bibliography{reference}
}

\end{document}